\RequirePackage{fix-cm}

\documentclass[twocolumn]{svjour3}
\smartqed 

\usepackage{graphicx}

\usepackage{cite}

\usepackage{times}
\usepackage{epsfig}
\usepackage{graphicx}
\usepackage{graphics}
\usepackage{amsmath}
\usepackage{amssymb}
\usepackage{textcomp}
\usepackage{tabularx}
\usepackage{rotating}
\usepackage{verbatim}
\usepackage{colortbl}
\usepackage[table]{xcolor}
\usepackage{caption}
\usepackage{booktabs}
\usepackage{gensymb}
\usepackage{mathtools}
\usepackage{subfig}
\usepackage{multirow,stackengine}
\usepackage[normalem]{ulem}

\usepackage[letterspace=100]{microtype}

\newcolumntype{$}{>{\global\let\currentrowstyle\relax}}
\newcolumntype{^}{>{\currentrowstyle}}
\newcommand{\rowstyle}[1]{\gdef\currentrowstyle{#1}%
	#1\ignorespaces
}

\usepackage{pifont}

\usepackage[pagebackref=true,breaklinks=true,letterpaper=true,colorlinks,bookmarks=false]{hyperref}

\newcommand\blfootnote[1]{
	\begingroup
	\renewcommand\thefootnote{}\footnote{#1}
	\addtocounter{footnote}{-1}
	\endgroup
}

\begin{document}
	
	\title{3D Semantic Scene Completion: a Survey%
	}

	\author{Luis Rold\~ao\textsuperscript{1,2}         \and
		Raoul de Charette\textsuperscript{1}     \and
		Anne Verroust-Blondet\textsuperscript{1}
	}

	\date{}

	\twocolumn[{
		\renewcommand\twocolumn[1][]{#1}%
		\maketitle
		
		\begin{center}
			
			\centering
			
			\renewcommand{\arraystretch}{0.6}
			\begin{tabular}{c c c}
				
				\multicolumn{2}{c}{\scriptsize INDOOR} & \scriptsize OUTDOOR \\
				
				\cmidrule(lr){0-1} \cmidrule(lr){3-3}
				
				\scriptsize \stackunder{\textbf{NYUv2}~\cite{Silberman2012IndoorSA} \textbf{/ NYUCAD}~\cite{Guo2018ViewVolumeNF}}{(Real-world / Synthetic)}
				& \scriptsize \stackunder{\textbf{SUNCG}~\cite{Song2017SemanticSC}}{(Synthetic)}  
				& \scriptsize \stackunder{\textbf{SemanticKITTI}~\cite{Behley2019SemanticKITTIAD}}{(Real-world)}   \\
				
				\includegraphics[width=0.27\linewidth]{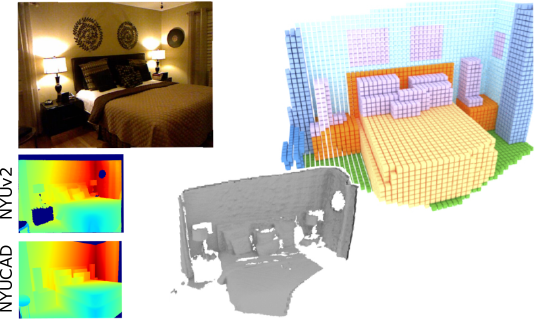} &
				\includegraphics[width=0.30\linewidth]{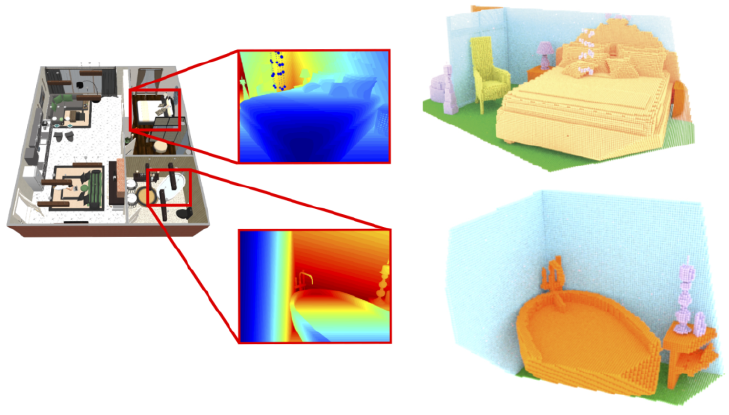} & 
				\includegraphics[width=0.38\linewidth]{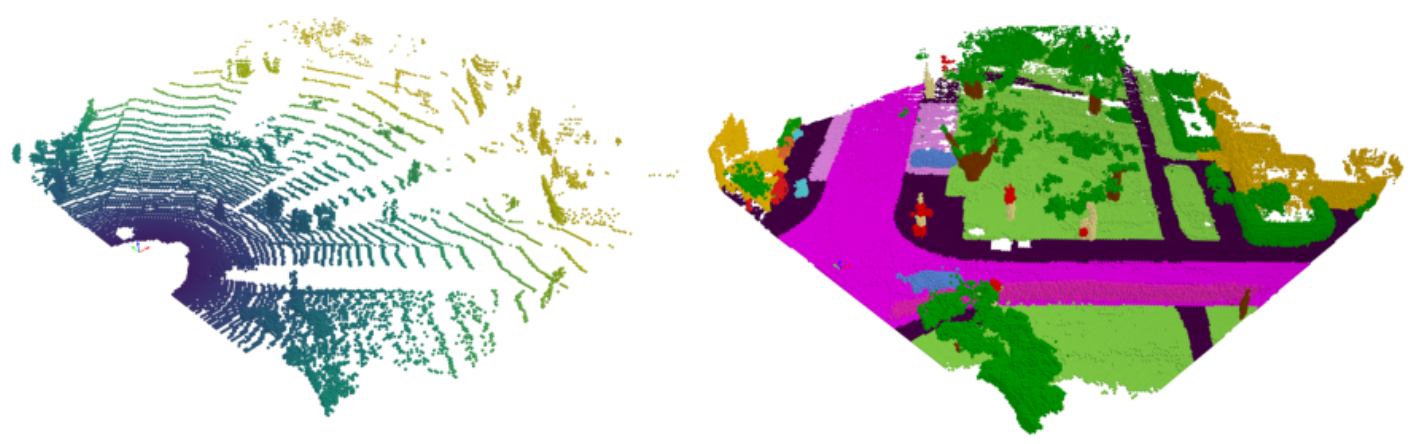} \\
				
				\multicolumn{2}{c}{\includegraphics[width=0.48\linewidth]{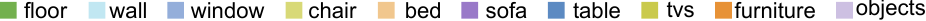}} 
				& \includegraphics[width=0.36\linewidth]{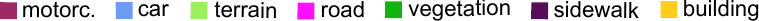} \\
				
			\end{tabular}
					
			\captionof{figure}{\textbf{Popular datasets for Semantic Scene Completion (SSC).} From an incomplete input view, the SSC task consists in the joint estimation of both geometry and semantics of the scene. The figure shows the 4 most popular datasets for SSC, each showing input data and ground truth. The complexity of SSC lies in the completion of unobserved / occluded regions and in the sparse supervision signal (notice that real ground truth is incomplete).
				(Images source \cite{Song2017SemanticSC, Roldao2020LMSCNetLM}).
			}
			\label{fig:teaser}

		\end{center}
	}]
	
	\blfootnote{
		\\
		L. Rold\~ao (luis.roldao@inria.fr / luis.roldao@akka.eu).\\
		R. de Charette (raoul.de-charette@inria.fr).\\
		A. Verroust-Blondet (anne.verroust@inria.fr).\\
		\textsuperscript{1} Inria, Paris, France. \\
		\textsuperscript{2} AKKA Research, Guyancourt, France.
	}
	
	\begin{abstract}
		Semantic Scene Completion (SSC) aims to jointly estimate the complete geometry and semantics of a scene, assuming partial sparse input. 
		In the last years following the multiplication of large-scale 3D datasets, SSC has gained significant momentum in the research community because it holds unresolved challenges. Specifically, SSC lies in the ambiguous completion of large unobserved areas and the weak supervision signal of the ground truth. This led to a substantially increasing number of papers on the matter.\\
		This survey aims to identify, compare and analyze the techniques providing a critical analysis of the SSC literature on both methods and datasets. 
		Throughout the paper, we provide an in-depth analysis of the existing works covering all choices made by the authors while highlighting the remaining avenues of research.
		SSC performance of the SoA on the most popular datasets is also evaluated and analyzed.
	\end{abstract}
	
	\definecolor{car}{rgb}{0.39215686, 0.58823529, 0.96078431}
	\definecolor{bicycle}{rgb}{0.39215686, 0.90196078, 0.96078431}
	\definecolor{motorcycle}{rgb}{0.11764706, 0.23529412, 0.58823529}
	\definecolor{truck}{rgb}{0.31372549, 0.11764706, 0.70588235}
	\definecolor{other-vehicle}{rgb}{0.39215686, 0.31372549, 0.98039216}
	\definecolor{person}{rgb}{1.        , 0.11764706, 0.11764706}
	\definecolor{bicyclist}{rgb}{1.        , 0.15686275, 0.78431373}
	\definecolor{motorcyclist}{rgb}{0.58823529, 0.11764706, 0.35294118}
	\definecolor{road}{rgb}{1.        , 0.        , 1.        }
	\definecolor{parking}{rgb}{1.        , 0.58823529, 1.        }
	\definecolor{sidewalk}{rgb}{0.29411765, 0.        , 0.29411765}
	\definecolor{other-ground}{rgb}{0.68627451, 0.        , 0.29411765}
	\definecolor{building}{rgb}{1.        , 0.78431373, 0.        }
	\definecolor{fence}{rgb}{1.        , 0.47058824, 0.19607843}
	\definecolor{vegetation}{rgb}{0.        , 0.68627451, 0.        }
	\definecolor{trunk}{rgb}{0.52941176, 0.23529412, 0.        }
	\definecolor{terrain}{rgb}{0.58823529, 0.94117647, 0.31372549}
	\definecolor{pole}{rgb}{1.        , 0.94117647, 0.58823529}
	\definecolor{traffic-sign}{rgb}{1.        , 0.        , 0.    }    
	
	\makeatletter
	\newcommand{\car@semkitfreq}{3.92}
	\newcommand{\bicycle@semkitfreq}{0.03}
	\newcommand{\motorcycle@semkitfreq}{0.03}
	\newcommand{\truck@semkitfreq}{0.16}
	\newcommand{\othervehicle@semkitfreq}{0.20}
	\newcommand{\person@semkitfreq}{0.07}
	\newcommand{\bicyclist@semkitfreq}{0.07}
	\newcommand{\motorcyclist@semkitfreq}{0.05}
	\newcommand{\road@semkitfreq}{15.30}  %
	\newcommand{\parking@semkitfreq}{1.12}
	\newcommand{\sidewalk@semkitfreq}{11.13}  %
	\newcommand{\otherground@semkitfreq}{0.56}
	\newcommand{\building@semkitfreq}{14.1}  %
	\newcommand{\fence@semkitfreq}{3.90}
	\newcommand{\vegetation@semkitfreq}{39.30}  %
	\newcommand{\trunk@semkitfreq}{0.51}
	\newcommand{\terrain@semkitfreq}{9.17} %
	\newcommand{\pole@semkitfreq}{0.29}
	\newcommand{\trafficsign@semkitfreq}{0.08}
	\newcommand{\semkitfreq}[1]{{\csname #1@semkitfreq\endcsname}}
	
	\definecolor{ceiling}{rgb}{0.8359375, 0.1484375, 0.15625}
	\definecolor{floor}{rgb}{0.16796875, 0.625, 0.015625}
	\definecolor{wall}{rgb}{0.6171875, 0.84375, 0.89453125}
	\definecolor{window}{rgb}{0.4453125, 0.6171875, 0.8046875}
	\definecolor{chair}{rgb}{0.796875, 0.796875, 0.35546875}
	\definecolor{bed}{rgb}{0.99609375, 0.7265625, 0.46484375}
	\definecolor{sofa}{rgb}{0.57421875, 0.3984375, 0.734375}
	\definecolor{table}{rgb}{0.1171875, 0.46484375, 0.70703125}
	\definecolor{tvs}{rgb}{0.625, 0.734375, 0.12890625}
	\definecolor{furniture}{rgb}{0.99609375, 0.49609375, 0.046875}
	\definecolor{objects}{rgb}{0.765625, 0.68359375, 0.8359375}
	
	\makeatletter
	\newcommand{\ceiling@nyuvtwofreq}{0.74}
	\newcommand{\floor@nyuvtwofreq}{12.44}
	\newcommand{\wall@nyuvtwofreq}{9.67}
	\newcommand{\window@nyuvtwofreq}{2.12}
	\newcommand{\chair@nyuvtwofreq}{2.03}
	\newcommand{\bed@nyuvtwofreq}{9.17}
	\newcommand{\sofa@nyuvtwofreq}{6.78}
	\newcommand{\table@nyuvtwofreq}{4.14}
	\newcommand{\tvs@nyuvtwofreq}{0.53}  %
	\newcommand{\furniture@nyuvtwofreq}{36.64}
	\newcommand{\objects@nyuvtwofreq}{15.74}  %
	\newcommand{\nyuvtwofreq}[1]{{\csname #1@nyuvtwofreq\endcsname}}

	\makeatletter
	\newcommand{\ceiling@suncgfreq}{2.68}
	\newcommand{\floor@suncgfreq}{12.27}
	\newcommand{\wall@suncgfreq}{33.55}
	\newcommand{\window@suncgfreq}{5.79}
	\newcommand{\chair@suncgfreq}{1.80}
	\newcommand{\bed@suncgfreq}{5.95}
	\newcommand{\sofa@suncgfreq}{4.94}
	\newcommand{\table@suncgfreq}{2.90}
	\newcommand{\tvs@suncgfreq}{0.36}  %
	\newcommand{\furniture@suncgfreq}{15.04}
	\newcommand{\objects@suncgfreq}{14.73}  %
	\newcommand{\suncgfreq}[1]{{\csname #1@suncgfreq\endcsname}}
	
	\section{Introduction }
\label{sec:intro}

Understanding the 3D surroundings is a natural ability for humans, who are capable to leverage prior knowledge to estimate geometry and semantics, even in large occluded areas.
This proves more difficult for computers, which has drawn wide interest from computer vision researchers in recent years~\cite{Guo2020DeepLF}.
Indeed, 3D scene understanding is a crucial feature for many applications, such as robotic navigation or augmented reality, where geometrical and semantics understanding is key to leverage interaction with the real world~\cite{GargSDMCCW21}.
Nonetheless, vision sensors only provide partial observations of the world given their limited field of view, sparse sensing, and measurement noise, capturing a partial and incomplete representation of the scene.

To address this, Scene Completion (SC) has been used for a long time to infer the complete geometry of a scene given one or more 2D/3D observations, historically using a wide variety of more or less sophisticated interpolation methods. 
With recent advances of 3D deep learning, Semantic Scene Completion (SSC) has been introduced as an extension of SC, where semantics and geometry are \textit{jointly} inferred for the whole scene, departing from the idea that they are entangled~\cite{Song2017SemanticSC}. 
Consequently, along with the addition of semantics, the SSC task has significantly departed from original SC in terms of nature and sparsity of the input data.

Fig.~\ref{fig:teaser} shows samples of input and ground truth for the four most popular SSC datasets. 
The complexity of the SSC task lies in the sparsity of the input data (see holes in depth or LiDAR input), and the incomplete ground truth (resulting of frame aggregation) providing a rather weak guidance. Different from object-level completion~\cite{Avetisyan2019Scan2CADLC,Yuan2018PCNPC} or from scene reconstruction where multiple views are aggregated~\cite{ZollhoferSGTNKK18,PintoreMGPPG20}, 
SSC requires in-depth understanding of the entire scene heavily relying on learned priors to resolve ambiguities. 
The increasing number of large scale datasets~\cite{Behley2019SemanticKITTIAD, Song2017SemanticSC, Silberman2012IndoorSA, Straub2019TheRD} have encouraged new SSC works in the last years. 
On connex topics, 3D (deep) vision has been thoroughly reviewed~\cite{lu2020deep,Guo2020DeepLF,liu2019deep,LiMZLC20} including surveys on 3D representations~\cite{ahmed2018survey} and task-oriented reviews like 3D semantic segmentation~\cite{zhang2019review,xie2020linking}, 3D reconstruction~\cite{ZollhoferSGTNKK18,PintoreMGPPG20}, 3D object detection~\cite{jiao2019survey}, etc. Still, no survey seems to exist on this hot SSC topic and navigating the literature is not trivial.

In this article, we propose the first comprehensive and critical review of the Semantic Scene Completion~(SSC) literature, focusing on methods and datasets. 
With this systematic survey, we expose a critical analysis of the SSC knowledge and highlight the missing areas for future research.
We aim to provide new insights to informed readers and help new ones navigate in this emerging field which gained significant momentum in the past few years. 
To the best of our knowledge this is the first SSC survey, which topic has only been briefly covered in recent 3D surveys~\cite{lu2020deep,Guo2020DeepLF}.

To study SSC, this paper is organized as follows. 
We first introduce and formalize the problem of Semantic Scene Completion in Sec.~\ref{sec:problem}, briefly brushing closely related topics. Ad-hoc datasets employed for the task and introduction to common 3D scene representations are covered in Sec.~\ref{sec:datasets_scene_rep}. We study the existing works in Sec.~\ref{sec:3d_scene_completion}, highlighting the different input encodings, deep architectures, design choices, and training strategies employed. The section ends with an analysis of the current performances, followed by a discussion in Sec.~\ref{sec:discussion}.

	\section{Problem statement} \label{sec:problem}

\begin{figure}[!tbp]
	\centering
	\includegraphics[width=1.0\linewidth]{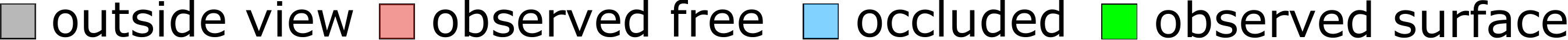} \\
	\subfloat[Cameras]{\includegraphics[width=0.45\linewidth]{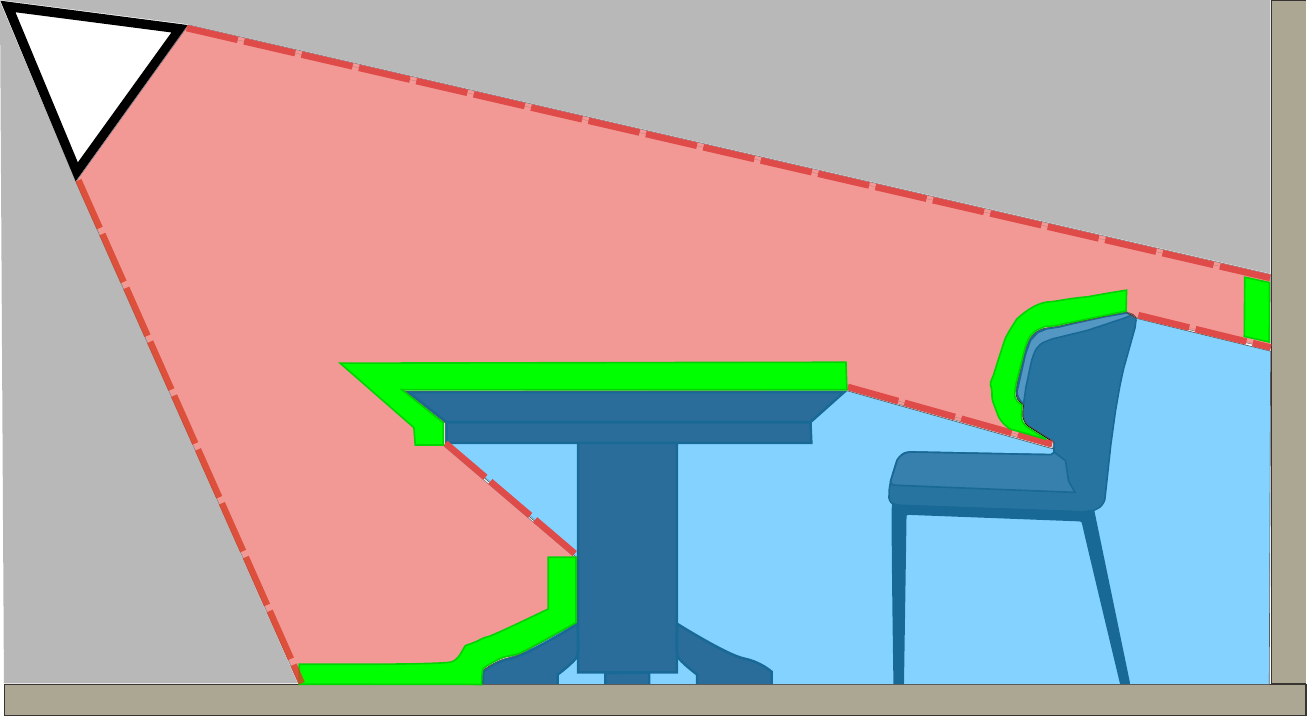}\label{fig:sensors_RGBD}}
	\hfill
	\subfloat[LiDAR]{\includegraphics[width=0.45\linewidth]{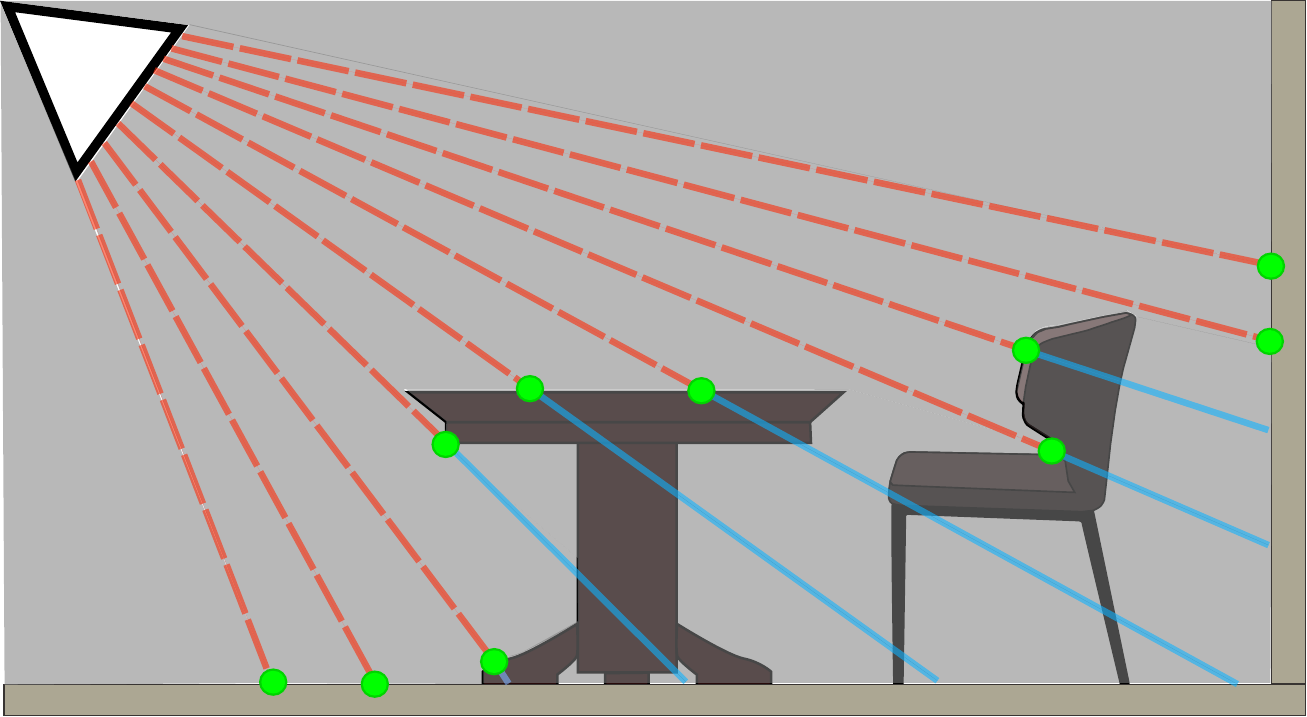}\label{fig:sensors_LIDAR}}
	\caption{\textbf{Scene acquisition.} 
	A camera (RGB, RGB-D, Depth) senses dense volumes but produces noisy depth measurements \protect\subref{fig:sensors_RGBD}, while LiDAR --~more accurate~-- is significantly sparser \protect\subref{fig:sensors_LIDAR}. (Inspired from: \cite{Song2017SemanticSC}).}
	\label{fig:SC_sensors_problem}
\end{figure}

Let $x$ be an incomplete 3D representation of a scene, Semantic Scene Completion (SSC) is the function $f(.)$ inferring a dense semantically labeled scene $\hat{y}$ such that $f(x) = \hat{y}$, best approximates the real 3D scene $y$. 
Most often, $x$ is significantly sparser than $y$ and the complexity lies in the inherent ambiguity, especially where large chunks of data are missing, due to sparse sensing or occlusions (see Fig. \ref{fig:SC_sensors_problem}). Subsequently, the problem cannot be addressed by interpolating data in $x$ and is most often solved by learning priors from $(x, y)$ pairs of sparse input and dense 3D scenes with semantic labels.

The nature of the sparse 3D input $x$ greatly affects the task complexity. While 3D data can be obtained from a wide variety of sensors, RGB-D/stereo cameras or LiDARs are commonly employed. 
The former, for example, provides a dense description of the visible surfaces where missing regions correspond to occluded areas, as shown in~Fig.~\ref{fig:sensors_RGBD}. This reduces the SSC task to estimating semantic completion only in the occluded regions~\cite{Song2017SemanticSC}.
Conversely, LiDAR data provides considerably sparser sensing, with density decreasing afar
and point-wise returns from laser beams cover an infinitesimal portion of the space leading to a high proportion of unknown volume, as shown in Fig.~\ref{fig:sensors_LIDAR}. 

The rest of this section provides the reader with a brief overview of related tasks, covering early works and foundations.

\subsection{Related tasks}\label{section:background}

\begin{table*}
	\scriptsize
	\setlength{\tabcolsep}{0.0035\linewidth}
	\newcommand{\nclass}[1]{\scriptsize{#1}}
	
	\def\mystrut{\rule{0pt}{1.5\normalbaselineskip}}
	\centering
	\rowcolors[]{3}{black!3}{white}
	
	\begin{tabular}{$l | ^c ^c ^c ^c ^c ^c ^c ^c ^c ^c ^c ^c ^c ^c}
		\toprule
		Dataset & Year & Type & Nature & Input$\mapsto$Ground truth & 3D Sensor & \# Classes & \multicolumn{6}{c}{Tasks*} & \#Sequences & \#Frames \\ 
		& & & & & & & SSC & DE & SPE & SS & OC & SNE \\
		\midrule 
		
		\rowstyle{\bfseries}NYUv2~\cite{Silberman2012IndoorSA}\textsuperscript{a} & 2012 & Real-world\tiny{$^{\dag}$} & Indoor & RGB-D $\rightarrow$ Mesh/Voxel & RGB-D & \nclass{894~(11)} & \checkmark & \checkmark & \checkmark & \checkmark & \checkmark &   & 1449 & 795/654 \\
		
		SUN3D~\cite{Xiao2013SUN3DAD} & 2013 & Real-world & Indoor & RGB-D $\rightarrow$ Points & RGB-D & - & \checkmark & \checkmark & \checkmark & \checkmark & \checkmark & & 254 & - \\
		
		\rowstyle{\bfseries}NYUCAD~\cite{Firman2016StructuredPO}\textsuperscript{b} & 2013 & Synthetic & Indoor & RGB-D $\rightarrow$ Mesh/Voxel & RGB-D & \nclass{894~(11)} & \checkmark & \checkmark & \checkmark & \checkmark & \checkmark &   & 1449 & 795/654 \\
		
		SceneNet~\cite{Handa2015SceneNetUR} & 2015 & Synthetic & Indoor & RGB-D $\rightarrow$ Mesh & RGB-D$^{\ddagger}$ & 11 & \checkmark & \checkmark & & \checkmark & \checkmark & & 57 & - \\
		
		SceneNN~\cite{Hua2016SceneNNAS} & 2016 & Real-world & Indoor & RGB-D $\rightarrow$ Mesh & RGB-D & 22 & \checkmark & \checkmark & & \checkmark & \checkmark & & 100 & - \\
		
		\rowstyle{\bfseries}SUNCG~\cite{Song2017SemanticSC} & 2017 & Synthetic & Indoor & Depth $\rightarrow$ Mesh/Voxel & RGB-D$^{\ddagger}$ & \nclass{84~(11)} & \checkmark & & & \checkmark & & & 45622 & 139368/470 \\
		
		Matterport3D~\cite{Chang2017Matterport3DLF} & 2017 & Real-world & Indoor & RGB-D $\rightarrow$ Mesh & 3D Scanner & \nclass{40~(11)} & \checkmark & \checkmark &  & \checkmark &  &  & 707 & 72/18 \\
		
		ScanNet~\cite{Dai2017ScanNetR3} & 2017 & Real-world & Indoor & RGB-D $\rightarrow$ Mesh & RGB-D & \nclass{20~(11)} & \checkmark &  \checkmark &  & \checkmark & \checkmark & \checkmark & 1513 & 1201/312\\
		
		2D-3D-S~\cite{Armeni2017Joint2D} & 2017 & Real-world & Indoor & RGB-D $\rightarrow$ Mesh & 3D Scanner & \nclass{13} & \checkmark & \checkmark &  & \checkmark & \checkmark  & \checkmark &  270 & - \\
		
		SUNCG-RGBD~\cite{Liu2018SeeAT}\textsuperscript{c} & 2018 & Synthetic & Indoor & RGB-D $\rightarrow$ Mesh/Voxel & RGB-D$^{\ddagger}$ & \nclass{84~(11)} & \checkmark & & & \checkmark & & & 45622 & 13011/499 \\
		
		\rowstyle{\bfseries}SemanticKITTI\cite{Behley2019SemanticKITTIAD} & 2019 & Real-world & Outdoor & Points/RGB $\rightarrow$ Points/Voxel & Lidar-64 & \nclass{28~(19)} & \checkmark & \checkmark &  & \checkmark &  & & 22 & 23201/20351\\
		
		SynthCity~\cite{Griffiths2019SynthCityAL} & 2019 & Synthetic & Outdoor & Points $\rightarrow$ Points & Lidar$^{\ddagger}$ & \nclass{9} & \checkmark & & & \checkmark & & & 9 & - \\
		
		CompleteScanNet~\cite{Wu2020SCFusionRI}\textsuperscript{d} & 2020 & Real-world\tiny{$^{\dag}$} & Indoor & RGB-D $\rightarrow$ Mesh/Voxel & RGB-D & \nclass{11} & \checkmark & \checkmark &  & \checkmark & \checkmark & \checkmark & 1513 & 45448/11238 \\
		
		SemanticPOSS~\cite{Pan2020SemanticPOSSAP} & 2020 & Real-world & Outdoor & Points/RGB $\rightarrow$ Points & Lidar-40 & \nclass{14} & \checkmark & \checkmark &  & \checkmark & & & 2988 & - \\

		\toprule

	\end{tabular}\\

	{\scriptsize $^{\ddagger}$ Simulated sensor.}
	{\scriptsize $^{\dag}$ Synthetically augmented.}
	{\scriptsize \textsuperscript{a} Mesh annotations from~\cite{Guo2013SupportSP}.}
	{\scriptsize \textsuperscript{b} Derivates from \textbf{NYUv2}~\cite{Silberman2012IndoorSA} by rendering depth images from mesh annotation.} 
	\\
	{\scriptsize \textsuperscript{c} Derivates from subset of \textbf{SUNCG}~\cite{Song2017SemanticSC} where missing RGB images were rendered.}
	{\scriptsize \textsuperscript{d} Derivates from \textbf{ScanNet}~\cite{Dai2017ScanNetR3} by fitting CAD models to dense mesh.}
	
	{\scriptsize * SSC: Semantic Scene Completion; DE: Depth Estimation; SPE: Sensor Pose Estimation; SS: Semantic Segmentation; } 
	{\scriptsize  OC: Object Classification; SNE: Surface Normal Estimation} \\

	\caption{\textbf{SSC datasets}. We list here datasets readily usable for the SSC task in chronological order.
	Popular datasets are \textbf{bold} and previewed in Fig.~\ref{fig:teaser}. Classes show the total number of semantic classes and when it differs, SSC classes in parenthesis.}
	\label{table:SSC_ready_datasets}
\end{table*}

Semantic scene completion inspires from closely related tasks like shape completion, semantic segmentation, and more recently semantic instance completion.
Thereof, SSC benefits from individual insights of these tasks which we briefly review their literature pointing to the respective surveys.

\paragraph{Completion.}

Completion algorithms initially used interpolation~\cite{Davis2002FillingHI} or energy minimization~\cite{SorkineHornung2004LeastsquaresM, Nealen2006LaplacianMO, Kazhdan2006PoissonSR} techniques to complete small missing regions. Completion works were first devoted to object shape completion, which infers occlusion-free object representation. For instance, some trends exploit symmetry~\cite{Pauly2008DiscoveringSR,  Sipiran2014ApproximateSD, Sung2015DatadrivenSP, de20193d, Thrun2005ShapeFS} and are reviewed in~\cite{Mitra2013SymmetryI3}. Another common approach is to rely on \textit{prior} 3D models to best fit sparse input~\cite{Li2015DatabaseAssistedOR, Pauly2005Examplebased3S,  Shen2012StructureRB, Li2017ShapeCF, Rock2015Completing3O}. In recent years, model-based techniques and new large-scale datasets enlarged the scope of action by enabling inference of complete occluded parts in both scanned objects~\cite{Dai2017ShapeCU, Yuan2018PCNPC, Rezende2016UnsupervisedLO, Yang2019Dense3O, Sharma2016VConvDAEDV, Han2017HighResolutionSC, Smith2017ImprovedAS, Stutz2018Learning3S, Varley2017ShapeCE, Park2019DeepSDFLC} and entire scenes~\cite{Dai2019SGNNSG, Firman2016StructuredPO,Zimmermann2017LearningFA}. 
Moreover, contemporary research shows promising results on challenging multi-object reconstruction from as single RGB image~\cite{Engelmann2020FromPT}.
For further analysis, we refer readers to related surveys~\cite{Yang2019Dense3O, Han2019Imagebased3O, Mitra2013SymmetryI3}.

\paragraph{Semantics.}

Traditional segmentation techniques reviewed in~\cite{Nguyen20133DPC} were based on hand-crafted features, statistical rules, and bottom-up procedures, combined with traditional classifiers.
The advances in deep learning have reshuffled the cards~\cite{Xie2020LinkingPW}.
Initial 3D deep techniques relied on multiviews processed by 2D CNNs~\cite{Su2015MultiviewCN,Boulch2017UnstructuredPC,Boulch2018SnapNet3P} and were quickly replaced by the use of 3D CNNs which operate on voxels~\cite{Maturana2015VoxNetA3,Riegler2017OctNetLD,Tchapmi2017SEGCloudSS,Wang2018AdaptiveOA,Meng2019VVNetVV, Dai20183DMVJ3}, tough suffering from memory and computation shortcomings.
Point-based networks~\cite{Qi2017PointNetDH,Qi2017PointNetDL,Thomas2019KPConvFA,Landrieu2018LargeScalePC,Wang2019DynamicGC,Li2018PointCNNCO} remedied this problem by operating on points and quickly became popular for segmentation, though generative tasks are still a challenge. Therefore, if the point space is pre-defined, SSC can be seen as a point segmentation task \cite{Zhong2020SemanticPC}.
We refer readers to dedicated surveys on semantic segmentation~\cite{Xie2020LinkingPW,Gao2020AreWH}.

\paragraph{Semantic Instance Completion.}

Different from SSC, semantic instance completion is applied only at the objects instance level and cannot complete scene background like walls, floor, etc. 
Most existing methods require instance detection and complete the latter with CAD models \cite{Nan2012ASA,Avetisyan2019Scan2CADLC,Shao2012AnIA} or a completion head \cite{Hou_2020_CVPR, Nie_2021_CVPR, Hou_2020_CVPR,Mller2020SeeingBO,Yuan2018PCNPC}. Notably, \cite{Avetisyan2020SceneCADPO} detects both objects and the scene layout to output a complete lightweight CAD representation. These approaches are commonly based on 3D object detection \cite{Song2016DeepSS,Qi2019DeepHV,Shi2020FromPT} or instance segmentation \cite{Hou20193DSIS3S}. Furthermore, some recent methods tackle multiple object completion from single images either through individual object detection and reconstruction \cite{Izadinia2016IM2C, Gkioxari2019MeshR, Kundu20183DRCNNI3,Yuan2018PCNPC} or joint completion by understanding the scene structure \cite{Nie2020Total3DUnderstandingJL,Popov2020CoReNetC3}. 
\\

Song \textit{et al.}~\cite{Song2017SemanticSC} were the first to 
address
semantic segmentation and scene completion jointly, showing that both tasks can mutually benefit. Since then, many SSC works gather ideas from the above described lines of work and are extensively reviewed in Sec.~\ref{sec:3d_scene_completion}.

\section{Datasets and representations}\label{sec:datasets_scene_rep}

This section presents existing SSC datasets~(Sec.~\ref{sec:datasets}) and commonly used 3D representations for SSC~(Sec.~\ref{section:3D_representations}).

\subsection{Datasets}\label{sec:datasets}

A comprehensive list of all SSC ready datasets is shown in Tab.~\ref{table:SSC_ready_datasets}. We denote as \textit{SSC ready} any dataset containing pairs of sparse/dense data with semantics label.
Note that while 14 datasets meet these criteria, only half has been used for SSC among which the four most popular are  bold in the table and previewed in Fig.~\ref{fig:teaser}.
Overall, there is a prevalence of indoor stationary datasets~\cite{Silberman2012IndoorSA, Song2017SemanticSC, Wu2020SCFusionRI, Chang2017Matterport3DLF, Dai2017ScanNetR3,Armeni2017Joint2D,Hua2016SceneNNAS,Handa2015SceneNetUR} as opposed to outdoors~\cite{Behley2019SemanticKITTIAD,Pan2020SemanticPOSSAP,Griffiths2019SynthCityAL}.

\paragraph{Datasets creation.} Synthetic datasets can easily be obtained by sampling 3D object meshes~\cite{Song2017SemanticSC,Handa2015SceneNetUR,Fu20203DFRONT3F} or simulating sensors in virtual environments~\cite{Dosovitskiy2017CARLAAO, Gaidon2016VirtualWorldsAP, Ros2016TheSD, Griffiths2019SynthCityAL}. Their evident advantage is the virtually free labeling of data, though transferability of synthetically learned features is arguable. 
Real datasets are quite costly to record and annotate, and require a significant processing effort.
Indoor datasets~\cite{Chang2017Matterport3DLF, Silberman2012IndoorSA, Dai2017ScanNetR3,Xiao2013SUN3DAD,Hua2016SceneNNAS,Armeni2017Joint2D,Wu2020SCFusionRI} are commonly captured with {RGB-D} or stereo sensors.
Conversely, outdoor datasets~\cite{Behley2019SemanticKITTIAD,Caesar2019nuScenesAM,Pan2020SemanticPOSSAP} are often recorded with LiDAR and optional camera. They are dominated by autonomous driving applications, recorded in (peri)-urban environment, and must subsequently account for dynamic agents (e.g. moving objects, ego-motion, etc.). 

\begin{figure}[!tbp]
	\centering
	\subfloat[NYUv2~\cite{Silberman2012IndoorSA}]{\includegraphics[height=0.39\linewidth]{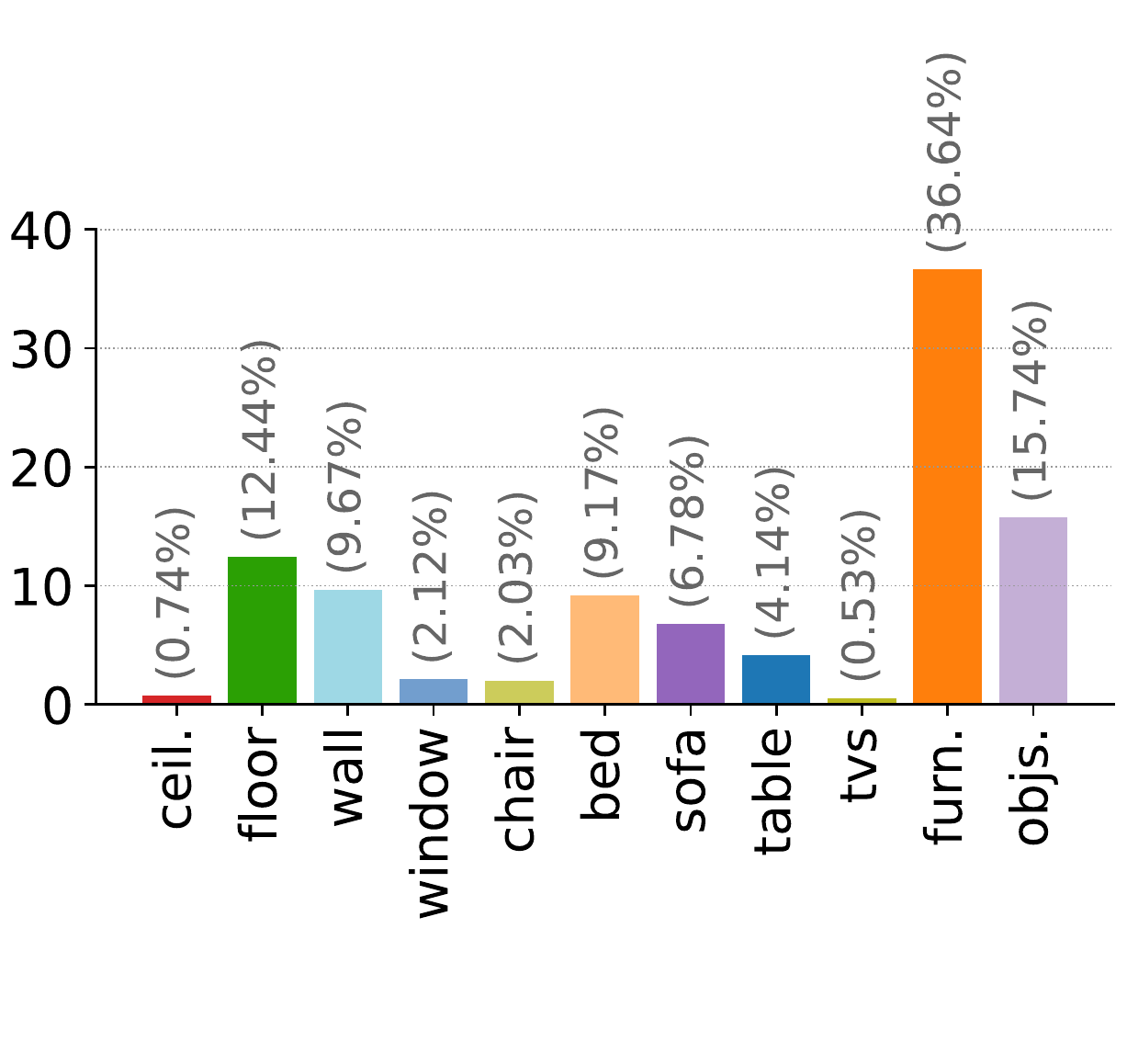}\label{fig:freq_nyuvtwo}}
	\subfloat[SemanticKITTI~\cite{Behley2019SemanticKITTIAD}]{\includegraphics[height=0.39\linewidth]{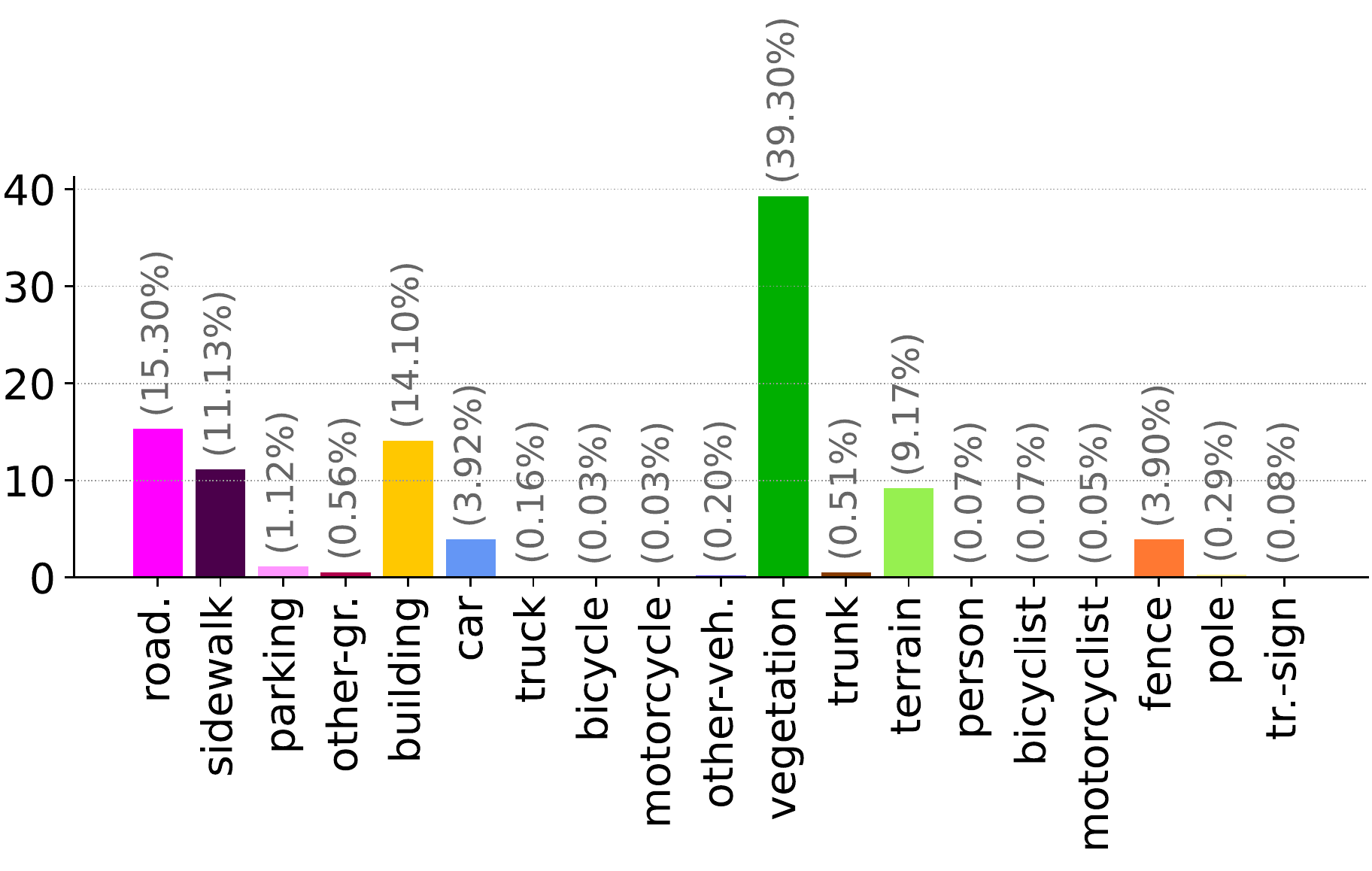}\label{fig:freq_semkit}}
	\caption{\textbf{Semantics distribution.} Class-wise frequencies of the most popular real datasets prove to be highly imbalanced.}
	\label{fig:dset_frequenciess}
\end{figure}

\begin{figure}[!tbp]
	\centering
	\subfloat[Misslabeling~\cite{Silberman2012IndoorSA}]{\includegraphics[width=0.38\linewidth]{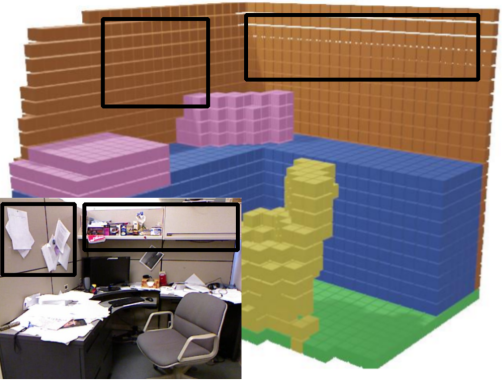}\label{fig:dset_misslabeling}}
	\hspace{0.55cm}
	\subfloat[Sparse sensing~\cite{Behley2019SemanticKITTIAD}]{\includegraphics[width=0.38\linewidth]{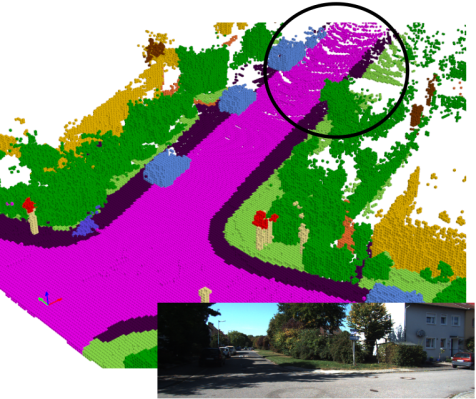}\label{fig:dset_sparsity}}
	\\
	\subfloat[Object motion~\cite{Behley2019SemanticKITTIAD}]{\includegraphics[width=0.38\linewidth]{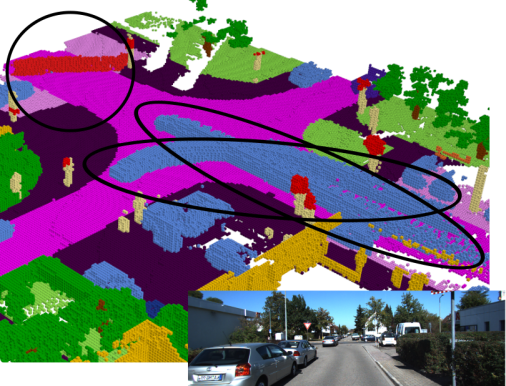}\label{fig:dset_accumulation}}
	\hspace{0.55cm}
	\subfloat[Occlusions~\cite{Behley2019SemanticKITTIAD}]{\includegraphics[width=0.38\linewidth]{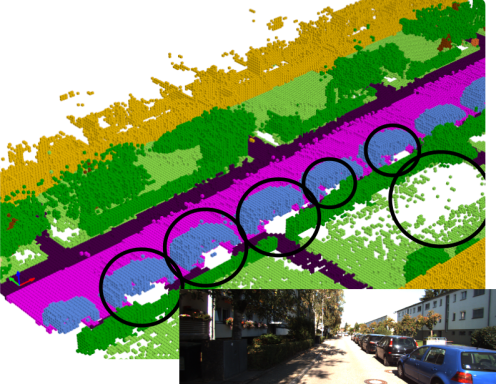}\label{fig:dset_occlusions}}
	\caption{\textbf{Inaccuracies of SSC ground truth.} Providing semantics and geometrics annotation in real-world datasets proves to be complex, and the resulting process is imperfect -- which lead to noisy supervision signal. Ground truth mislabeling is a well known bias \protect\subref{fig:dset_misslabeling}. Sparsity is common in LiDAR-based datasets~\protect\subref{fig:dset_sparsity}. Object motion causes temporal traces~\protect\subref{fig:dset_accumulation}, and occlusions are inevitable in dynamic scenes~\protect\subref{fig:dset_occlusions}.}
	\label{fig:dataset_errors}
\end{figure}

\paragraph{Ground truth generation.} 
An evident challenge is the dense annotation of such datasets. 
Specifically, while indoor stationary scenes can be entirely captured from multiple views or rotating apparatus, 3D outdoor dynamic scenes are virtually impossible to capture \textit{entirely} as it would require ubiquitous scene sensing.\\
Subsequently, ground truth $y$ is obtained from the aggregation and labeling of sparse sequential data $\{y_{0}, y_{1}, ..., y_{T}\}$ over a small time window $T$. 
Multi-frame registration is usually leveraged to that matters, assuming that consecutive frames have an overlapping field of view.
For RGB-D datasets, mostly stationary and indoors, it is commonly achieved from Structure from Motion~(SfM)~\cite{Nair2012HighAT, Saputra2018VisualSA} or visual SLAM~\cite{Saputra2018VisualSA} (vSLAM), which cause holes, missing data, and noisy annotations~\cite{FuentesPacheco2012VisualSL}. 
Such imperfections are commonly reduced by inferring dense complete geometry of objects with local matching of CAD models~\cite{Firman2016StructuredPO,Guo2013SupportSP,Wu2020SCFusionRI}, or post-processing hole filling techniques \cite{Straub2019TheRD}.
In outdoor settings, point cloud registration techniques~\cite{Pomerleau2015ARO, Bellekens2014ASO} enable the co-location of multiple LiDAR measurements into a single reference coordinate system~\cite{Behley2019SemanticKITTIAD}.\\
Interestingly, while frequently referred to as \textit{dense}, the ground truth scenes are often noisy and non-continuous in real datasets, being in fact an \textit{approximation} of the real scene. 
Firstly, regardless of the number of frames used, some portions of the scene remain occluded, see Fig. \ref{fig:dset_occlusions}, especially in dynamic environments. 
Secondly, sensors accuracy and density tend to steadily decrease with the distance, as in Fig.~\ref{fig:dset_sparsity}. 
Thirdly, rigid registration can only cope with viewpoint changes, which leads to dynamic objects (e.g. moving cars) producing \textit{traces}, which impact on the learning priors is still being discussed~\cite{Rist2020SemanticSC,Roldao2020LMSCNetLM, Yan2020SparseSS}, see Fig. \ref{fig:dset_accumulation}. 
Finally, another limitation lies in the sensors, which only sense the geometrical surfaces and not the volumes, turning all solid objects into shells.
To produce semantics labels, the common practice is to observe the aggregated 3D data from multiple virtual viewpoints to minimize the labeling ambiguity. This process is tedious and prone to errors\footnote{Authors of SemanticKITTI report that semantic labeling a hectar of 3D data takes approx. 4.5hr~\cite{Behley2019SemanticKITTIAD}.}, visible in Fig.~\ref{fig:dset_misslabeling}. Ultimately, semantics distribution is highly imbalanced as shown in the two most used indoor/outdoor datasets Fig.~\ref{fig:dset_frequenciess}. 

\paragraph{Indoor datasets.} From Tab.~\ref{table:SSC_ready_datasets}, \textbf{NYUv2}~\cite{Silberman2012IndoorSA} (aka NYU-Kinect) is the most popular indoor \textit{real-world} dataset, composed of mainly office and house room scenery. Despite complete 3D ground truth scene volumes not being originally provided, they have been generated in~\cite{Guo2013SupportSP} by using 3D models and 3D boxes or planes for producing complete synthetically augmented meshes of the scene, generating 1449 pairs of RGB-D images and 3D semantically annotated volumes. Extension of the mesh volumes to 3D grids has been done in \cite{Song2017SemanticSC}. However, mesh annotations are not always perfectly aligned with the original depth images. To solve this, many methods also report results by using depth maps rendered from the annotations directly as in~\cite{Firman2016StructuredPO}. This variant is commonly named as \textbf{NYUCAD} and provides simplified data pairs at the expense of geometric detail loss. Additional indoor real-world datasets as \textbf{Matterport3D}~\cite{Chang2017Matterport3DLF}, \textbf{SceneNN}~\cite{Hua2016SceneNNAS} and \textbf{ScanNet}~\cite{Dai2017ScanNetR3} can be used for completing entire rooms from incomplete 3D meshes~\cite{Dai2019SGNNSG, Dai2018ScanCompleteLS}. The latter has also been synthetically augmented from 3D object models and referred to as \textbf{CompleteScanNet}~\cite{Wu2020SCFusionRI}, providing cleaner annotations. Additionally, smaller \textbf{SUN3D}~\cite{Xiao2013SUN3DAD} provides RGB-D images along with registered semantic point clouds. 
Note that datasets providing 3D meshes or point clouds easily be voxelized as detailed in~\cite{Song2017SemanticSC}. 
Additionally, Stanford \textbf{2D-3D-S}~\cite{Armeni2017Joint2D} provides 360$^{\circ}$ RGB-D images, of interest for completing entire rooms \cite{Dourado2020SemanticSC}. Due to real datasets small sizes, low scene variability, and annotation ambiguities, \textit{synthetic} \textbf{SUNCG}~\cite{Song2017SemanticSC} (aka SUNCG-D) was proposed, being a large scale dataset with pairs of depth images and complete synthetic scene meshes. An extension containing RGB modality was presented in~\cite{Liu2018SeeAT} and known as \textbf{SUNCG-RGBD}. Despite its popularity, the dataset is no longer available due to copyright infringement\footnote{https://futurism.com/tech-suing-facebook-princeton-data}, evidencing a lack of synthetic indoor datasets for SSC, that could be addressed using \textbf{SceneNet}~\cite{Handa2015SceneNetUR}.

\paragraph{Outdoor datasets.} \textbf{SemanticKITTI}~\cite{Behley2019SemanticKITTIAD} is the most popular large-scale dataset and currently the sole providing single sparse and dense semantically annotated point cloud pairs from \textit{real-world} scenes. It derivates from the popular odometry benchmark of the \textbf{KITTI} dataset~\cite{Geiger2013VisionMR}, which provides careful registration and \textit{untwisted} point clouds considering vehicle's ego-motion. Furthermore, voxelized dense scenes were later released as part of an evaluation benchmark with a hidden test set~\footnote{https://competitions.codalab.org/competitions/22037}. The dataset presents big challenges given the high scene variability and the high class imbalance naturally present in outdoor scenes (Fig.~\ref{fig:freq_semkit}).
\textbf{SemanticPOSS}~\cite{Pan2020SemanticPOSSAP} also provides single sparse semantic point clouds and sensor poses in same format as the latter to ease its implementation. Furthermore, \textit{synthetic} \textbf{SynthCity} additionally provides dense semantic point clouds and sensor poses. It has the advantage of excluding dynamic objects, which solves the effect of object motion~(cf.~Fig.~\ref{fig:dset_accumulation}), but not occlusions~(cf.~Fig.~\ref{fig:dset_occlusions}). 

\subsection{Scene representations} \label{section:3D_representations}

\begin{figure}[!tbp]
	\centering
	\subfloat[Point Cloud ]{\includegraphics[height=0.30\columnwidth]{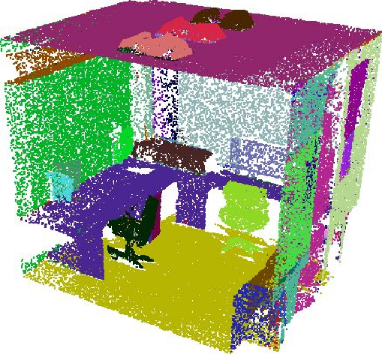}\label{fig:rep_pointcloud}}\hspace{0.1\columnwidth}
	\subfloat[Voxel Grid]{\includegraphics[height=0.30\columnwidth]{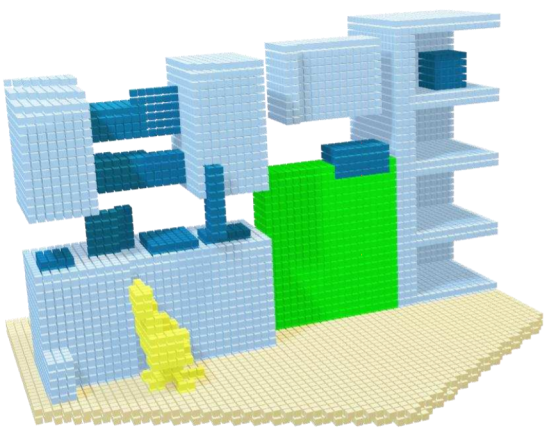}\label{fig:rep_grid}}\\
	\subfloat[Implicit Surface]{\includegraphics[height=0.30\columnwidth]{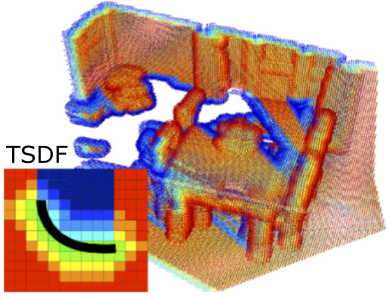}\label{fig:rep_tsdf}}\hspace{0.1\columnwidth}
	\subfloat[Mesh]{\includegraphics[height=0.30\columnwidth]{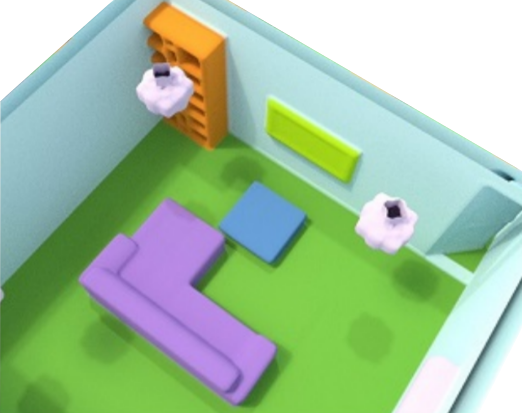}\label{fig:rep_mesh}}
	\caption{\textbf{SSC representations.} Several 3D representations co-exist in the literature. Its choice has a major impact on the method to use, as well as the memory or the computation needs. (Source: \cite{Wang2018SGPNSG,Wang2019ForkNetMV,Song2017SemanticSC,Dai2018ScanCompleteLS})}
	\label{fig:representations}
\end{figure}

We now detail the common 3D representation for SSC output, shown in Fig.~\ref{fig:representations}. While voxel grid or point cloud are the most used, other implicit/explicit representations are of interest for applications like rendering, fluid simulation, etc.

\paragraph{Point Cloud} is a convenient and memory efficient representation, which expresses scene geometry in the 3D continuous world as a set of points lying on its surface. Different from others, point cloud omits definition of the free space and lacks geometrical structure. 
Few works have in fact applied point-based SSC. On objects, PCN~\cite{Yuan2018PCNPC} was the first to address point-based completion followed by~\cite{Wang2020SoftPoolNetSD, Wang2020CascadedRN, Zhang2020DetailPP, Wang2019PatchBasedP3, Wen2020PointCC, Wang2020PointCC, Tchapmi2019TopNetSP, Xie2020GRNetGR, Huang2020PFNetPF}. However, point generation on scenes is more challenging and \cite{Zhong2020SemanticPC,Rist2020SemanticSC} circumvent this need by querying semantic classes at given locations.

\paragraph{Occupancy grid} encodes scene geometry as a 3D grid, in which cells describe semantic occupancy of the space. Opposed to point clouds, grids conveniently define neighborhood with adjacent cells, and thus enable easy application of 3D CNNs,
which facilitates to extend deep learning architectures designed for 2D data into 3D \cite{Song2017SemanticSC, Garbade2019TwoS3, Guo2018ViewVolumeNF, Roldao2020LMSCNetLM, Li2020DepthBS, Zhang2018EfficientSS, Zhang2019CascadedCP, Chen20203DSS, Chen20193DSS, Dai2019SGNNSG, Dai2018ScanCompleteLS, Dourado2019EdgeNetSS, Dourado2020SemanticSC,Cherabier2018LearningPF}. However, the representation suffers from constraining limitations and efficiency drawbacks since it represents both occupied and free regions of the scene, leading to high memory and computation needs. 
Voxels are also commonly used as a support for implicit surface definition which we now describe. 

\paragraph{Implicit surface} encodes geometry as a gradient field expressing the signed distance to the closest surface, known as the Signed Distance Function (SDF). While fields are continuous by nature, for ease of implementation they are commonly encoded in a discrete manner. The value of the gradient field is estimated at specific locations, typically at the voxel centers~\cite{Dai2018ScanCompleteLS, Dai2019SGNNSG}, or at the point locations for point clouds~\cite{Rist2020SemanticSC}. 
Implicit surface may also be used as input~\cite{Zhang2018EfficientSS, Wang2019ForkNetMV, Chen20193DSS, Song2017SemanticSC, Dai2019SGNNSG, Dai2018ScanCompleteLS, Dourado2019EdgeNetSS, Dourado2020SemanticSC, Li2020DepthBS, Chen20203DSS, Zhang2019CascadedCP} to reduce the sparsity of the input data, at the expense of greedy computation. For numerical reason, most works encode in fact a flipped version (cf. f-TSDF, sec. \ref{sec:input_encoding}).
Meshes, explained in detail below, can be obtained from the implicit surface, by using meshification algorithms such as marching cubes~\cite{Lorensen1987MarchingCA}. 

\paragraph{Mesh} enables an explicit surface representation of the scene by a set of polygons. Implementing deep-learning algorithms directly on 3D meshes is challenging and most works obtain the mesh from intermediate implicit voxel-based TSDF representations~\cite{Dai2018ScanCompleteLS, Dai2019SGNNSG} by minimizing distance values within voxels and applying meshification algorithms~\cite{Lorensen1987MarchingCA}. Other alternatives contemplate applying view inpainting as in~\cite{Han2019DeepRL} or using parametric surface elements~\cite{Groueix2018AtlasNetAP}, which are more oriented to object/shape completion.
Furthermore, recent learning-based algorithms such as Deep Marching Cubes~\cite{Liao2018DeepMC} enable to obtain continuous meshes end-to-end from well sampled point clouds, but similarly have not been applied to fill large areas of missing information or scenes.

	\section{Semantic Scene Completion}  \label{sec:3d_scene_completion}

The seminal work of Song \textit{et al.}~\cite{Song2017SemanticSC} first addressed Semantic Scene Completion (SSC) with the observation that semantics and geometry are `tightly intertwined'. 
While there have been great progress lately, the best methods still perform little below 30\% mIoU on the most challenging SemanticKITTI benchmark~\cite{Behley2019SemanticKITTIAD}, advocating that there is a significant margin for improvement. 

Inferring a dense 3D scene from 2D or sparse 3D inputs is in fact an ill-posed problem since the input data are not sufficient to resolve all ambiguities. 
As such, apart from~\cite{Zheng2013BeyondPC, Lin2013HolisticSU,Geiger2015Joint3O,Firman2016StructuredPO}, all existing works rely on deep learning to learn semantics and geometric priors from large scale datasets reviewed in Sec.~\ref{sec:datasets}. 

In the following, we provide a comprehensive survey of the SSC literature. Unlike some historical computer vision tasks, for SSC we found little consensus and a wide variety of choices exists. As such we also focus on the remaining avenues of research to foster future works.

The survey is organized in sections that follow a standard SSC pipeline, with each section analyzing the different line of choices. 
We start in Sec.~\ref{sec:input_encoding} by reviewing the various input encoding strategies, broadly categorized into 2D/3D, and discuss their influence on the global problem framing.
Following this, we study SSC deep architectures in Sec.~\ref{sec:architecture_choices}. While initially, the task was addressed with vanilla 3D CNNs, other architectures have been leveraged such as 2D/3D CNNs, point-based networks, or various hybrid proposals. 
Sec.~\ref{sec:design} presents important design choices, such as contextual aggregation which greatly influences any geometrical task like SSC, or lightweight designs to leverage the burden of 3D networks spanning from compact 3D representations to custom convolutions.
We discuss the training strategies in Sec.~\ref{sec:training}, along with the losses and their benefits. Finally, a grouped evaluation of the metrics, methods performance and network efficiency is in Sec.~\ref{sec:evaluation}.

We provide the reader with a digest overview of the field, chronologically listing methods in Tab.~\ref{tab:methods} -- where columns follow the paper structure. 
Because SSC definition may overlap with some \textit{reconstruction methods} that also address semantics, we draw the inclusion line in that SSC must also complete semantics \textit{and} geometry of \textit{unseen} regions. We in-distinctively report as SSC any method meeting these criteria.
Looking at Tab.~\ref{tab:methods}, it illustrates both the growing interest in this task and the current lack of consensus on input encoding, architectures, etc.

\begin{table*}   
	\scriptsize
	\setlength{\tabcolsep}{0.0035\linewidth}
	\newcommand{\codecaffe}[0]{{\tiny{}Caffe}}
	\newcommand{\codetf}[0]{{\tiny{}TF}}
	\newcommand{\codept}[0]{{\tiny{}PyTorch}}
	\newcommand{\codenon}[0]{{\tiny{}-}}
	\newcommand{\entryremove}[1]{\sout{#1}}
	
	\newcommand{\losses}[1]{\tiny{#1}}
	\newcommand{\fusion}[1]{\tiny{#1}}
	\newcommand{\nofusion}[0]{}
	\newcommand{\refin}[1]{\scriptsize{#1}}
	
	\newcommand{\archvv}[0]{view-volume}
	\newcommand{\archpoint}[0]{point-based}
	\newcommand{\archhyb}[0]{hybrid}
	\newcommand{\archvol}[0]{volume}
	
	\newcommand{\skipcon}[0]{}
	\newcommand{\dilconv}[0]{DC}
	\newcommand{\PDA}[0]{PDA}
	\newcommand{\CntxtAware}[1]{\tiny{#1}}
	\newcommand{\Optim}[1]{\tiny{#1}}
	
	\definecolor{color1}{rgb}{0.0, 0.5, 0.69}
	\definecolor{color2}{rgb}{1.0, 0.22, 0.0}
	\definecolor{color3}{rgb}{0.0, 0.8, 0.6}
	\definecolor{color4}{rgb}{0.93, 0.53, 0.18}
	\newcommand{\SemLoss}[1]{\textcolor{color1}{#1}}
	\newcommand{\RecLoss}[1]{\textcolor{color2}{#1}}
	\newcommand{\CsyLoss}[1]{\textcolor{color3}{#1}}
	\newcommand{\AdvLoss}[1]{}
	
	\def\mystrut{\rule{0pt}{1.5\normalbaselineskip}}
	
	\centering
	\rowcolors[]{4}{black!3}{white}
	\begin{tabular}{rl | c c c  |  c c c  | c |  c | c | c | c | c | c c c c | c | c c c c | c c}
		\toprule
		
		& 
		& 
		\multicolumn{6}{c|}{\stackunder{Input Encoding}{(Sec.~\ref{sec:input_encoding})}} &  
		\multicolumn{1}{c|}{\stackunder{Architecture}{(Sec.~\ref{sec:architecture_choices})}} & 
		\multicolumn{5}{c|}{\stackunder{Design choices}{(Sec.~\ref{sec:design})}} & 
		\multicolumn{5}{c|}{\stackunder{Training}{(Sec. ~\ref{sec:training})}} & 
		\multicolumn{4}{c|}{\stackunder{Evaluation}{(Sec. ~\ref{sec:evaluation})}} & 
		\multicolumn{2}{c}{Open source}\\
		
		& & \multicolumn{3}{c|}{2D} & \multicolumn{3}{c|}{3D} & & & & & & & & & & & & & & & & & \\
		
		&
		& 
		\rotatebox{90}{RGB} & 
		\rotatebox{90}{Depth/Range/HHA} & 
		\rotatebox{90}{Other {\tiny{}(seg., normals, etc.)}} & 
		\rotatebox{90}{Occ. Grid} & 
		\rotatebox{90}{TSDF} & 
		\rotatebox{90}{Point Cloud} &  
		\rotatebox{90}{Network type} &
		\rotatebox{90}{Contextual Awareness} &  
		\rotatebox{90}{Position Awareness} & 
		\rotatebox{90}{Fusion Strategies} & 
		\rotatebox{90}{Lightweight Design} & 
		\rotatebox{90}{Refinement} & 
		\rotatebox{90}{End-to-end} & 
		\rotatebox{90}{Coarse-to-fine} & 
		\rotatebox{90}{Multi-scale} & 
		\rotatebox{90}{Adversarial} & 
		\rotatebox{90}{Losses} & 
		\rotatebox{90}{NYU\textsuperscript{b}~\cite{Silberman2012IndoorSA,Firman2016StructuredPO}} & 
		\rotatebox{90}{SUNCG\textsuperscript{c}~\cite{Song2017SemanticSC,Liu2018SeeAT}} & 
		\rotatebox{90}{SemanticKITTI~\cite{Behley2019SemanticKITTIAD}} & 
		\rotatebox{90}{Other~\cite{Chang2017Matterport3DLF,Dai2017ScanNetR3,Wu2020SCFusionRI,Liu2018SeeAT,Armeni2017Joint2D}} & 
		\rotatebox{90}{Framework} & 
		\rotatebox{90}{Weights}\\
		\midrule
		
		2017 & SSCNet~\cite{Song2017SemanticSC}\textsuperscript{a} & & & & & \checkmark & & \archvol & \CntxtAware{\skipcon~\dilconv} & & \nofusion & \Optim{} & \refin{} & \checkmark & & & & \losses{\SemLoss{CE}} &  \checkmark & \checkmark & \checkmark & & \codecaffe & \checkmark \\

		2018\mystrut & Guedes et al.~\cite{Guedes2018SemanticSC} & \checkmark & & & & \checkmark & & \archvol & \CntxtAware{\dilconv} & \Optim{} & \fusion{E} & & \refin{} & \checkmark & & & & \losses{\SemLoss{CE}} & \checkmark & & & & \codenon & \\
        & ScanComplete~\cite{Dai2018ScanCompleteLS} & & & & & \checkmark & & \archvol & \CntxtAware{\skipcon} & & \nofusion & \Optim{GrpConv} & \refin{} & & \checkmark & & & \losses{\RecLoss{$\ell_{1}$}~\SemLoss{CE}} &  & \checkmark & & \checkmark & \codetf & \checkmark \\
		& VVNet~\cite{Guo2018ViewVolumeNF} & & \checkmark & \checkmark & & & &  \archvv & \CntxtAware{\skipcon~\dilconv} & & \fusion{E} & \Optim{} & \refin{} & \checkmark & & & & \losses{\SemLoss{CE}} &  \checkmark & \checkmark & & & \codetf & \checkmark \\
		
		& Cherabier et al.~\cite{Cherabier2018LearningPF} 
		& \checkmark & & 
		& & \checkmark & 	
		& \archvol			
		& \CntxtAware{\PDA}	
		& 					
		& \fusion{E}		
		& \Optim{MSO}		
		& 					
		& \checkmark & \checkmark & \checkmark &  
		& \losses{\SemLoss{CE}} 
		& & & & \checkmark 	
		& \codenon & 		
		\\
		
		& VD-CRF~\cite{zhang2018semantic} & & & & & \checkmark & & \archvol & \CntxtAware{\dilconv} & &  \nofusion & \Optim{} &  \refin{\checkmark} & \checkmark & & & & \losses{\SemLoss{CE}} & \checkmark & \checkmark & & & \codenon & \\
		& ESSCNet~\cite{Zhang2018EfficientSS} & & & & & \checkmark & & \archvol & \CntxtAware{\skipcon} & & \nofusion & \Optim{GrpConv Sparse} & \refin{} & & & \checkmark & & \losses{\SemLoss{CE}} &  \checkmark & \checkmark & & & \codept & \checkmark \\
		& ASSCNet~\cite{Wang2018AdversarialSS} & & \checkmark & & & & & \archvv & \CntxtAware{Mscale. CAT} & & \nofusion & \Optim{} & \refin{} & & & & \checkmark & \losses{\SemLoss{CE}~\AdvLoss{ADV}} &  \checkmark & \checkmark & & & \codetf & \\
		& SATNet~\cite{Liu2018SeeAT} & \checkmark & \checkmark & & & &  & \archvv & \CntxtAware{ASPP} & & \fusion{M/L} & \Optim{} & \refin{} & \checkmark & & & & \losses{\SemLoss{CE}} &  \checkmark & \checkmark & & & \codept & \checkmark \\

		2019\mystrut & DDRNet~\cite{Li2019RGBDBD} & \checkmark & \checkmark & & & & & \archvv & \CntxtAware{LW-ASPP} & & \fusion{M} & \Optim{DDR}  & \refin{} & \checkmark & & & & \losses{\SemLoss{CE}} &  \checkmark & \checkmark & & & \codept & \checkmark \\
		& TS3D~\cite{Garbade2019TwoS3}\textsuperscript{a} & \checkmark & & & \checkmark & & & \archhyb & \CntxtAware{\skipcon~\dilconv} & & \fusion{E} & \Optim{} & \refin{} & \checkmark & & & & \losses{\SemLoss{CE}} &  \checkmark & \checkmark & \checkmark & & \codenon & \\ 
		& EdgeNet~\cite{Dourado2019EdgeNetSS} & \checkmark & & & & \checkmark & & \archvol & \CntxtAware{\dilconv} & \checkmark & \fusion{M} & \Optim{} & \refin{} & \checkmark & & & & \losses{\SemLoss{CE}} &  \checkmark & \checkmark & & & \codenon{} & \\
		& SSC-GAN~\cite{Chen20193DSS} & & & & & \checkmark & & \archvol & \CntxtAware{\dilconv} & & \nofusion & \Optim{} & \refin{} & & & & \checkmark & \losses{\textcolor{red}{\RecLoss{BCE}~\SemLoss{CE}~\AdvLoss{ADV}}} &  \checkmark & \checkmark & & & \codenon{} & \\
		& TS3D+RNet~\cite{Behley2019SemanticKITTIAD} & & \checkmark & & \checkmark & & & \archhyb & \CntxtAware{\skipcon~\dilconv} & & \fusion{E} & \Optim{} & \refin{} & \checkmark & & & & \losses{\SemLoss{CE}} &  & & \checkmark & & \codenon{} & \\ 
		& TS3D+RNet+SATNet~\cite{Behley2019SemanticKITTIAD} & & \checkmark & & \checkmark & & & \archhyb & \CntxtAware{\skipcon~\dilconv} & & \fusion{E} & \Optim{} & \refin{} & \checkmark & & & & \losses{\SemLoss{CE}} &  & & \checkmark & & \codenon{} & \\ 
		& ForkNet~\cite{Wang2019ForkNetMV} & & & & & \checkmark & & \archvol & \CntxtAware{\dilconv} & & \nofusion & & \refin{} & & & & \checkmark & \losses{\RecLoss{BCE}~\SemLoss{CE}~\AdvLoss{ADV}} &  \checkmark & \checkmark & & & \codetf & \checkmark \\
        & CCPNet~\cite{Zhang2019CascadedCP} & & & & & \checkmark & & \archvol & \CntxtAware{CCP~\dilconv} & & \nofusion & \Optim{GrpConv} & \refin{\checkmark} & \checkmark & & & & \losses{\SemLoss{CE}} &  \checkmark & \checkmark & & & \codenon{} & \\
		& AM$^{2}$FNet~\cite{Chen2019AM2FNetAM}  & \checkmark & & & & \checkmark & & \archhyb & \CntxtAware{\skipcon~\dilconv} & \checkmark & \fusion{M} & \Optim{} & \refin{\checkmark} & \checkmark & & & & \losses{\RecLoss{BCE}~\SemLoss{CE}} & \checkmark & & & & \codenon & \\

		2020\mystrut & GRFNet~\cite{Liu20203DGR} & \checkmark & \checkmark & & & & & \archvv & \CntxtAware{LW-ASPP~\dilconv} & & \fusion{M} & \Optim{DDR} & \refin{} & \checkmark & & & & \losses{\SemLoss{CE}} &  \checkmark & \checkmark & & & \codenon & \\
		& Dourado et al.~\cite{Dourado2020SemanticSC} & \checkmark & & & & \checkmark & & \archvol & \CntxtAware{\dilconv} & \checkmark & \fusion{E} & & \checkmark & \checkmark & & & & \losses{\SemLoss{CE}} & \checkmark &  & & & \codenon{} & \\
		& AMFNet~\cite{Li2020AttentionbasedMF} & \checkmark & \checkmark & & & & & \archvv & \CntxtAware{LW-ASPP} & & \fusion{L} & \Optim{RAB} & \refin{} & \checkmark & & & & \losses{\SemLoss{CE}} &  \checkmark & \checkmark & & & \codenon{} & \\
		& PALNet~\cite{Li2020DepthBS} & & \checkmark & & & \checkmark & & \archhyb & \CntxtAware{FAM~\dilconv} &  \checkmark & \fusion{M} & \Optim{} & \refin{} & \checkmark & & & & \losses{\SemLoss{PA}} &  \checkmark & \checkmark & & & \codept & \checkmark \\
		& 3DSketch~\cite{Chen20203DSS} & \checkmark & & & & \checkmark & & \archhyb & \CntxtAware{\skipcon~\dilconv} & \checkmark & \fusion{M} & \Optim{DDR} & \refin{} & \checkmark &  & & \checkmark & \losses{\RecLoss{BCE}~\SemLoss{CE}~\CsyLoss{CCY}} &  \checkmark & \checkmark & & & \codept & \checkmark \\
        & AIC-Net~\cite{Li2020AnisotropicCN} & \checkmark & \checkmark & & & & & \archvv & \CntxtAware{FAM~AIC} & & \fusion{M} & \Optim{Anisotropic} & \refin{} & \checkmark & & & & \losses{\SemLoss{CE}} &  \checkmark & \checkmark & & & \codept & \checkmark \\
        & Wang et al.~\cite{Wang2020DeepOC} & & & & \checkmark & & & \archvol & \CntxtAware{\skipcon} & & \nofusion & \Optim{Octree-based} & \refin{} & \checkmark & & & & \losses{\RecLoss{BCE}~\SemLoss{CE}} &  & \checkmark & & & \codenon{} & \\
        & L3DSR-Oct~\cite{Wang2019Learning3S} & & & & & \checkmark & & \archvol & \CntxtAware{} & &  \nofusion & \Optim{Octree-based} & & & \checkmark & & & \losses{\RecLoss{BCE}~\SemLoss{CE}} & & \checkmark & \checkmark & \checkmark & \codenon & \\ 
        & IPF-SPCNet~\cite{Zhong2020SemanticPC} & \checkmark & & & & & \checkmark & \archhyb & \CntxtAware{\skipcon} & & \fusion{E} & \Optim{} & \refin{} & \checkmark & & & & \losses{\SemLoss{CE}} &  \checkmark & & & & \codenon{} & \\
        & Chen et al.~\cite{Chen2020RealTimeSS} & & & & & \checkmark & & \archvol & \CntxtAware{GA Module} & & \nofusion & \Optim{} & \refin{} & \checkmark & & & & \losses{\RecLoss{BCE}~\SemLoss{CE}} &  \checkmark & \checkmark & & & \codenon{} & \\
		& LMSCNet~\cite{Roldao2020LMSCNetLM} & & & & \checkmark & & & \archvv & \CntxtAware{\skipcon~MSFA} & & \nofusion & \Optim{2D} & \refin{} & \checkmark & & \checkmark & & \losses{\SemLoss{CE}} &  \checkmark & & \checkmark & & \codept & \checkmark \\
		& SCFusion~\cite{Wu2020SCFusionRI} & & & & \checkmark & & & \archvol & \CntxtAware{\dilconv} & & \nofusion & \Optim{} & \refin{\checkmark} &  & & & \checkmark & \losses{\SemLoss{CE}~\AdvLoss{ADV}}  & & & \checkmark & \checkmark & \codenon{} & \\
		& S3CNet~\cite{Cheng2020S3CNet} & & \checkmark & \checkmark & & \checkmark & & \archhyb & \CntxtAware{\skipcon} &  & \fusion{L} & \Optim{Sparse} & \refin{\checkmark} & \checkmark & & & & \losses{\RecLoss{BCE}~\SemLoss{CE~PA}} &  & & \checkmark & & \codenon{} &  \\
		& JS3C-Net~\cite{Yan2020SparseSS} & & & & \checkmark & & & \archvol & \CntxtAware{\skipcon} &  & \nofusion & \Optim{Sparse} & \refin{} & \checkmark & & & & \losses{\SemLoss{CE}} &  & & \checkmark & & \codept & \checkmark \\
		& Local-DIFs~\cite{Rist2020SemanticSC} & & & & & & \checkmark & \archpoint & \CntxtAware{\skipcon} & \Optim{} & \nofusion &  & \refin{} & \checkmark & & & & \losses{\RecLoss{BCE}~\SemLoss{CE}~\CsyLoss{SCY}} &  & & \checkmark & & \codenon{} & \\
		
		2021 & SISNet~\cite{Cai2021SemanticSC} & \checkmark & & & & \checkmark & & \archhyb & & & \fusion{M} & \Optim{DDR} & &  \checkmark & & & &  \losses{\SemLoss{CE}} & \checkmark & \checkmark & & \checkmark & \codept & \\
		\bottomrule
	\end{tabular}\\
	{\scriptsize \textsuperscript{a} These original works were significantly extended in~\cite{Guo2018ViewVolumeNF}, \cite{Behley2019SemanticKITTIAD}, or ~\cite{Roldao2020LMSCNetLM}.} \\
	{\scriptsize \textsuperscript{b} Includes both NYUv2\cite{Silberman2012IndoorSA}, NYUCAD\cite{Firman2016StructuredPO}.}
	{\scriptsize \textsuperscript{c} Includes both SUNCG\cite{Song2017SemanticSC}, SUNCG-RGBD\cite{Liu2018SeeAT}.}
	{\scriptsize \textsuperscript{d} Includes both ScanNet\cite{Dai2017ScanNetR3}, CompleteScanNet\cite{Wu2020SCFusionRI}.}\\
	{\scriptsize Contextual Awareness - DC, Dilated Convolutions. (LW)-ASPP, (Lightweight) Atrous Spatial Pyramid Pooling. CCP, Cascaded Context Pyramid. FAM, Feature Aggregation Module. AIC, Anisotropic Convolutional Module. GA, Global Aggregation. MSFA, Multi-scale Feature Aggregation. PDA, Primal-Dual Algorithm.}\\
	{\scriptsize Fusion Strategies - E, Early. M, Middle. L, Late.}\\
	{\scriptsize Lightweight Design - GrpConv, Group Convolution. DDR, Dimensional Decomposition Residual Block. RAB, Residual Attention Block. MSO, MultiScale Optimization.}\\
	{\scriptsize Losses - \RecLoss{Geometric:} BCE, Binary Cross Entropy. $\ell_{1}$, L1 norm. \SemLoss{Semantic}: CE, Cross Entropy. PA, Position Awareness. \CsyLoss{Consistency:} CCY, Completion Consistency. SCY, Spatial Semantics Consistency.
	}
	\caption{\textbf{Semantic Scene Completion (SSC) methods.}
	} 
	\label{tab:methods}
\end{table*}
\begin{figure}
	\centering
	\subfloat[Surface]{\includegraphics[width=0.40\linewidth]{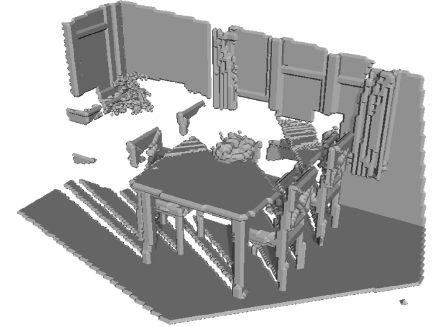}\label{fig:tsdf_vis_surface}}
	\hspace{0.1\columnwidth}
	\subfloat[Projective TSDF]{\includegraphics[width=0.46\linewidth]{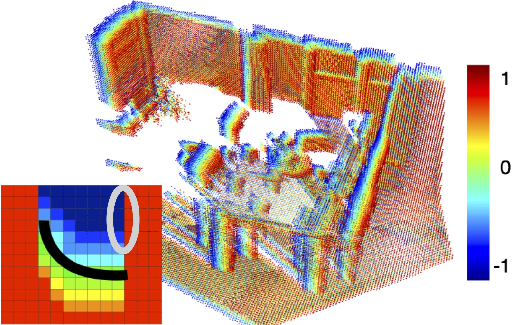}\label{fig:tsdf_ptsdf}}\\
	\subfloat[TSDF]{\includegraphics[width=0.40\linewidth]{Figures/TSDF.png}\label{fig:tsdf_tsdf}}
	\hspace{0.1\columnwidth}
	\subfloat[Flipped TSDF]{\includegraphics[width=0.46\linewidth]{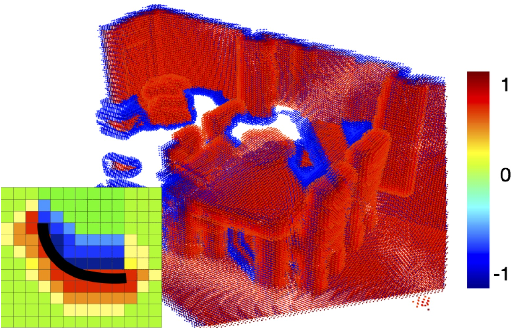}\label{fig:tsdf_ftsdf}}
	\caption{\textbf{TSDF variants}. Projective TSDF \protect\subref{fig:tsdf_ptsdf} is fast to obtain but view-dependent. TSDF \protect\subref{fig:tsdf_tsdf} ~\cite{Dai2018ScanCompleteLS,Chen20193DSS,Wang2019ForkNetMV,Wang2020DeepOC,Chen2020RealTimeSS,Chen20203DSS} solves the view dependency but gradient is stronger at farther areas from the surface, being inadequate for learning-based methods. In contrast, f-TSDF \protect\subref{fig:tsdf_ftsdf}~\cite{Song2017SemanticSC,Zhang2018EfficientSS,Dourado2019EdgeNetSS,Zhang2019CascadedCP,Cheng2020S3CNet,Dourado2020SemanticSC,Chen2019AM2FNetAM} has strongest gradient near the surface.  (Source:~\cite{Song2017SemanticSC}).}
	\label{fig:tsdf_examples}
\end{figure}

\subsection{Input Encoding} \label{sec:input_encoding}
Given the 3D nature of the task, there is an evident benefit of using 3D inputs as it already withholds geometrical insights. As such, it is easier to leverage sparse 3D surface as input in the form of occupancy grid, distance fields, or point clouds. Another line of work uses RGB data in conjunction with depth data since they are spatially aligned and easily handled by 2D CNNs.

\paragraph{3D grid-based.}
In most works, 3D occupancy grid (aka Voxel Occupancy) is used~\cite{Roldao2020LMSCNetLM, Garbade2019TwoS3, Yan2020SparseSS,Wu2020SCFusionRI}, encoding each cell as either \textit{free} or \textit{occupied}. Such representation is conveniently encoded as binary grids, easily compressed (cf. Sec.~\ref{sec:optimization_strategies}), but provides little input signal for the network. 
An alternative richer encoding is the use of TSDF (Fig.~\ref{fig:tsdf_tsdf}), where the signed distance $d$ to the closest surface is computed at given 3D locations (usually, voxel centers), as in~\cite{Dai2019SGNNSG,Dai2018ScanCompleteLS,Chen20203DSS,Cherabier2018LearningPF}. 
Instead of only providing input signal at the measurement locations like occupancy grids or point cloud, TSDF provides a richer supervision signal for the network. The sign for example provides guidance on which part of the scene is occluded in the input. The greedy 3D computation can be avoided using projective TSDF (p-TSDF, cf. Fig.~\ref{fig:tsdf_ptsdf}) which only computes the distance along the sensing path~\cite{Newcombe2011KinectFusionRD}, but with the major drawback of being view-dependent. 
Highlighted by Song \textit{et al.}~\cite{Song2017SemanticSC}, another limitation of TSDF or {p-TSDF} lies in the strong gradients being in the free space area rather than close to the surface where the networks need guidance. This is noticeable in Fig.~\ref{fig:tsdf_ptsdf}, \ref{fig:tsdf_tsdf} since the red/blue gradients are afar from the surface.
To move strong gradients closer to the surface, they introduced flipped TSDF ({f-TSDF}, Fig.~\ref{fig:tsdf_ftsdf}) such that $\text{f-TSDF} = sign(\text{TSDF})(d_\text{max} - d)$ with $d_\text{max}$ the occlusion boundary, showing improvement in both network guidance and performance~\cite{Song2017SemanticSC}. However, the field yet lacks thorough study on the benefit of TSDF encoding. While f-TSDF is still commonly used in recent SSC~\cite{Zhang2018EfficientSS,zhang2018semantic,Dourado2019EdgeNetSS,Dourado2020SemanticSC,Zhang2019CascadedCP,Li2020DepthBS,Chen2019AM2FNetAM} or SC~\cite{denninger20203d}, other approaches stick with original TSDF~\cite{Dai2018ScanCompleteLS,zhang2018semantic,Chen20203DSS,Wang2019ForkNetMV}. Furthermore, the benefit of f-TSDF over TSDF is questioned in~\cite{zhang2018semantic,Garbade2019TwoS3} with experiments showing it may slightly harm the performance or bring negligible improvement.
Except for the lighter p-TSDF, all TSDF-variants require high computation times and hinder real time implementations.

\paragraph{3D point cloud.} Despite the benefit of such 
representation, only three recent SSC works~\cite{Zhong2020SemanticPC,Rist2020SemanticSC,Rist2020SCSSnetLS} use point cloud as input encoding. In~\cite{Zhong2020SemanticPC}, RGB is used in conjunction to augment point data with RGB features, whereas~\cite{Rist2020SemanticSC, Rist2020SCSSnetLS} reproject point features in the top-view space. 
Despite the few SSC methods using points input, it is commonly used for object completion~\cite{Yuan2018PCNPC,Huang2019DeepNN,Xie2020GRNetGR,Wang2020CascadedRN}. 

\paragraph{2D representations.} Alternatively, depth maps or range images (i.e. 2D polar-encoded LiDAR data) are common 2D representations storing geometrical information, therefore suitable candidates for the SSC task. Indeed many works~\cite{Behley2019SemanticKITTIAD,Cheng2020S3CNet,Guo2018ViewVolumeNF,Li2020AnisotropicCN,Li2020DepthBS,Li2020AttentionbasedMF,Li2019RGBDBD,Lin2013HolisticSU,Liu2018SeeAT,Wang2018AdversarialSS} used either representation alone or in conjunction with other modalities. Opposite to point cloud but similarly to grid representation, such encoding enables meaningful connexity of the data and cheap processing with 2D CNNs. 
Some works propose to transform depth into an HHA image~\cite{Gupta2014LearningRF}, 
which keeps more effective information when compared to the single channel depth encoding~\cite{Guo2018ViewVolumeNF, Liu2018SeeAT}. 
Additional non-geometrical modalities such as RGB or LiDAR refraction intensity provide auxiliary signals specifically useful to infer semantic labels~\cite{Behley2019SemanticKITTIAD,Garbade2019TwoS3,Liu2018SeeAT,Li2019RGBDBD,Liu20203DGR,Li2020AttentionbasedMF,Chen20203DSS,Zhong2020SemanticPC,Cherabier2018LearningPF}. In practice, some works have shown that good enough semantic labels can be inferred directly from depth/range images~\cite{Behley2019SemanticKITTIAD,Yan2020SparseSS,Cherabier2018LearningPF} to guide SSC.

Interestingly, the vast majority of the literature only accounts for surface information while ignoring any free space data provided by the sensors (see Fig.~\ref{fig:SC_sensors_problem}). While \textit{free space} labels might be noisy, we believe it provides an additional signal for the network which was found beneficial in~\cite{Rist2020SemanticSC,Wu2020SCFusionRI}. Conversely, Roldao et al.~\cite{Roldao2020LMSCNetLM} relate that encoding \textit{unknown} information brought little improvement. 

\subsection{Architecture choices}\label{sec:architecture_choices}

\begin{figure}
  \centering

  \subfloat[View-Volume Nets., 3D-backbone:~\cite{Guo2018ViewVolumeNF,Liu2018SeeAT,Li2019RGBDBD,Liu20203DGR,Li2020AttentionbasedMF,Li2020AnisotropicCN}, 2D-backbone.:~\cite{Roldao2020LMSCNetLM}]{
  \renewcommand{\arraystretch}{0.5}
  \begin{tabular}{c | c}		
	\rotatebox{90}{\scriptsize ~~~~~~~3D-backbone} & \includegraphics[width=0.87\linewidth]{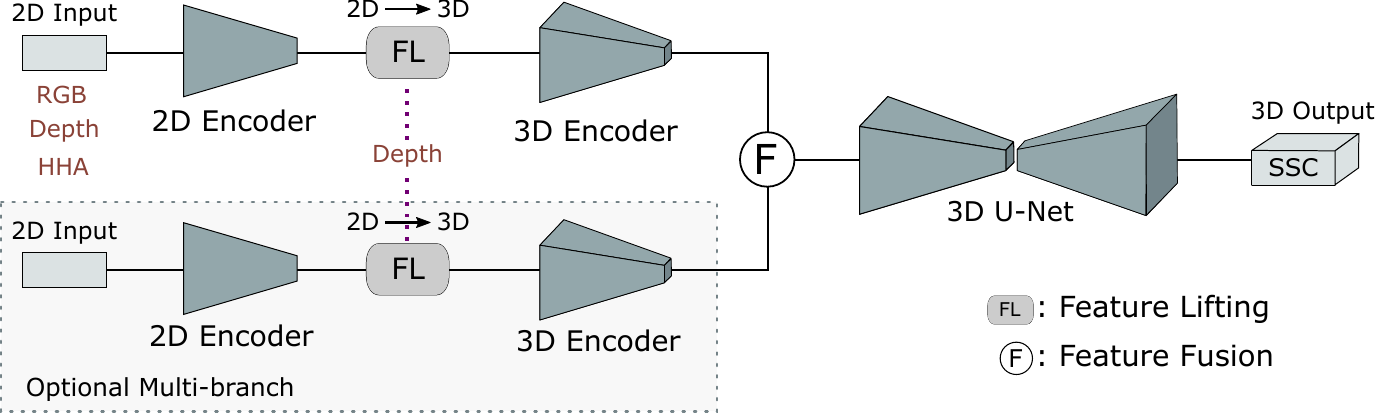}\\
	\multicolumn{2}{c}{}\\
	\rotatebox{90}{\scriptsize 2D-backbone} & \includegraphics[width=0.87\linewidth]{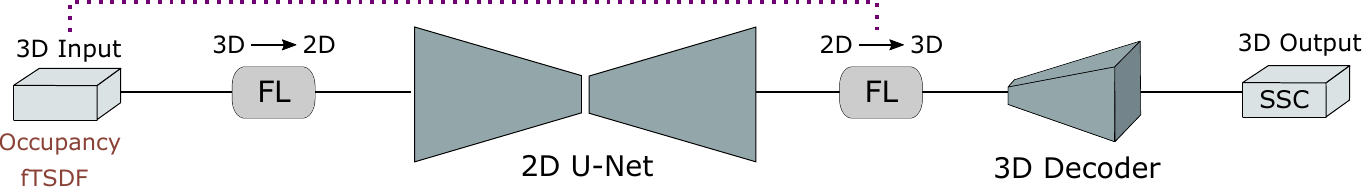} \\
	\label{fig:arch_view_volume}
  \end{tabular}}

  \subfloat[Volume Nets.~\cite{Song2017SemanticSC,Guedes2018SemanticSC,zhang2018semantic,Zhang2018EfficientSS,Dourado2019EdgeNetSS,Chen20193DSS,Wang2019ForkNetMV,Zhang2019CascadedCP,Wang2020DeepOC,Chen2020RealTimeSS,Yan2020SparseSS,Dourado2020SemanticSC,Wu2020SCFusionRI,Cherabier2018LearningPF,Dai2018ScanCompleteLS,Wang2019Learning3S}]{~~~~~\includegraphics[width=0.48\linewidth]{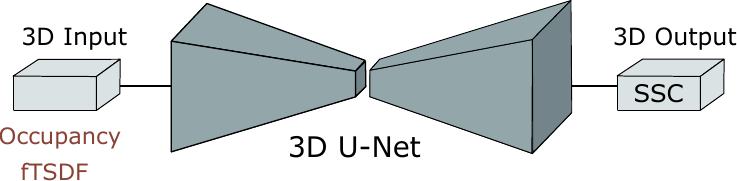}\label{fig:arch_volume}}~~~~~
  \hfill
  \subfloat[Point-based Nets.~\cite{Zhong2020SemanticPC}]{~\includegraphics[width=0.39\linewidth]{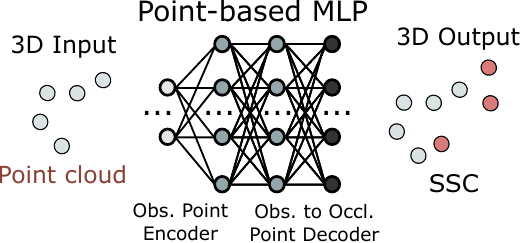}\label{fig:arch_point_based}}~
  \\
  
  \subfloat[Hybrid Nets., point-2D:~\cite{Rist2020SemanticSC, Rist2020SCSSnetLS}, parallel-2D-3D:~\cite{Cheng2020S3CNet}, point-augmented:~\cite{Zhong2020SemanticPC}, mix-2D-3D:~\cite{Garbade2019TwoS3,Behley2019SemanticKITTIAD,Li2020DepthBS,Chen20203DSS,Cai2021SemanticSC}]{
  \renewcommand{\arraystretch}{0.5}
  \begin{tabular}{c | c}
  	\rotatebox{90}{\scriptsize ~~point-2D} & \includegraphics[width=0.87\linewidth]{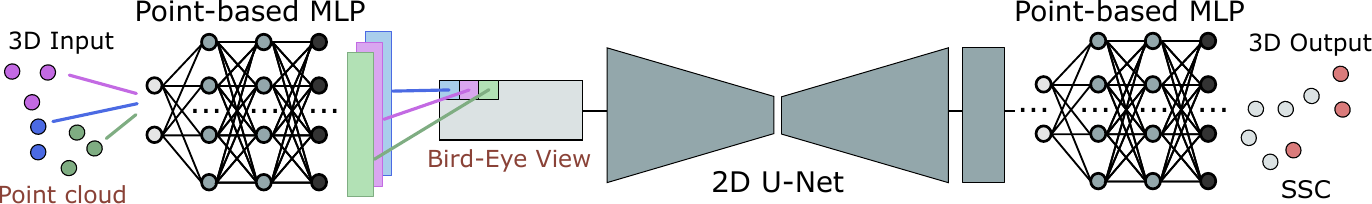}\\
	\multicolumn{2}{c}{}\\
	\rotatebox{90}{\scriptsize par.-2D-3D} & \includegraphics[width=0.87\linewidth]{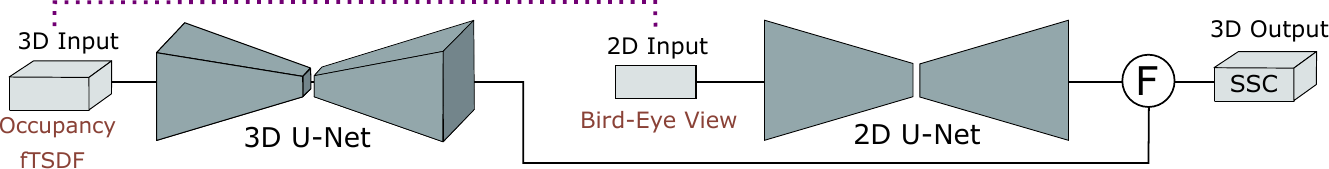}\\
	\multicolumn{2}{c}{}\\
	\rotatebox{90}{\scriptsize ~~~point-aug.} & \includegraphics[width=0.87\linewidth]{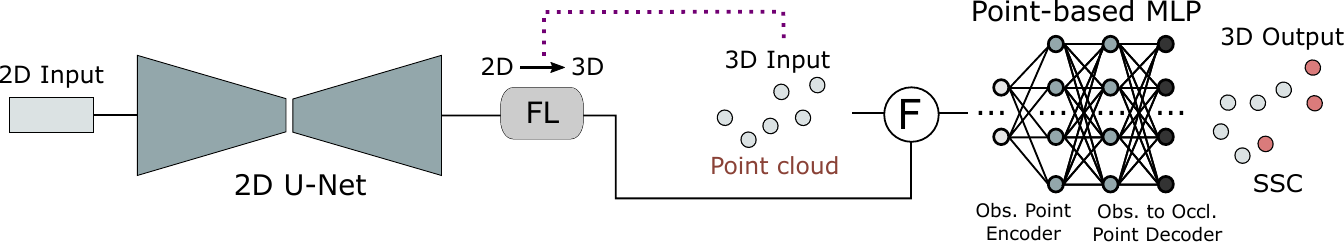}\\
	\multicolumn{2}{c}{}\\
	\rotatebox{90}{\scriptsize mix-2D-3D} & \includegraphics[width=0.87\linewidth]{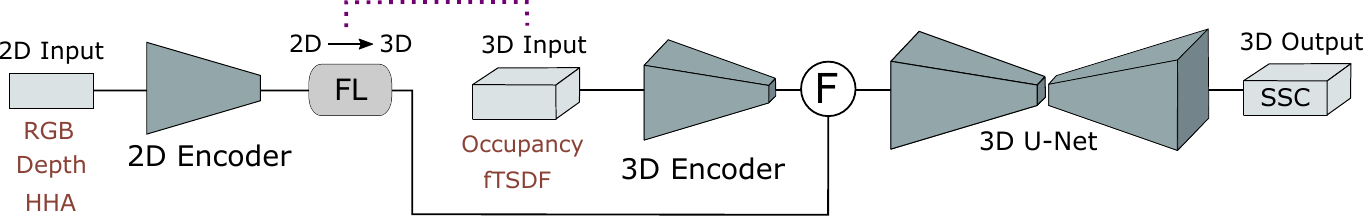}\\
	\label{fig:arch_hybrid}
\end{tabular}}

  \caption{\textbf{Architectures for SSC.} For compactness, we do not display all connections but rather focus on the global architectures and information exchanges between the different networks. \raisebox{.5pt}{\textcircled{\raisebox{-.15pt} {\scriptsize{}F}}} stands for any type of fusion.}
  
  \label{fig:architecture_schemes}

\end{figure}

Directly linked to the choice of input encoding, we broadly categorize architectural choices into 4 groups. In detail: \textit{Volume networks} leveraging 3D CNNs to convolve volumetric grid representations, \textit{Point-based networks} which compute features on points locations, \textit{View-Volume networks} that learn the 2D-3D mapping of data with 2D and 3D CNNs, and \textit{Hybrid networks} that use various networks to combine modalities of different dimension.
All architectures output $N\times{}C$ data ($N$ the spatial dimensions and $C$ the number of semantic classes) where the last dimension is the probability of either semantic class at the given location. In most works, the final prediction is the softmax output with the class probabilities.
We refer to Fig.~\ref{fig:architecture_schemes} for a visual illustration of either architecture type.

\paragraph{Volume networks.} As they are convenient for processing grid data 3D CNNs~(Fig.~\ref{fig:arch_volume}) are the most popular for SSC~\cite{Song2017SemanticSC,Guedes2018SemanticSC,zhang2018semantic,Zhang2018EfficientSS,Dourado2019EdgeNetSS,Chen20193DSS,Wang2019ForkNetMV,Zhang2019CascadedCP,Wang2020DeepOC,Chen2020RealTimeSS,Yan2020SparseSS,Dourado2020SemanticSC,Wu2020SCFusionRI,Cherabier2018LearningPF,Dai2018ScanCompleteLS,Wang2019Learning3S}. Since completion heavily requires contextual information it is common practice to use a U-Net architecture~\cite{Ronneberger2015UNetCN} (see Fig.~\ref{fig:architecture_schemes}), i.e. encoder-decoder with skip connections. The benefit of the latter is not only to provide contextual information but also to enable meaningful coarser scene representation, used in SSC for outputting multi-scale predictions~\cite{Roldao2020LMSCNetLM,Zhang2018EfficientSS} or for enabling coarse-to-fine refinement~\cite{Dai2018ScanCompleteLS}.

There are two important limitations of 3D CNNs: their cubically growing memory need, and the dilation of the sparse input manifold due to convolutions. 
To circumvent both, one can use sparse 3D networks like SparseConvNet~\cite{Graham20183DSS,Zhang2018EfficientSS} or Minkowski Network~\cite{choy20194d} which conveniently operates only on input geometry, thus allowing high grid resolution where each cell contains typically a single point. 
While they were found highly beneficial for most 3D tasks, they show limited interest for SSC since the output is expected to be denser than the network input. Considering the output to be \textit{sparse}, Dai \textit{et al.}~\cite{Dai2019SGNNSG} used a sparse encoder and a custom sparse generative decoder to restrict the manifold dilation, applied for SC rather than SSC. This is beneficial but cannot cope with large chunks of missing data. An alternative is~\cite{Yan2020SparseSS} that first performs pointwise semantic labeling using a sparse network. 
To enable a \textit{dense} SSC output in~\cite{Zhang2018EfficientSS}, authors merge the output of multiple \textit{shared} SparseConvNets applied on sub-sampled non-overlapping sparse grids. 
Both~\cite{Zhang2018EfficientSS,Dai2019SGNNSG} are a clever use of sparse convolutions but somehow limit the memory and computational benefit of the latter. In~\cite{Cherabier2018LearningPF}, variational optimization is used to regularize the model and avoid the need for greedy high-capacity 3D CNN.

\paragraph{View-volume networks.} 
To take advantage of 2D CNNs efficiency, a common strategy is to use them in conjunction with 3D CNNs as in~\cite{Guo2018ViewVolumeNF, Liu2018SeeAT, Li2020AnisotropicCN, Li2019RGBDBD, Liu20203DGR, Roldao2020LMSCNetLM,Li2020AttentionbasedMF}, see Fig.~\ref{fig:arch_view_volume}. Two different schemes have been identified from the literature. The most common, as in~\cite{Guo2018ViewVolumeNF, Liu2018SeeAT, Li2020AnisotropicCN, Li2019RGBDBD, Liu20203DGR,Li2020AttentionbasedMF}, employs a 2D CNN encoder to extract 2-dimensional features from 2D texture/geometry inputs (RGB, depth, etc.), which are then lifted to 3D and processed by 3D CNN (Fig.~\ref{fig:arch_view_volume},~3D-backbone). 
Optional modalities may be added with additional branches and mid-fusion scheme~\cite{Liu2018SeeAT, Li2020AnisotropicCN, Li2019RGBDBD, Liu20203DGR}. In brief, the use of sequential 2D-3D CNNs conveniently benefits of
 different neighboring definitions, since 2D neighboring pixels might be far away in 3D and vice versa, thus providing rich feature representation, at the cost of increased processing.
To address this limitation, the second scheme (Fig.~\ref{fig:arch_view_volume},~2D-backbone) projects 3D input data into 2D, then processed with normal 2D CNN~\cite{Roldao2020LMSCNetLM} significantly lighter than its 3D counterpart. 
The resulting 2D features are then lifted back to 3D and decoded with 3D convolutions, retrieving the third dimension before the final prediction.
This latter scheme is irrefutably lighter in computation and memory footprint but better suited for outdoor scenes (cf. Sec.~\ref{sec:evaluation}), as the data exhibits main variance along two axes (i.e. longitudinal and lateral).

\paragraph{Point-based networks.} 
To ultimately prevent discretization of the input data, a few recent works~\cite{Zhong2020SemanticPC,Rist2020SemanticSC,Rist2020SCSSnetLS} employ point-based networks, see Fig.~\ref{fig:arch_point_based}. 
In 2018,~\cite{Yuan2018PCNPC} first proposed to apply permutation-invariant architecture~\cite{Qi2017PointNetDH} to object completion with promising results. 
However, its use for generative SSC is hindered by the limited points generation capacity, the need for fixed size output, and the use of global features extraction.
To date, only SPC-Net~\cite{Zhong2020SemanticPC} relies solely on a point-based network -- $\mathcal{X}$Conv~\cite{Li2018PointCNNCO} -- to predict the semantics of observed and unobserved points. The fixed size limitation is circumvented by assuming a regular distribution of unobserved points, addressing the problem as a point segmentation task.
Overall, we believe point-based SSC has yet attracted too few works and is a viable avenue of research.

\paragraph{Hybrid networks.} 
Several other works combine architectures already mentioned, which we refer to as \textit{hybrid networks}~\cite{Rist2020SemanticSC, Rist2020SCSSnetLS,Cheng2020S3CNet,Zhong2020SemanticPC,Garbade2019TwoS3,Behley2019SemanticKITTIAD,Li2020DepthBS,Chen20203DSS}, see Fig.~\ref{fig:arch_hybrid}.
A common 2D-3D scheme combines 2D and 3D features (e.g. RGB and f-TSDF) through expert modality networks in a common latent space decoded via a 3D CNN~\cite{Garbade2019TwoS3, Chen20203DSS} (Fig.~\ref{fig:arch_hybrid},~mix-2D-3D). This enables additional benefit with the combined use of texture and geometrical features. 
Similarly, IPF-SPCNet~\cite{Zhong2020SemanticPC} performs semantic segmentation from RGB image on an initial 2D CNN and lifts image labels to augment ad-hoc 3D points (Fig.~\ref{fig:arch_hybrid},~point-augmented). In~\cite{Rist2020SemanticSC,Rist2020SCSSnetLS}, PointNet~\cite{Qi2017PointNetDH} encodes geometrical features from sub-set of points later convolved in a bird eye view (BEV) projection with 2D CNN in a hybrid architecture manner (Fig.~\ref{fig:arch_hybrid},~point-2D). 
Another strategy employs parallel 2D-3D branches to process the same data with different neighborhoods definition contained in the 2D projected image and 3D grid as in~\cite{Li2020DepthBS}. 
Recently, S3CNet~\cite{Cheng2020S3CNet} combines 3D voxelized f-TSDF and normal features with a 2D BEV~\cite{Chen2017Multiview3O}, passing both modalities through sparse encoder-decoder networks for late fusion (Fig.~\ref{fig:arch_hybrid},~parallel-2D-3D), achieving impressive results in outdoor scenes. A similar architecture is proposed by~\cite{Abbasi2018Deep3S} to perform what they refer to as \textit{scene extrapolation} \cite{Song2018Im2Pano3DE3}, by performing extrapolation of a half-known scene into a complete one.

\subsection{Design choices}
\label{sec:design}

The significant sparsity difference between the input data and the expected \textit{dense} output imposes special design choices to be made, specifically to ensure efficient flow of information. 
In the following, we elaborate on the most important ones such as contextual awareness~(Sec.~\ref{sec:mscale_aggregation}), position awareness~(Sec.~\ref{sec:position_awareness}), and fusion strategies~(Sec.~\ref{sec:fusion}). 
Finally, we detail lightweight designs (Sec.~\ref{sec:optimization_strategies}) to efficiently process 3D large extent of sparse data, and the common refinement processes~(Sec.~\ref{sec:refinement}).

\subsubsection{Contextual awareness} \label{sec:mscale_aggregation}

\begin{figure}
	\centering
	\subfloat[Serial]{\includegraphics[width=0.92\linewidth]{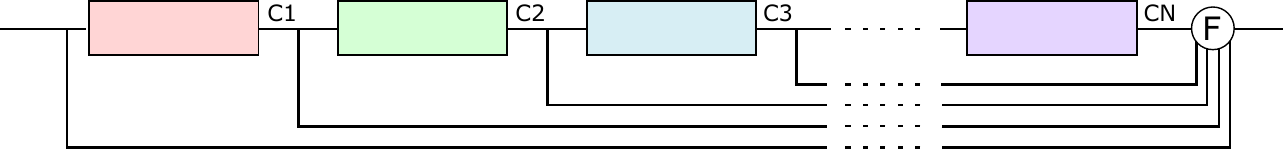}\label{fig:context_serial}}\\
	\subfloat[Self-cascaded]{\includegraphics[width=0.63\linewidth]{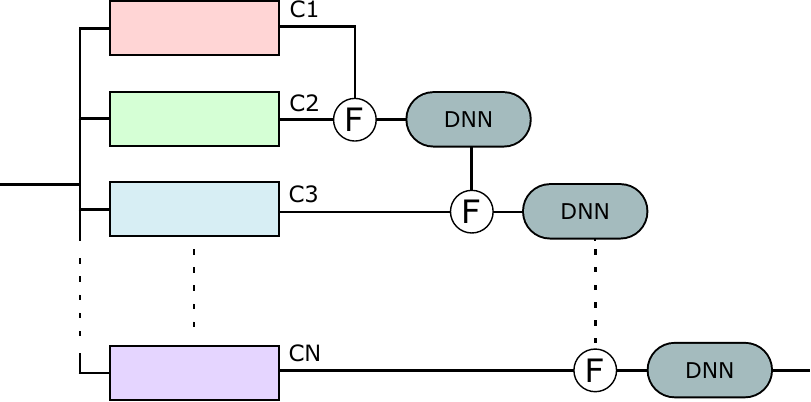}\label{fig:context_cascade}}
	\hfill
	\subfloat[Parallel]{\includegraphics[width=0.34\linewidth]{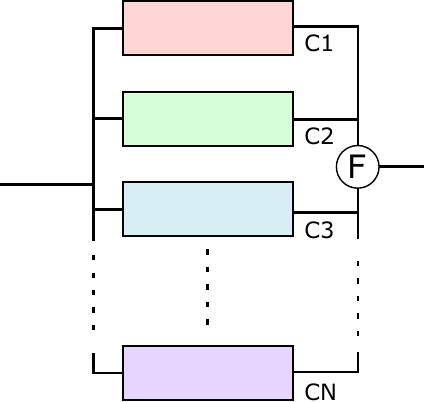}\label{fig:context_parallel}}
	\caption{\textbf{Multi-scale contextual aggregation.} While context is indubitably important for SSC, different strategies are used to aggregate features from various spatial/scale contexts. Color blocks stand for convolutions with different dilation rates. \raisebox{.5pt}{\textcircled{\raisebox{-.15pt} {\scriptsize{}F}}} stands for any type of fusion.}
	\label{fig:mscale_aggregation}
\end{figure}

To correctly complete the missing information in the scene, it is necessary to gather contextual information from multiple scales, which enables to disambiguate between similar objects present in the scene. This makes it possible to capture both local geometric details and high-level contextual information~\cite{Song2017SemanticSC}. 
A common strategy to accomplish this is to add skip connections between different convolutional layers to aggregate features from different receptive fields~\cite{Song2017SemanticSC, Roldao2020LMSCNetLM, Zhang2018EfficientSS, Guo2018ViewVolumeNF, Garbade2019TwoS3,Liu2018SeeAT,Chen20193DSS,Dourado2019EdgeNetSS,Zhang2019CascadedCP,Chen20203DSS,Dai2019SGNNSG,Dourado2020SemanticSC}. 
Additionally, serial context aggregation with multi-scale feature fusion can be used as proposed in \cite{Li2020DepthBS}, Fig.~\ref{fig:context_serial}.
In~\cite{Song2017SemanticSC}, the use of dilated convolutions (aka ‘atrous’)~\cite{Yu2016MultiScaleCA} are proposed to increase receptive fields and gather context information at low computational cost. The strategy became popular among most works~\cite{Song2017SemanticSC,Guo2018ViewVolumeNF,Garbade2019TwoS3,Liu2018SeeAT,Wang2019ForkNetMV,Dourado2019EdgeNetSS,Dourado2020SemanticSC,Zhang2019CascadedCP,Li2020AttentionbasedMF,Chen20203DSS,Li2020DepthBS,Roldao2020LMSCNetLM,Wu2020SCFusionRI}. Such convolutions are only suitable for dense networks (as opposed to sparse networks), and even then should only be applied in deeper layers of the network after dilation of the input manifold. 
In~\cite{Liu2018SeeAT}, a feature aggregation module is introduced by using Atrous Spatial Pyramid Pooling blocks (ASPP)~\cite{Chen2018DeepLabSI}, which exploits multi-scale features by employing multiple parallel filters with different dilation rates, Fig.~\ref{fig:context_parallel}. A lightweight ASPP is presented in~\cite{Li2019RGBDBD}. 
Dilated convolutions in the ASPP module can be replaced by Residual dilated blocks~\cite{He2016DeepRL} to increase spatial context and improve gradient flow. A Guided Residual Block (GRB) to amplify fused features and a Global Aggregation module to aggregate global context through 3D global pooling are proposed in~\cite{Chen2020RealTimeSS}. 
An additional feature aggregation strategy is presented in~\cite{Zhang2019CascadedCP}, where multi-context aggregation is achieved by a cascade pyramid architecture, Fig.~\ref{fig:context_cascade}. 
In~\cite{Cherabier2018LearningPF} multi-scale features are aggregated together following a Primal-Dual optimization algorithm~\cite{pock2011diagonal}, which ensures semantically stable predictions and further acts as a regularizer for the learning.

\subsubsection{Position awareness} \label{sec:position_awareness}

Geometric information contained in voxels at different positions has high variability, i.e. \textit{Local Geometric Anisotropy}. In particular, voxels inside an object are homogeneous and likely to belong to the same semantic category as their neighbors. Conversely, voxels at the surface, edges, and vertices of the scene provide richer geometric information due to the higher variance of their surroundings. To deal with this, PALNet~\cite{Li2020DepthBS} proposes a Position Aware loss (cf. Sec.~\ref{sec:losses}), which consists of a cross entropy loss with individual voxel weights assigned according to their geometric anisotropy, 
providing slightly faster convergence and improving results. Likewise, AM$^{2}$FNet~\cite{Chen2019AM2FNetAM} supervises contour information by an additional cross entropy loss as a supplementary cue for segmentation. 

In the same line of work, EdgeNet~\cite{Dourado2019EdgeNetSS} calculates Canny edges~\cite{Canny1986ACA} then fused with an f-TSDF obtained from the depth image. Hence, it increases the gradient along the geometrical edges of the scene. Additionally, detection of RGB edges enables the identification of objects lacking geometrical saliency. The same network is used in \cite{Dourado2020SemanticSC} to predict complete scenes from panoramic RGB-D images.

Similarly,~\cite{Chen20203DSS} introduces an explicit and compact geometric embedding from depth information by predicting a 3D sketch containing scene edges from an input TSDF. They show that this feature embedding is resolution-insensitive, which brings high benefit, even from partial low-resolution observations.

\subsubsection{Fusion strategies}\label{sec:fusion}

SSC requires outputting both geometry and semantics. Though highly coupled --~geometry helping semantics and vice-versa~--, there is a natural benefit to use inputs of different natures for example to provide additional texture or geometry insights. 
We found that about two-thirds of the literature use multi-modal inputs though it appears less trendy in most recent works (see Tab.~\ref{tab:methods} col `Input'). 
For the vast majority of multi-input works, RGB is used alongside various geometrical input~\cite{Guedes2018SemanticSC,Liu2018SeeAT,Li2019RGBDBD,Garbade2019TwoS3,Dourado2019EdgeNetSS,Behley2019SemanticKITTIAD,Chen2019AM2FNetAM,Dourado2020SemanticSC,Liu20203DGR,Li2020AttentionbasedMF,Chen20203DSS,Li2020AnisotropicCN,Zhong2020SemanticPC} as it is a natural candidate for semantics. 
Even without color, the fusing of 2D and 3D modalities is often employed because it enables richer feature estimation. This is because 2D and 3D neighborhoods differ, since 2D data results of planar projection along the optical axis of the sensor. 
Subsequently, a common strategy consists of fusing geometrical features processed with different 2D / 3D encoding to obtain richer local scene descriptors. 
In~\cite{Behley2019SemanticKITTIAD} depth and occupancy are fused while~\cite{Li2020DepthBS} uses depth along with TSDF-like data. As mentioned earlier (cf. Sec.~\ref{sec:input_encoding}), TSDF provides a gradient field easing the network convergence.
Finally, application-oriented fusion is also found such as fusing bird eye view along with geometrical inputs as in~\cite{Cheng2020S3CNet} -- which is better suited for outdoor SSC.

\begin{figure}
	\centering
	
	\subfloat[Early Fusion~\cite{Guedes2018SemanticSC,Guo2018ViewVolumeNF,Garbade2019TwoS3,Zhong2020SemanticPC,Dourado2020SemanticSC,Behley2019SemanticKITTIAD,Cherabier2018LearningPF}]{\includegraphics[width=0.65\linewidth]{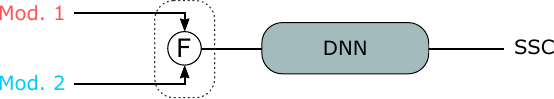}\label{fig:fusion_early}}\\

	\subfloat[Middle Fusion, single-stage:~\cite{Liu2018SeeAT,Dourado2019EdgeNetSS,Li2020DepthBS,Chen20203DSS}, multi-stage.:~\cite{Li2019RGBDBD,Liu20203DGR,Li2020AnisotropicCN,Chen2019AM2FNetAM}]{
		\renewcommand{\arraystretch}{0.6}
		\begin{tabular}{c | c}		
			\rotatebox{90}{\scriptsize single-stage} & \includegraphics[width=0.65\linewidth]{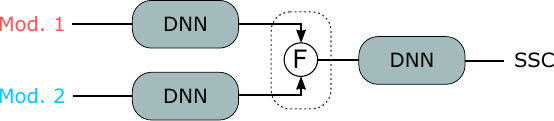}\\
			\multicolumn{2}{c}{}\\
			\rotatebox{90}{\scriptsize ~~~~~~~~~multi-stage} & \includegraphics[width=0.65\linewidth]{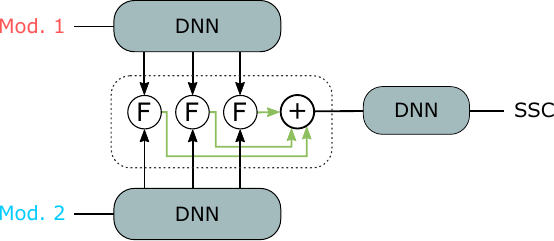} \\
			\label{fig:fusion_middle}
	\end{tabular}}

	\definecolor{modalityone}{rgb}{1, 0.3515, 0.3515}
	\definecolor{modalitytwo}{rgb}{0.1484, 0.8242, 19}
	
	\subfloat[Late Fusion~\cite{Li2020AttentionbasedMF,Liu2018SeeAT,Cheng2020S3CNet}]{\includegraphics[width=0.68\linewidth]{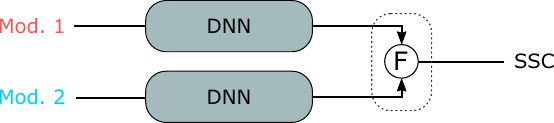}\label{fig:fusion_late}}
	
	\caption{\textbf{Fusion Strategies.} To accommodate for multiple input modalities (\textcolor{modalityone}{Mod. 1}, \textcolor{modalitytwo}{Mod. 2}), several fusion strategies are found in the literature. Here, \raisebox{.5pt}{\textcircled{\raisebox{-.15pt} {\scriptsize{}F}}} stands for fusion and might be any type of fusion like concat~\raisebox{.5pt}{\textcircled{\raisebox{-.15pt} {\scriptsize{}C}}}, sum~\raisebox{.5pt}{\textcircled{\raisebox{-.15pt} {\scriptsize{}+}}}, multiply~\raisebox{.5pt}{\textcircled{\raisebox{-.15pt} {\scriptsize{}$\times$}}}, convolutions, etc.}
	\label{fig:fusion_schemes}
\end{figure}

We group the type of fusion in three categories, shown in Fig.~\ref{fig:fusion_schemes}. Fusion applied at the input level (\textit{Early fusion}), at the mid-level features (\textit{Middle fusion}), or at the late/output level (\textit{Late fusion}). They are respectively referred to as \textit{E}, \textit{M}, and \textit{L} in column `Fusion strategies'~Tab.~\ref{tab:methods}. 

\paragraph{Early fusion.} The most trivial approach is to concatenate input modalities~\cite{Guedes2018SemanticSC,Guo2018ViewVolumeNF,Garbade2019TwoS3,Zhong2020SemanticPC,Dourado2020SemanticSC,Behley2019SemanticKITTIAD,Cherabier2018LearningPF} before any further processing, see Fig. \ref{fig:fusion_early}. There are two strategies here: when spatially aligned (e.g. RGB/Depth) inputs can be concatenated channel-wise; alternatively, inputs can be projected to a shared 3D common space (\textit{aka} features lifting). For spatially aligned modalities, it is common to use pairs of normals/depth~\cite{Guo2018ViewVolumeNF} or RGB/semantics~\cite{Behley2019SemanticKITTIAD} and to process them with 2D CNNs. The second strategy lifts any 2D inputs to 3D -- assuming depth information and accurate inter-sensors calibration -- and processes it with 3D networks. This has been done with 
RGB/depth~\cite{Guedes2018SemanticSC}, depth/semantics~\cite{Garbade2019TwoS3,Cherabier2018LearningPF}, points/semantics~\cite{Zhong2020SemanticPC}. Except when using points, this second strategy leads to a sparse tensor since not all 3D cells have features.
Noteworthy,~\cite{Garbade2019TwoS3,Behley2019SemanticKITTIAD,Zhong2020SemanticPC,Cherabier2018LearningPF} use semantics, which is first estimated either from RGB or depth-like data.
A 2D or 3D network processes the concatenated tensor, and while it logically outperforms single-modality input~\cite{Guo2018ViewVolumeNF,Dourado2019EdgeNetSS,Garbade2019TwoS3} there seems to be little benefit to apply early fusion.

\paragraph{Middle fusion.} To exploit all modalities, middle fusion uses expert networks that learn modality-centric features. A straightforward fusion strategy is employed in~\cite{Chen20203DSS,Dourado2019EdgeNetSS,Li2020DepthBS,Liu2018SeeAT} where the features are simply concatenated and processed with a U-Net style architecture (cf. Fig. \ref{fig:fusion_middle},~single-stage), which improves over early fusion but still limits the exchange of information between modalities.
The information flow is improved in~\cite{Li2019RGBDBD,Liu20203DGR,Li2020AnisotropicCN,Chen2019AM2FNetAM} by fusing modality-centric features in a multi-stage manner (cf. Fig. \ref{fig:fusion_middle},~multi-stage); meaning that low-level features from different modalities are fused together and aggregated with fused mid/high level features gathered similarly. While ultimately the number of fusion stages shall be a function of the input/output size, 3 stages are often used~\cite{Li2019RGBDBD,Li2020AnisotropicCN,Chen2019AM2FNetAM}, though~\cite{Liu20203DGR} claims 4 stages boost performances with similar input/output. The fused mechanism can be a simple summation~\cite{Li2019RGBDBD} or concatenation~\cite{Li2020AnisotropicCN}, but~\cite{Chen2019AM2FNetAM,Liu20203DGR} benefit from smarter selective fusion schemes using respectively RefineNet~\cite{lin2017refinenet} and Gated Recurrent Fusion.\\
Overall, the literature consensus is that middle fusion is highly efficient for SSC. The ablation studies of Liu \textit{et al.}~\cite{Liu20203DGR} reports that any selective fusion schemes bring at least a $20\%$ performance boost over simple sum/concat/max schemes.

\paragraph{Late fusion.} Few works use late fusion for SSC~\cite{Li2020AttentionbasedMF,Cheng2020S3CNet,Liu2018SeeAT}, see Fig. \ref{fig:fusion_late}. 
The straightforward strategy in~\cite{Li2020AttentionbasedMF} is to apply fusion -- namely, element-wise multiplication -- of two SSC branches (a 3D guidance branch, and a semantic completion branch), followed by a softmax. The benefit still appears little (5 to 10\%) given the extra computational effort. Similarly, color and geometry branches are concatenated and shallowly convolved before softmax in \cite{Liu2018SeeAT}, also providing a small benefit (less than 3\%). A unique strategy was proposed in the recent S3CNet~\cite{Cheng2020S3CNet} where the output of parallel 2D top-view \textit{and} 3D SSC are fused together in a semantic-wise manner. While it was only evaluated on outdoor scenes -- which setup naturally minimizes vertically overlapping semantic labels -- ablation reports an overall 20\% boost.\\

Summarizing the different strategies, \textit{Middle fusion} appears to be the best general SSC practice, though \textit{Late fusion} was found beneficial in some specific settings. On fused modalities, RGB/geometry fusion boosts performance but at the cost of an additional sensor need, but even using fusion of geometrical input with different encodings is highly beneficial. An interesting insight from~\cite{Dourado2019EdgeNetSS,Chen20203DSS} advocates that RGB \textit{or} geometry can be fused with edge features as they provide additional boundaries guidance for the SSC network.

\subsubsection{Lightweight designs} \label{sec:optimization_strategies}

A few techniques for lightweight designs are often applied for SSC with the aim of addressing two separate problems: how to improve the memory or computation efficiency, and how to design meaningful convolutions to improve the information flow. We detail either problem and its solutions below.

\paragraph{Memory and computation efficiency.}
Voxel grids are often used as input/output encoding of the 3D data since current datasets provide ground truth in such a format. 
However, only a tiny portion of the voxels are occupied which makes the naive \textit{dense} grid inefficient in memory and computation. 
Memory wise, a few works use compact hierarchical 3D representation inspired from pre-deep learning, like Kd-Tree~\cite{bentley1975multidimensional} and Octree~\cite{meagher1982geometric}. 
Octree-based deep networks are often used for learning object reconstruction~\cite{Wang2017OCNNOC, Riegler2017OctNetFusionLD, Riegler2017OctNetLD, Wang2018AdaptiveOA} though little applied on real semantic scene completion problem~\cite{Cherabier2018LearningPF,Wang2019Learning3S,Wang2020DeepOC}. 
Meanwhile, deep {Kd}-Networks~\cite{Klokov2017EscapeFC} proposal seems less appropriate and has not yet been applied to SSC.
Computation-wise,~\cite{Cherabier2018LearningPF} proposed a custom network architecture with adjustable multi-scale branches in which inference and backpropagation can be run in parallel, subsequently enabling faster training and good performance with low-capacity dense 3D CNNs.
Alternatively, few SSC or SC works~\cite{Zhang2018EfficientSS, Dai2019SGNNSG} use sparse networks like SparseConvNet~\cite{Graham20183DSS} or Minkowski~\cite{choy20194d} which operate only in active locations through a hash table. 
While sparse convolutions are very memory/computation efficient, they are less suitable for completion, since they deliberately avoid filling empty voxels to prevent dilation of the input domain. 
To remedy this for the SSC task, dense convolutions are still applied in the decoder, which subsequently reduces sparse networks efficiency.\\
Overall, while Kd-/Octree- networks are highly memory efficient, the complexity of their implementation has restricted a wider application. Contrastingly, sparse networks~\cite{Graham20183DSS,choy20194d} are more used \cite{Zhang2018EfficientSS,Cheng2020S3CNet,Yan2020SparseSS,Dai2019SGNNSG}.

\paragraph{Efficient convolutions.}
A key observation is the spatial redundancy of data since neighboring voxels contain similar information. To exploit such redundancy,~\cite{Zhang2018EfficientSS} proposes Spatial Group Convolutions (SGC) to divide input volume into different sparse tensors along the spatial dimensions which are then convolved with shared sparse networks. 
A similar strategy is followed by~\cite{Dai2018ScanCompleteLS}, dividing the volumetric space into a set of eight interleaved voxel groups and performing an auto-regressive prediction~\cite{Reed2017ParallelMA}. 
Dilated convolutions are also widely used for semantic completion methods~\cite{Song2017SemanticSC,Guo2018ViewVolumeNF,Garbade2019TwoS3,Liu2018SeeAT,Wang2019ForkNetMV,Chen20193DSS,Li2019RGBDBD,Dourado2019EdgeNetSS,Zhang2019CascadedCP,Li2020AttentionbasedMF,Chen20203DSS,Li2020DepthBS,Liu20203DGR,Dourado2020SemanticSC,Roldao2020LMSCNetLM,Chen2020RealTimeSS}, since they increase receptive fields at small cost, providing large context, which is crucial for scene understanding as discussed in Sec.~\ref{sec:mscale_aggregation}.
Dilated convolutions with \textit{separated kernels} are proposed in~\cite{Zhang2019CascadedCP} by separating the input tensor into subvolumes. This enables to reduce the number of parameters and consider depth profiles in which depth values are continuous only in neighboring regions.\\
\noindent{}DDRNet~\cite{Li2019RGBDBD} also introduces Dimensional Decomposition Residual (DDR) block, decomposing 3D convolutions into three consecutive layers along each dimension, 
subsequently reducing the network parameters. 
In \cite{Li2020AnisotropicCN}, this concept is extended with the use of anisotropic convolutions, where the kernel size of each 1D convolution is adaptively learned during training to model the dimensional anisotropy.

\subsubsection{Refinement} \label{sec:refinement}
Refinement is commonly used in many vision tasks, but little applied in SSC. 
VD-CRF~\cite{zhang2018semantic} extends SSCNet~\cite{Song2017SemanticSC} by applying Conditional Random Field (CRF) to refine output consistency, achieving little over 4\% gain. Additionally, S3CNet~\cite{Cheng2020S3CNet} presents a 3D spatial propagation network \cite{Liu2017LearningAV} to refine segmentation results after fusion of 2D semantically completed bird eye view image and 3D grid.
Additional partial refinement is applied in \cite{Dourado2020SemanticSC,Wu2020SCFusionRI} to fuse SSC predictions from different viewpoints, by either softmax applied to overlapping partitions \cite{Dourado2020SemanticSC,Wu2020SCFusionRI} or an occupancy-based fusion policy \cite{Wu2020SCFusionRI}. 
Though few works address the refinement problem, some notable performance boosts are found in the literature, thus being an encouraging topic to explore.

\subsection{Training}
\label{sec:training}

We now detail the SSC training process, starting with the SSC losses (Sec.~\ref{sec:losses}), and subsequently the implemented training strategies (Sec.~\ref{sec:training_strategies}).

\subsubsection{Losses} \label{sec:losses}

We classify the SSC losses found in the literature in 3 broad categories: \textit{geometric~losses} which optimize geometrical accuracy, \textit{semantics~losses} which optimize semantics prediction, and \textit{consistency~losses} which guide the overall completion consistency.
Note that other non-SSC losses are often added and that the type of SSC losses are commonly mixed --~specifically, geometric+semantics~\cite{Dai2018ScanCompleteLS, Wang2020DeepOC, Chen2020RealTimeSS, Cheng2020S3CNet} or all three types~\cite{Chen20193DSS, Chen20203DSS, Wang2019ForkNetMV, Rist2020SemanticSC, Rist2020SCSSnetLS}. We refer to Tab.~\ref{tab:methods} for a quick overview of the losses used by each method.\\
In this section, we also refer to $\hat{y}$ as SSC prediction and $y$ as ground truth, though for clarity we add a subscript notation to distinguish between SSC encoding. For example, $y^\text{mesh}$ corresponds to the ground truth mesh.

\paragraph{Geometric losses.} 
These losses penalize the geometrical distance of the output $\hat{y}$ to ground truth $y$, in a self-unsupervised manner.

On occupancy grids outputs ($\hat{y}^\text{occ}$), \textbf{Binary Cross-Entropy} loss (BCE) is most often used~\cite{Rist2020SemanticSC,Wang2020DeepOC,Chen2020RealTimeSS,Cheng2020S3CNet, Rist2020SCSSnetLS} to discriminate \textit{free} voxels from \textit{occupied}. Assuming a binary class mapping where all non-free semantic classes map to `occupy'. It writes:
\begin{equation}\label{eq:BCELoss}
	\mathcal{L}_{\text{BCE}} = - \frac{1}{N} \sum_{i=0}^{N} \hat{y}^\text{occ}_{i}log(y^\text{occ}_i) - (1-\hat{y}^\text{occ}_{i})log(1-y^\text{occ}_i) \,,
\end{equation}
with $N$ the number of voxels.
The drawback of such loss is that it provides little guidance to the network due to its sparsity. Smoother guidance can be provided by outputting an implicit surface ($\hat{y}^\text{SDF}$) through minimization of the predicted signed distance values in $\hat{y}^\text{SDF}$ and corresponding SDF-encoded mesh ($y^\text{SDF}$) -- using $\boldsymbol{\ell_{1}}$ or $\boldsymbol{\ell_{2}}$ norms.

On points outputs ($\hat{y}^\text{pts}$), if SSC is approached as a generative task, the above losses could also be used to penalize distance to a ground truth mesh, though it might be more suitable to apply points-to-points distances, thus assuming a ground truth point cloud ($y^\text{pts}$). 
To that end, permutation invariant metrics as the \textbf{Chamfer Distance} (CD)~\cite{Fan2017APS} or the \textbf{Earth Mover's Distance} (EMD)~\cite{Kurenkov2018DeformNetFD, Fan2017APS} have been employed for object completion tasks~\cite{Fan2017APS, Yuan2018PCNPC} but have not been explored yet for SSC because of their computational greediness~\cite{Fan2017APS}. We highlight that such losses could provide an additional geometric supervision signal when used in conjunction with semantic losses described below.

\paragraph{Semantic losses.}
Such losses are suitable for occupancy grids or points and can accommodate for either $C$ classes (considering only semantics classes of occupied voxels or points) or $C+1$ classes (considering all voxels/points and `free space' being the additional class). Note that only the second case ($C+1$ classes) enforce reconstruction, so the first one ($C$ classes) would require additional \textit{geometric losses}.
\textbf{Cross-Entropy} loss (CE) is the preferred loss for SSC~\cite{Song2017SemanticSC, Roldao2020LMSCNetLM, Zhang2018EfficientSS, Guo2018ViewVolumeNF, Garbade2019TwoS3, Liu2018SeeAT, Li2019RGBDBD, Dourado2019EdgeNetSS, Zhang2019CascadedCP, Chen20193DSS, Li2020AnisotropicCN, Li2020AttentionbasedMF,Cherabier2018LearningPF}, it models classes as independent thus considering the latter to be equidistant in the semantic space.
Formally, supposing $(y,\hat{y})$ it writes:
\begin{equation}\label{eq:CELoss}
	\mathcal{L}_{\text{CE}} = - \frac{1}{C} \sum_{i=0}^{N}\sum_{c=0}^{N} w_{c}\hat{y}_{i, c}\log \left( \frac{e^{y_{i,c}}}{\textstyle \sum_{c^{\prime}}^{C} e^{y_{i,c^{\prime}}}} \right) \,,
\end{equation}
assuming here that $y$ is the one-hot-encoding of the classes (i.e. $y_{i, c}=1$ if $y_{i}$ label is $c$ and otherwise $y_{i, c}=0$). In practice, $(y,\hat{y})$ can be either occupancy grids $(y^\text{occ},\hat{y}^\text{occ})$ or points $(y^\text{pts},\hat{y}^\text{pts})$.
A rare practice from~\cite{Wang2019ForkNetMV} is to address classification with BCE (Eq.~\ref{eq:BCELoss}) through the sum of $C$ binary classification problems between each semantic class and the free class. However, such a practice is unusual and arguably beneficial. 

Recently, PALNet~\cite{Li2020DepthBS} proposed the \textbf{Position Aware} loss~(PA), a weighted cross-entropy accounting for the local semantics entropy to encourage sharper semantics/geometric gradients in the completion~(cf.~Sec~\ref{sec:position_awareness}). The loss writes:
\begin{equation}
	\mathcal{L}_{\text{PA}} = - \frac{1}{N} \sum_{i=0}^{N}\sum_{c=0}^{C} (\lambda + \alpha W_{\text{LGA}_i})\hat{y}^\text{occ}_{i, c}\log \left( \frac{e^{y^\text{occ}_{i,c}}}{\textstyle \sum_{c^{\prime}}^{C} e^{y^\text{occ}_{i,c^{\prime}}}} \right) \,,
\end{equation}
with $\lambda$ and $\alpha$ being simple base and weight terms, and $W_{\text{LGA}_i}$ being the \textit{Local Geometric Anisotropy} of $i$ that scales accordingly to the semantic entropy in its direct vicinity (i.e.~$W_{\text{LGA}}$ lowers in locally smooth semantics areas). We refer to~\cite{Li2020DepthBS} for an in-depth explanation. From the latter, $\mathcal{L}_{\text{PA}}$ leads to a small performance gain of $1$--$3\%$. Noteworthy, this loss could easily accommodate point clouds as well.\\

Note that \textit{geometric} or \textit{semantics} losses can only be computed on \textit{known} ground truth location, due to the ground truth sparsity. Additionally, because SSC is a highly imbalanced problem~(cf.~Fig.~\ref{fig:dset_frequenciess}), class-balancing strategy is often used.

\paragraph{Consistency losses.} 
Different from most \textit{semantics} losses, these losses~\cite{Chen20203DSS,Rist2020SemanticSC, Rist2020SCSSnetLS} provide a self-supervised semantic signal.
In~\cite{Chen20203DSS} the \textbf{completion consistency} (CCY) of predictions from multiple partitioned sparse inputs is enforced via a Kullback-Leibler divergence. 
Differently, \cite{Rist2020SemanticSC, Rist2020SCSSnetLS} enforces \textbf{spatial semantics consistency} (SCY) by minimizing the Jenssen-Shannon divergence of semantic inference between a given spatial point and some given \textit{support points}. This self-supervision signal is available at any position within the scene. However, the strategy for \textit{support points} is highly application dependent and while suitable for outdoor scenes which have repetitive semantic patterns, we conjecture it might not scale as efficiently to cluttered indoor scenes.\\

Overall, few self-supervised or even unsupervised strategies exist and we believe that such type of new losses~\cite{Zhang2021AMF} should be encouraged.

\subsubsection{Training strategies}
\label{sec:training_strategies}

\begin{figure}
	\centering
		
	\subfloat[End-to-end, single-scale:~\cite{Song2017SemanticSC, Guo2018ViewVolumeNF, Li2019RGBDBD, Dourado2019EdgeNetSS, Zhang2019CascadedCP, Li2020DepthBS, Liu20203DGR, Li2020AttentionbasedMF, Li2020AnisotropicCN,Chen2020RealTimeSS,Wang2020DeepOC,Garbade2019TwoS3,Rist2020SemanticSC,Cherabier2018LearningPF,Dourado2020SemanticSC}, multi-scale:~\cite{Zhang2018EfficientSS, Roldao2020LMSCNetLM}, multi-modal supervision:~\cite{Li2020AttentionbasedMF,Garbade2019TwoS3,Liu2018SeeAT,Zhong2020SemanticPC}]{
	\begin{tabular}{l|c c|c}
		
		\renewcommand{\arraystretch}{0.0001}
		\setlength{\tabcolsep}{0.0001\linewidth}
		\rotatebox{90}{\scriptsize single-scale} & \includegraphics[width=0.32\linewidth]{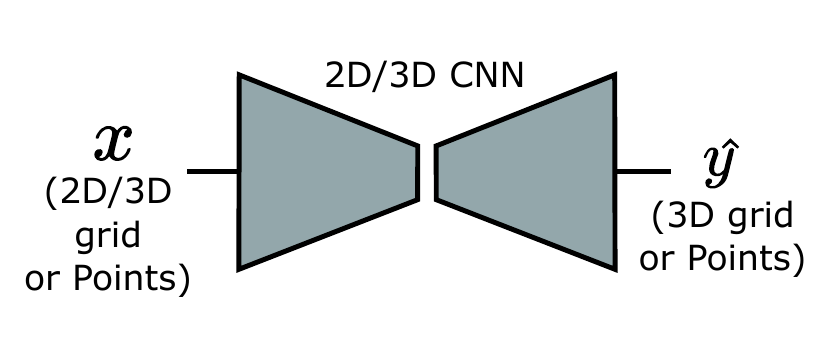} &
		\rotatebox{90}{\scriptsize multi-scale} &
		\includegraphics[width=0.33\linewidth]{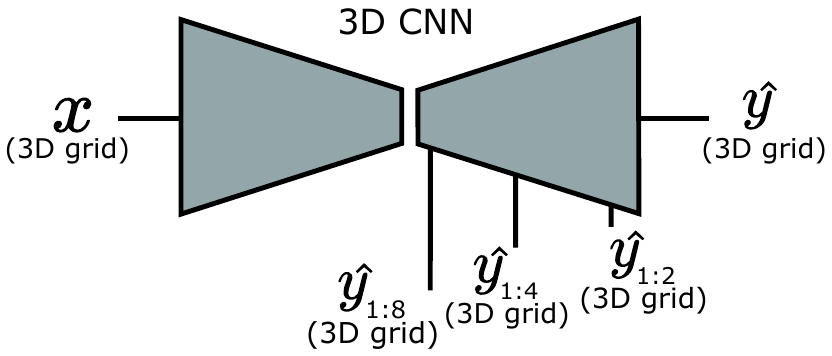}\\
		\multicolumn{4}{c}{}\\
		\rotatebox{90}{\scriptsize supervision} & \multicolumn{3}{c}{\includegraphics[width=0.78\linewidth]{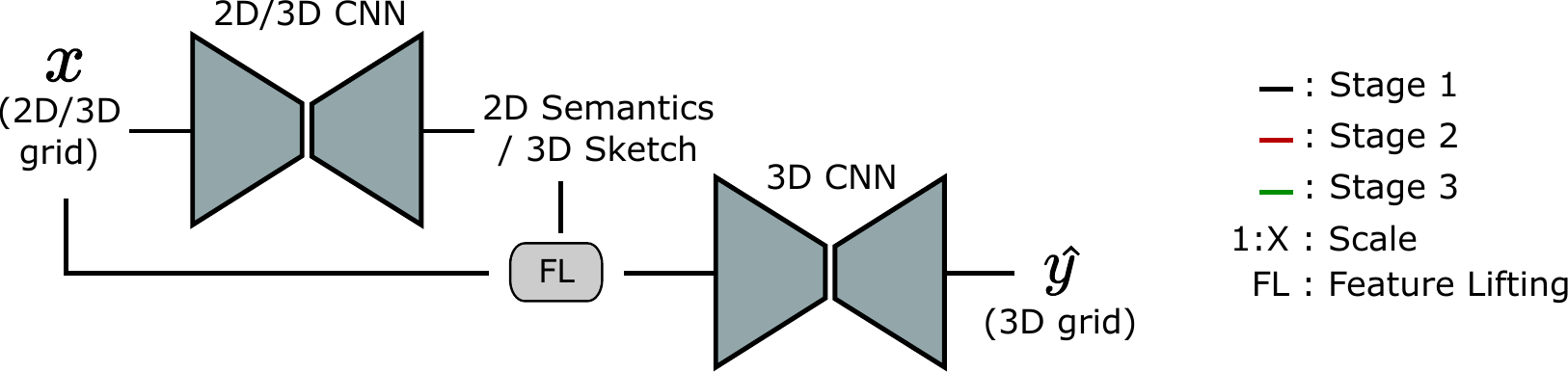}}\\
		\rotatebox{90}{\scriptsize ~~~~~~~~~~multi-modal} & \multicolumn{3}{c}{\includegraphics[width=0.7\linewidth]{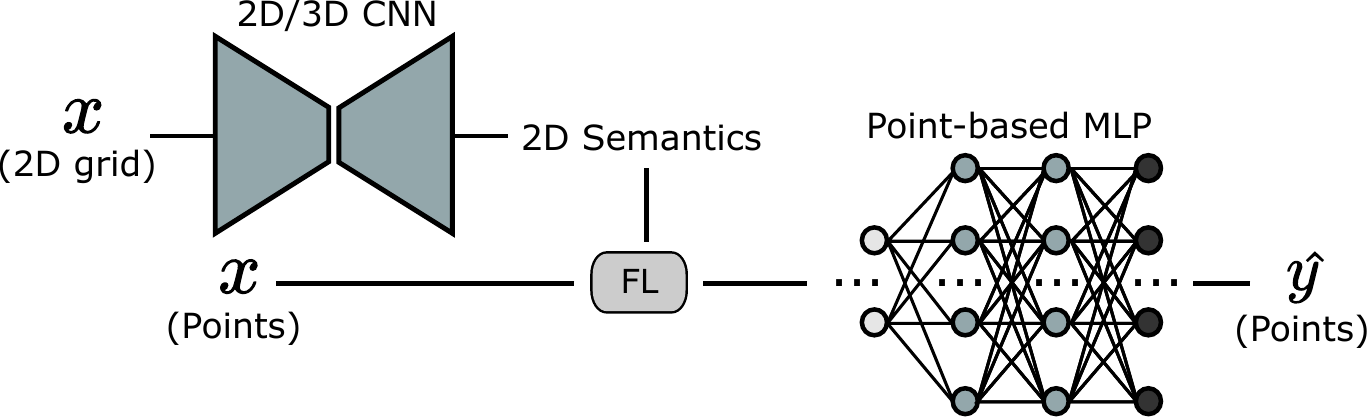}} \\
	\end{tabular}

	\label{fig:training_strategies_end_to_end}\label{fig:training_strategies_multiscale}\label{fig:training_strategies_two_stage}}
	\hfill

	\subfloat[Coarse-to-fine~\cite{Dai2018ScanCompleteLS}]{\includegraphics[width=0.49\linewidth]{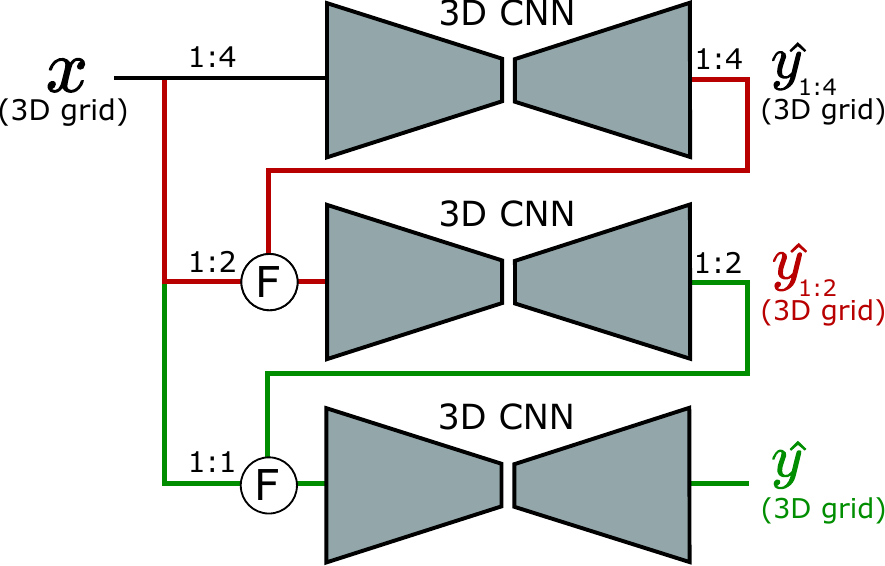}\label{fig:training_strategies_coarse_to_fine}}
	\hfill
	\subfloat[Adversarial~\cite{Wang2018AdversarialSS, Wang2019ForkNetMV, Chen20193DSS,Wu2020SCFusionRI}]{\includegraphics[width=0.48\linewidth]{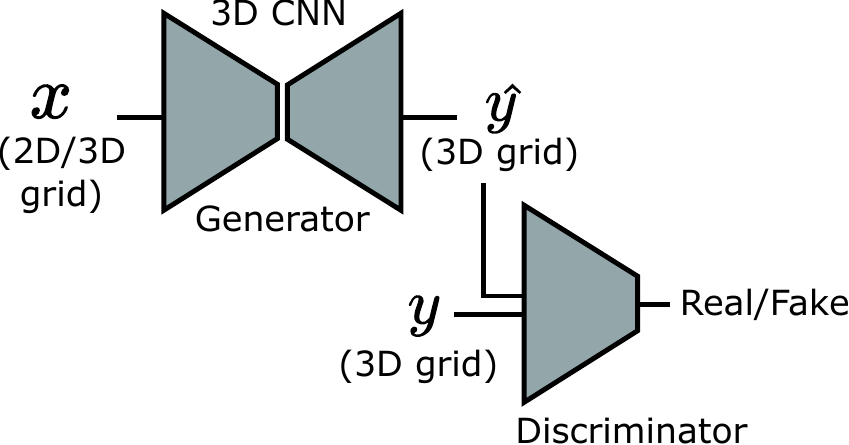}\label{fig:training_strategies_adversarial}}
	\caption{\textbf{Training strategies.} Most SSC architectures are trained end-to-end~\protect\subref{fig:training_strategies_end_to_end} outputting single or multi-scale SSC. Additionally, multi-modal supervision training commonly lift semantic features calculated on sparse input to a second stage network. Coarse-to-fine~\protect\subref{fig:training_strategies_coarse_to_fine}, similarly to multi-scale relies on multiple size predictions, but trains in a multi-stage coarse to fine manner. Finally, Adversarial training~\protect\subref{fig:training_strategies_adversarial} discriminates between ground truth and predicted scenes. \raisebox{.5pt}{\textcircled{\raisebox{-.15pt} {\scriptsize{}F}}} stands for fusion of any type.}
	\label{fig:training_strategies}
\end{figure}

The vast majority of SSC works are trained end-to-end for \textit{single-scale} reconstruction, although others prefer \textit{multi-scale} supervision to output SSC at different resolutions. 
Few works also employ \textit{multi-modal supervision} commonly relying on auxiliary 2D semantic segmentation. 
It is also possible to train $n$-stage networks with \textit{coarse-to-fine} strategies, or even train with \textit{adversarial} learning to enforce realism.
Strategies are illustrated in Fig.~\ref{fig:training_strategies} (with link color indicating the stage) and reviewed below.

\paragraph{End-to-end.}
Most architectures (Fig.~\ref{fig:training_strategies_end_to_end},~top left) are trained end-to-end and output a \textit{single scale} SSC~\cite{Song2017SemanticSC, Guo2018ViewVolumeNF, Li2019RGBDBD, Dourado2019EdgeNetSS, Zhang2019CascadedCP, Li2020DepthBS, Liu20203DGR, Li2020AnisotropicCN, Chen2020RealTimeSS,Garbade2019TwoS3,Dourado2020SemanticSC,Cherabier2018LearningPF,Rist2020SemanticSC,Wang2020DeepOC} -- often similar to the input size. Training that way is straightforward and often offers minimal memory footprint. 
Noteworthy,~\cite{Dai2019SGNNSG} -- which does geometric completion only -- gradually increases sparsity during training to ease the completion of large missing chunks.\\
To guide the training, multi-scale SSC outputs can also be supervised, typically from the early layers of a U-Net decoder. A simple, yet efficient \textit{multi-scale} strategy~\cite{Roldao2020LMSCNetLM, Zhang2018EfficientSS} is to minimize the sum of SSC losses at different resolutions (Fig.~\ref{fig:training_strategies_end_to_end},~top right), thus also enforcing coarse SSC representations in the network. In~\cite{Zhang2018EfficientSS}, two different scales are predicted, versus four in~\cite{Roldao2020LMSCNetLM} providing down to 1:8 (1~over~8) downscaled SSC. In the latter, authors also report that the decoder can be ablated to provide very fast inference at coarsest resolution (370FPS at 1:8 scale).
When available, some works leverage multi-modal supervision relying on intermediate auxiliary tasks, typically 2D or 3D semantics, later used along original input data to infer the final SSC (Fig.~\ref{fig:training_strategies_two_stage},~bottom), as in~\cite{Li2020AttentionbasedMF,Garbade2019TwoS3,Zhong2020SemanticPC,Liu2018SeeAT}. The latter could also be trained in a two-stage manner.
In general, end-to-end training is conducted from scratch, though some~\cite{Dourado2019EdgeNetSS,Zhang2019CascadedCP,Liu2018SeeAT,Song2017SemanticSC,Guo2018ViewVolumeNF} report pretraining on the synthetic SUNCG dataset.

\paragraph{Coarse-to-fine.} 
ScanComplete~\cite{Dai2018ScanCompleteLS} also follows a multi-scale strategy somehow close to~\cite{Roldao2020LMSCNetLM, Zhang2018EfficientSS}, though training in a coarse-to-fine manner (Fig.~\ref{fig:training_strategies_coarse_to_fine}). In detail, three sequential training are achieved at increasingly higher resolutions, with each stage network taking as input the ad-hoc sparse input and the previous stage SSC prediction (for stage$>$1). 
Interestingly, no one explored a continuous curriculum learning setting, which could yield stabler training and performance improvement. 
Still,~\cite{Cherabier2018LearningPF} (intentionally omitted Fig.~\ref{fig:training_strategies_coarse_to_fine}) applies a unique coarse-to-fine proposal in a fully end-to-end manner, via parallel backpropagations in all scales.
Of similar spirit,~\cite{Dai2019SGNNSG} proposes an iteration-based progressive refinement during training for scene completion, but insights of such strategy are not deeply discussed. 

\paragraph{Adversarial.} 
Even SSC ground truth has large missing chunks of data, leading to ambiguous supervision. 
To address this,~\cite{Wang2018AdversarialSS, Wang2019ForkNetMV, Chen20193DSS,Wu2020SCFusionRI} use adversarial training~(Fig.~\ref{fig:training_strategies_adversarial}), since the discriminator provides an additional supervision signal.
This is straightforwardly implemented in~\cite{Wu2020SCFusionRI,Chen20193DSS}, where the discriminator classifies ground truth from generated SSC (aka real/fake).
In~\cite{Wang2018AdversarialSS, Wang2019ForkNetMV} of same authors, 2 discriminators are used in a somehow similar fashion to discriminate both the SSC output and the latent depth or semantics features to enforce deep shared representation.
Additionally, \cite{Chen20203DSS} employs a Conditional Variational Autoencoder (CVAE) to generate completed border sketches to be fed to the main SSC branch.
Despite few works on the matter, adversarial appears a logical choice to improve SSC consistency and provide additional self-supervision. Both~\cite{Wang2018AdversarialSS} and~\cite{Wu2020SCFusionRI} report a 10\%-15\% boost on several datasets.\\

Finally, on implementation -- where mentioned -- only~\cite{Dai2018ScanCompleteLS,Zhong2020SemanticPC,Wang2018AdversarialSS,Wang2019ForkNetMV,Roldao2020LMSCNetLM} train with Adam optimizer,~\cite{Chen20193DSS} with a mix of Adam/SGD, and all others use only SGD with momentum 0.9 and $10^{-4}$ weight decay, except for~\cite{Zhang2019CascadedCP,Dourado2019EdgeNetSS,Li2020AttentionbasedMF,Chen2020RealTimeSS,Wang2017OCNNOC,Wang2020DeepOC,Chen20193DSS} using $5\times{}10^{-4}$. The training most often uses standard learning rate scheduler~\cite{Li2019RGBDBD,Zhang2019CascadedCP,Li2020DepthBS,Liu20203DGR,Li2020AnisotropicCN,Chen2020RealTimeSS,Wang2017OCNNOC,Zhong2020SemanticPC,Garbade2019TwoS3,Wang2020DeepOC} though sophisticated scheduling~\cite{Dourado2019EdgeNetSS} or fixed learning rate~\cite{Li2020AttentionbasedMF} are also used.
Because of 3D greediness, the common practice is to train with small batch size of 1~\cite{Li2020AttentionbasedMF}, 2~\cite{Li2019RGBDBD}, 3~\cite{Dourado2019EdgeNetSS}, 4~\cite{Guo2018ViewVolumeNF,Zhang2019CascadedCP,Li2020DepthBS,Liu20203DGR,Li2020AnisotropicCN,Dourado2019EdgeNetSS,Roldao2020LMSCNetLM,Zhong2020SemanticPC,Chen20193DSS,Cherabier2018LearningPF}, 8~\cite{Wang2017OCNNOC,Wang2020DeepOC,Wang2018AdversarialSS,Wang2019ForkNetMV} or 16~\cite{Chen2020RealTimeSS} to fit in standard 12GB GPUs.

\subsection{Evaluation} \label{sec:evaluation}

We now provide an in-depth evaluation of the field, reviewing first the common metrics~(Sec. \ref{sec:eval_metrics}), the qualitative and quantitative performance of the literature~(Sec. \ref{sec:eval_performance}), and the networks' efficiency~(Sec. \ref{sec:eval_neteff}). 

\subsubsection{Metrics}\label{sec:eval_metrics}
\paragraph{Joint Semantics-Geometry.} The preferred metric for SSC is the  mean \textbf{Jaccard Index} or \textbf{mean Intersection over Union} (mIoU) \cite{Everingham2014ThePV}, which considers IoU of all semantic classes for prediction, without considering free space. It writes
\begin{equation}\label{eq:mIoU}
\text{mIoU} = \frac{1}{C} \sum_{c=1}^{C} \frac{\text{TP}_c}{\text{TP}_c + \text{FP}_c + \text{FN}_c} \,,
\end{equation}
where $\text{TP}_c$, $\text{FP}_c$ and $\text{FN}_c$ are the true positives, false positives and false negatives predictions for class $c$, respectively. Since ground truth is commonly semi-dense for real-world datasets, evaluation is performed in known space only.

\paragraph{Geometry only.} Because mIoU considers semantic classes, the pure geometrical reconstruction quality is not encompassed. 
Therefore \textbf{Intersection over Union} (\textit{IoU}), along with \textbf{Precision} and \textbf{Recall} are commonly used on the binary free/occupy scene representation, obtained by mapping all semantic classes to \textit{occupy}.

Alternatively, any distance metrics from Sec.~\ref{sec:losses} (i.e. $\boldsymbol{\ell_{1}}$, $\boldsymbol{\ell_{2}}$, \textbf{EMD} or \textbf{CD}) may be used as in~\cite{Dai2018ScanCompleteLS, Dai2019SGNNSG} though less used in real datasets, due to their lower precision when sparsity increases.\\

On common practice, we highlight that evaluation on real indoor or outdoor datasets is usually performed differently. 
This results of the common sensors setup, respectively RGB-D (indoor) and LiDAR (outdoor), providing significantly different density information. 
Referring to Fig.~\ref{fig:SC_sensors_problem}, in real indoor~\cite{Song2017SemanticSC} the geometrical IoU is evaluated on input occluded regions while the mIoU is evaluated on input occluded~(blue) and observed~(red) surfaces.
In real outdoor~\cite{Behley2019SemanticKITTIAD} the IoU and mIoU are commonly evaluated on the entire known space, regardless of whether regions were observed or occluded in the input.
Obviously, synthetic datasets can cope with either practice. In the following, we describe the common practices and report semantics metrics (mIoU) along with geometrical ones (Precision, Recall, IoU).

\subsubsection{Performance}\label{sec:eval_performance}
\begin{table*}
	\scriptsize
	\def\mystrut{\rule{0pt}{1.5\normalbaselineskip}}
	\centering
	\rowcolors[]{8}{black!3}{white}
	\setlength{\tabcolsep}{0.005\linewidth}
	\newcommand{\metvariant}[1]{{~~~~#1}}
	
	\def\colortops#1#2#3#4#5#6{
		\def\temp{#1}\ifx\temp\empty
		-
		\else
		\ifdim#1pt=#2pt\cellcolor{red!50}\textbf{#1}\else
		\ifdim#1pt=#3pt\cellcolor{red!40}#1\else
		\ifdim#1pt=#4pt\cellcolor{red!30}#1\else
		\ifdim#1pt=#5pt\cellcolor{red!20}#1\else
		\ifdim#1pt=#6pt\cellcolor{red!10}#1\else
		#1
		\fi\fi\fi\fi\fi
		\fi
	}
	\definecolor{rowblue}{rgb}{0.8627451, 0.9803922, 0.9803922}
	\newcommand{\nyuvtwo}[5]{
		\colortops{#1}{92.1}{85.0}{76.0}{74.8}{74.2}\def\temp{#1}\ifx\temp\empty\else#5\fi& 
		\colortops{#2}{94.5}{91.8}{90.8}{87.8}{85.4}\def\temp{#2}\ifx\temp\empty\else#5\fi& 
		\colortops{#3}{78.2}{73.4}{71.3}{63.5}{62.2}\def\temp{#3}\ifx\temp\empty\else#5\fi& 
		\colortops{#4}{52.4}{41.1}{38.5}{35.1}{34.4}\def\temp{#4}\ifx\temp\empty\else#5\fi 
	}
	\newcommand{\nyucad}[5]{
		\colortops{#1}{94.1}{91.3}{91.1}{90.6}{88.7}\def\temp{#1}\ifx\temp\empty\else#5\fi& 
		\colortops{#2}{92.6}{92.3}{92.2}{91.7}{91.2}\def\temp{#2}\ifx\temp\empty\else#5\fi& 
		\colortops{#3}{86.3}{84.2}{82.4}{82.2}{80.8}\def\temp{#3}\ifx\temp\empty\else#5\fi& 
		\colortops{#4}{63.5}{55.2}{53.2}{46.6}{46.2}\def\temp{#4}\ifx\temp\empty\else#5\fi 
	}
	\newcommand{\suncg}[5]{
		\colortops{#1}{98.2}{93.1}{92.6}{92.1}{90.8}\def\temp{#1}\ifx\temp\empty\else#5\fi& 
		\colortops{#2}{96.8}{96.5}{95.5}{95.2}{92.4}\def\temp{#2}\ifx\temp\empty\else#5\fi& 
		\colortops{#3}{91.4}{88.2}{88.1}{86.9}{84.8}\def\temp{#3}\ifx\temp\empty\else#5\fi& 
		\colortops{#4}{76.5}{74.8}{74.2}{70.5}{69.5}\def\temp{#4}\ifx\temp\empty\else#5\fi 
	}
	\newcommand{\semkit}[5]{
		\colortops{#1}{81.6}{80.5}{77.1}{71.5}{62.6}\def\temp{#1}\ifx\temp\empty\else#5\fi& 
		\colortops{#2}{88.3}{84.2}{83.4}{75.5}{73.5}\def\temp{#2}\ifx\temp\empty\else#5\fi& 
		\colortops{#3}{57.7}{56.7}{56.6}{55.3}{50.6}\def\temp{#3}\ifx\temp\empty\else#5\fi& 
		\colortops{#4}{29.5}{23.8}{22.7}{17.7}{17.6}\def\temp{#4}\ifx\temp\empty\else#5\fi 
	}

	\begin{tabular}{rl | c | ccc | c| ccc | c| ccc | c | ccc | c}
		\multicolumn{3}{c}{} & \multicolumn{12}{c}{\scriptsize \textbf{Indoor}} & 
		\multicolumn{4}{c}{\scriptsize \textbf{Outdoor}} \\
		\cmidrule(lr){4-15}\cmidrule(lr){16-19}
		\multicolumn{3}{c}{} & \multicolumn{4}{c|}{\scriptsize {Real-world}} & \multicolumn{8}{c|}{\scriptsize {Synthetic}} & 
		\multicolumn{4}{c}{\scriptsize {Real-world}} \\
		\multicolumn{3}{c}{} & \multicolumn{4}{c|}{NYUv2~\cite{Silberman2012IndoorSA}} & \multicolumn{4}{c|}{NYUCAD~\cite{Firman2016StructuredPO}} & \multicolumn{4}{c|}{SUNCG~\cite{Song2017SemanticSC}} & \multicolumn{4}{c}{SemanticKITTI~\cite{Behley2019SemanticKITTIAD}} \\
		\multicolumn{3}{c}{} & \multicolumn{4}{c|}{\scriptsize $60\times36\times60$} & \multicolumn{4}{c|}{\scriptsize $60\times36\times60$} & \multicolumn{4}{c|}{\scriptsize $60\times36\times60$} & \multicolumn{4}{c}{\scriptsize $256\times32\times256$} \\
		\midrule
		& Method & Input & Prec. & Recall & IoU & mIoU & Prec. & Recall & IoU & mIoU  & Prec. & Recall & IoU & mIoU & Prec. & Recall & IoU & mIoU \\
		\midrule
		2017 & SSCNet$^{\dag}$~\cite{Song2017SemanticSC}\textsuperscript{a} & G & \nyuvtwo{57.0}{94.5}{55.1}{24.7}{}&\nyucad{75.0}{92.3}{70.3}{}{} & \suncg{76.3}{95.2}{73.5}{46.4}{} & \semkit{31.7}{83.4}{29.8}{9.5}{}\\

		& \metvariant{SSCNet-full~\cite{Roldao2020LMSCNetLM}} & G & \nyuvtwo{}{}{}{}{} & \nyucad{}{}{}{}{} & \suncg{}{}{}{}{} & \semkit{59.6}{75.5}{50.0}{16.1}{}\\
		
		2018 & Guedes et al.~\cite{Guedes2018SemanticSC} & G+T & \nyuvtwo{62.5}{82.3}{54.3}{27.5}{} & \nyucad{}{}{}{}{} & \suncg{}{}{}{}{} & \semkit{}{}{}{}{}\\
		
		& VVNet$^{\dag}$~\cite{Guo2018ViewVolumeNF}  & G &  \nyuvtwo{74.8}{74.5}{59.5}{29.3}{$^{\ddagger}$}&\nyucad{91.1}{86.6}{79.8}{40.9}{$^{\ddagger}$} &\suncg{90.8}{91.7}{84.0}{66.7}{} & \semkit{}{}{}{}{}\\

		& VD-CRF$^{\dag}$~\cite{zhang2018semantic} & G &  \multicolumn{3}{r|}{\lsstyle$^{\dag}$ refer to Tab. \ref{tab:results_datasets_perf_pretrainSUNCG}}  & & \multicolumn{3}{r|}{\lsstyle$^{\dag}$ refer to Tab. \ref{tab:results_datasets_perf_pretrainSUNCG}}  & & \suncg{}{}{74.5}{48.8}{} & \semkit{}{}{}{}{}\\

		& ESSCNet~\cite{Zhang2018EfficientSS}\textsuperscript{b} & G & \nyuvtwo{71.9}{71.9}{56.2}{26.7}{} & \nyucad{}{}{}{}{} & \suncg{92.6}{90.4}{84.5}{70.5}{} & \semkit{62.6}{55.6}{41.8}{17.5}{}\\

		& SATNet$^{\dag}$~\cite{Liu2018SeeAT} & G+T & \multicolumn{3}{r|}{\lsstyle$^{\dag}$ refer to Tab. \ref{tab:results_datasets_perf_pretrainSUNCG}} & & \nyucad{}{}{}{}{} & \suncg{80.7}{96.5}{78.5}{64.3}{* }&\semkit{}{}{}{}{}\\

		2019 & DDRNet~\cite{Li2019RGBDBD} & G+T & \nyuvtwo{71.5}{80.8}{61.0}{30.4}{} & \nyucad{88.7}{88.5}{79.4}{42.8}{} & \suncg{}{}{}{}{} & \semkit{}{}{}{}{}\\
		
		& TS3D~\cite{Garbade2019TwoS3}\textsuperscript{{c}} & G+T & \nyuvtwo{}{}{60.0}{34.1}{} & \nyucad{}{}{76.1}{46.2}{} & \suncg{}{}{}{}{} & \semkit{31.6}{84.2}{29.8}{9.5}{}\\  
		
		& \metvariant{TS3D+DNet~\cite{ Behley2019SemanticKITTIAD}} & G & \nyuvtwo{}{}{}{}{} & \nyucad{}{}{}{}{} & \suncg{}{}{}{}{} & \semkit{25.9}{88.3}{25.0}{10.2}{}\\
		
		& \metvariant{TS3D+DNet+SATNet~\cite{Behley2019SemanticKITTIAD}} & G & \nyuvtwo{}{}{}{}{} & \nyucad{}{}{}{}{} & \suncg{}{}{}{}{} & \semkit{80.5}{57.7}{50.6}{17.7}{}\\
		
		& EdgeNet$^{\dag}$~\cite{Dourado2019EdgeNetSS} & G+T & \nyuvtwo{76.0}{68.3}{56.1}{27.8}{}&\nyucad{}{}{}{}{} & \suncg{93.1}{90.4}{84.8}{69.5}{* }&\semkit{}{}{}{}{}\\

		& SSC-GAN~\cite{Chen20193DSS} & G & \nyuvtwo{63.1}{87.8}{57.8}{22.7}{} & \nyucad{80.7}{91.1}{74.8}{42.0}{} & \suncg{83.4}{92.4}{	78.1}{55.6}{} & \semkit{}{}{}{}{}\\ 
		
		& ForkNet$^{\dag}$~\cite{Wang2019ForkNetMV} & G & \multicolumn{3}{r|}{\lsstyle$^{\dag}$ refer to Tab. \ref{tab:results_datasets_perf_pretrainSUNCG}} & & \nyucad{}{}{}{}{} & \suncg{}{}{86.9}{63.4}{$^\wedge$} & \semkit{}{}{}{}{} \\

		& CCPNet$^{\dag}$~\cite{Zhang2019CascadedCP} & G & \nyuvtwo{74.2}{90.8}{63.5}{38.5}{$^\wedge$}&\nyucad{91.3}{92.6}{82.4}{53.2}{$^\wedge$} &\suncg{98.2}{96.8}{91.4}{74.2}{$^\wedge$} & \semkit{}{}{}{}{}\\

		& AM$^{2}$FNet~\cite{Chen2019AM2FNetAM} & G+T  & \nyuvtwo{72.1}{80.4}{61.3}{31.7}{} & \nyucad{87.2}{91.0}{80.2}{44.6}{} & \suncg{}{}{}{}{} & \semkit{}{}{}{}{}\\
		
		2020 & GRFNet~\cite{Liu20203DGR} & G+T & \nyuvtwo{68.4}{85.4}{61.2}{32.9}{} & \nyucad{87.2}{91.0}{80.1}{45.3}{} & \suncg{}{}{}{}{} & \semkit{}{}{}{}{}\\
		
		& AMFNet~\cite{Li2020AttentionbasedMF} & G+T & \nyuvtwo{67.9}{82.3}{59.0}{33.0}{} & \nyucad{}{}{}{}{} & \suncg{}{}{}{}{} & \semkit{}{}{}{}{}\\
		
		& PALNet~\cite{Li2020DepthBS} & G & \nyuvtwo{68.7}{85.0}{61.3}{34.1}{} & \nyucad{87.2}{91.7}{80.8}{46.6}{} & \suncg{}{}{}{}{} & \semkit{}{}{}{}{}\\
		
		& 3DSketch~\cite{Chen20203DSS} & G+T & \nyuvtwo{85.0}{81.6}{71.3}{41.1}{} & \nyucad{90.6}{92.2}{84.2}{55.2}{} & \suncg{}{}{88.2}{76.5}{* }&\semkit{}{}{}{}{}\\
		
		& AIC-Net~\cite{Li2020AnisotropicCN} & G+T & \nyuvtwo{62.4}{91.8}{59.2}{33.3}{} & \nyucad{88.2}{90.3}{80.5}{45.8}{} & \suncg{}{}{}{}{} & \semkit{}{}{}{}{}\\
		
		& Wang et al.~\cite{Wang2020DeepOC} & G & \nyuvtwo{}{}{}{}{} & \nyucad{}{}{}{}{} & \suncg{92.1}{95.5}{88.1}{74.8}{} & \semkit{}{}{}{}{}\\
		
		& IPF-SPCNet~\cite{Zhong2020SemanticPC} & G+T & \nyuvtwo{70.5}{46.7}{39.0}{35.1}{} & \nyucad{83.3}{72.7}{63.5}{50.7}{} & \suncg{}{}{}{}{} & \semkit{}{}{}{}{}\\ 
		
		& Chen et al.~\cite{Chen2020RealTimeSS} & G & \nyuvtwo{}{}{73.4}{34.4}{} & \nyucad{}{}{82.2}{44.5}{} & \suncg{}{}{84.8}{63.5}{} & \semkit{}{}{}{}{}\\
		
		& LMSCNet~\cite{Roldao2020LMSCNetLM} & G & \nyuvtwo{}{}{}{}{} & \nyucad{}{}{}{}{} & \suncg{}{}{}{}{} & \semkit{77.1}{66.2}{55.3}{17.0}{}\\
		
		& \metvariant{LMSCNet-SS~\cite{Roldao2020LMSCNetLM}} & G & \nyuvtwo{}{}{62.2}{28.4}{$^{\ddagger}$}&\nyucad{}{}{}{}{} & \suncg{}{}{}{}{} & \semkit{81.6}{65.1}{56.7}{17.6}{}\\
		
		& S3CNet~\cite{Cheng2020S3CNet} & G & \nyuvtwo{}{}{}{}{} & \nyucad{}{}{}{}{} & \suncg{}{}{}{}{} & \semkit{}{}{45.6}{29.5}{}\\
		
		& JS3C-Net~\cite{Yan2020SparseSS} & G & \nyuvtwo{}{}{}{}{} & \nyucad{}{}{}{}{} & \suncg{}{}{}{}{} & \semkit{71.5}{73.5}{56.6}{23.8}{}\\
		
		& Local-DIFs~\cite{Rist2020SemanticSC} & G & \nyuvtwo{}{}{}{}{} & \nyucad{}{}{}{}{} & \suncg{}{}{}{}{} & \semkit{}{}{57.7}{22.7}{}\\	
		
		2021 & SISNet~\cite{Cai2021SemanticSC} & G+T & \nyuvtwo{92.1}{83.8}{78.2}{52.4}{} & \nyucad{94.1}{91.2}{86.3}{63.5}{} & \suncg{}{}{}{}{} & \semkit{}{}{}{}{}\\	
		
		\bottomrule
	\end{tabular}\\
	{\scriptsize \textsuperscript{a} Results in SemanticKITTI reported on ~\cite{Roldao2020LMSCNetLM}.}
	{\scriptsize \textsuperscript{b} Results in SemanticKITTI reported on ~\cite{Yan2020SparseSS}.}
	{\scriptsize \textsuperscript{c} Results in SemanticKITTI reported on ~\cite{Behley2019SemanticKITTIAD}.}
	\\
	{Input: Geometry (depth, range, points, etc.), Texture (RGB).}
	{\scriptsize * Texture input not used due to absence in SUNCG.}\\
	{\scriptsize $^{\dag}$ Results with SUNCG pre-training on NYUv2 and NYUCAD presented in Tab. \ref{tab:results_datasets_perf_pretrainSUNCG}.}
	{\scriptsize $^{\ddagger}$ Results provided by authors.}\\
	{\scriptsize $^{\wedge}$ Results reported at a different resolution. CCPNet: ($240\times144\times240$). ForkNet: ($80\times48\times80$).}
	\caption{\textbf{SSC performance on the most popular datasets without pretraining.} 
		The relatively low best mIoU scores on the challenging real outdoor SemanticKitti~\cite{Behley2019SemanticKITTIAD} ($29.5\%$) and real indoor NYUv2~\cite{Silberman2012IndoorSA} ($41.1\%$) shows the complexity of the task. 
		In the `method' column, we indicate variants with an offset.
		To better interpret the performance, column `Input' shows the type of input modality used where `G' is Geometry (depth, range, points, etc.) and `T' is Texture (RGB).
		Note that all indoor datasets commonly report performance for $60\times36\times60$ grids for historical reasons though 4x bigger input is commonly treated, cf. Sec.~\ref{sec:eval_performance}. Top 5 methods are highlighted in each column from red to white.}
	\label{tab:results_datasets_perf}
\end{table*}

We report the available mIoU and IoU performance on the most popular SSC datasets in Tab.~\ref{tab:results_datasets_perf}, which are all obtained from voxelized ground truth. For non-voxel methods the output is voxelized beforehand. Additionally, detailed classwise performance of top five methods for SemanticKITTI, NYUv2 and SUNCG are presented in Tabs.~\ref{table:semantic_kitti_semantics}, \ref{table:semantic_NYUv2} and \ref{table:semantic_SUNCG}, respectively. From the performance Tab.~\ref{tab:results_datasets_perf}, the mIoU of the best methods plateaus around $75-85\%$ on synthetic indoor dataset, $52\%$ on real indoor, and $30\%$ on real outdoor.
Importantly, note that \textit{most} indoor datasets performance are evaluated at 1:4 of the original ground truth resolution -- that is $60\times{}36\times{}60$ -- for historical reasons\footnote{\label{foot:indoorgtscale}In their seminal work, for memory reason Song \textit{et al.}~\cite{Song2017SemanticSC} evaluated SSC only at the 1:4 scale. Subsequently, to provide fair comparisons between indoor datasets and methods, most other indoor SSC have been using the same resolution despite the fact that higher resolution ground truth is available. Recent experiments in~\cite{Chen20203DSS} advocate that using higher input/output resolution boosts the SSC performance significantly.}. Additionally, some methods refine parameters on NYUv2 and NYUCAD after SUNCG pre-training and are shown separately in Tab.~\ref{tab:results_datasets_perf_pretrainSUNCG}. This makes indoor / outdoor performance comparison tricky. It is interesting to note that IoU -- geometrical completion (i.e. ignoring semantics) -- is way higher than best mIoU. In detail, best IoU are $78\%$ on real indoor, and $57\%$ on real outdoor.\\
Qualitative results of a dozen of methods are shown in Fig.~\ref{fig:qualitative_indoor_1} for indoor datasets, and Fig.~\ref{fig:qualitative_outdoor_1} for outdoor datasets.\\

Overall, one may note the synthetic to real best performance gap of indoor datasets, which is approx. ${10-35\%}$ mIoU and $10-18\%$ IoU. 
While a difference is expected, once again it highlights that geometry has a smaller synthetic/real domain gap compared to semantics. 
On a general note also, most methods perform significantly better on IoU than on mIoU, demonstrating the complexity of the \textit{semantics} scene completion. 
In fact, the ranking of methods differs depending on the metric. 
For example, on NYUv2 (indoor) the recently presented SISNet has best mIoU ($52.4\%$) and IoU ($78.2\%$) by a large margin thanks to their semantic instance completion block and iterative SSC architecture. By reliyng solely on SSC, CCPNet~\cite{Zhang2019CascadedCP} gets best indoor mIoU ($41.3\%$), although achieved through SUNCG pre-training. Conversely, 3DSketch~\cite{Chen20203DSS} achieves similar mIoU ($41.1\%$) by training solely on NYUv2 and Chen et al.~\cite{Chen2020RealTimeSS} is second overall on IoU ($73.4\%$) with the same setup. On
SemanticKITTI~(outdoor) S3CNet~\cite{Cheng2020S3CNet} has best mIoU ($29.5\%$) and Local-DIFs \cite{Rist2020SemanticSC} best IoU ($57.7\%$). 
Note also the large difference between best indoor/outdoor metrics. While only a handful of methods \cite{Song2017SemanticSC,Zhang2018EfficientSS,Liu2018SeeAT,Roldao2020LMSCNetLM} are evaluated in both setups, they indeed perform significantly worse on outdoor data -- though indoor/outdoor performance should be carefully compared given the different resolution. 
This is partially explained by the higher sparsity in outdoor datasets, visible in `input' of Fig.~\ref{fig:qualitative_outdoor_1}. 
Another explanation is the higher number of classes in SemanticKITTI versus NYU and the extreme class-imbalance setup given that minor classes are very rarely observed, see Fig.~\ref{fig:dset_frequenciess}. 
On general qualitative results, either indoor (Fig.~\ref{fig:qualitative_indoor_1}) or outdoor (Fig.~\ref{fig:qualitative_outdoor_1}) results show that predictions are accurate in large homogeneous areas (walls/ground, floor, buildings) and most errors occur at object boundaries. 
This is evident in Tab.~\ref{table:semantic_kitti_semantics}, where most methods achieve high performance in the largest classes of SemanticKITTI, but struggle with predictions in less represented ones (e.g. bicycle, motorcycle, person). Worth mentioning, S3CNet~\cite{Cheng2020S3CNet} achieves considerably larger scores in rare classes (+25\%, +37\%, +38\% respectively), more than twice when compared to next best classed scores. The reason for such behavior is regrettably not deeply explored in their work. 

\paragraph{Inputs.}
To ease interpretation, col `Input' in Tab.~\ref{tab:results_datasets_perf} shows the nature of input used, where `G' is Geometry of any type (depth, TSDF, points, etc.) -- possibly several -- and `T' is Texture (RGB).
From Tab.~\ref{tab:results_datasets_perf} using both geometry \textit{and} texture (G+T) performs among the best indoor, such as SISNet~\cite{Cai2021SemanticSC} using both RGB and TSDF, and 3DSketch~\cite{Chen20203DSS} which relies on textural edges and depth. Generally speaking, G+T enables the prediction of non-salient geometric objects (i.e. paints, windows, doors) as shown in Fig.~\ref{fig:qual_in_a} by the door predicted 3DSketch~\cite{Chen20203DSS} and missed by \mbox{SSCNet}~\cite{Song2017SemanticSC}. 
Noteworthy, among the best mIoU methods~\cite{Zhang2019CascadedCP,Cheng2020S3CNet,Chen20203DSS} all use TSDF-encoding as geometrical input.
On outdoor datasets, only TS3D~\cite{Garbade2019TwoS3} uses texture without significant improvement.  
More works are required to evaluate the benefit of RGB modality on outdoor data.

\begin{table*}
	\scriptsize
	\setlength{\tabcolsep}{0.005\linewidth}
	\newcommand{\classfreq}[1]{{~\tiny(\semkitfreq{#1}\%)}}  
	
	\def\mystrut{\rule{0pt}{1.5\normalbaselineskip}}
	\centering
	\rowcolors[]{2}{black!3}{white}
	
	\begin{tabular}{l| c | c c c c c c c c c c c c c c c c c c c|c}
		
		\toprule

		Method 
		& \rotatebox{90}{Input}
		& \rotatebox{90}{\textcolor{road}{$\blacksquare$} road\classfreq{road}} 
		& \rotatebox{90}{\textcolor{sidewalk}{$\blacksquare$} sidewalk\classfreq{sidewalk}}
		& \rotatebox{90}{\textcolor{parking}{$\blacksquare$} parking\classfreq{parking}} 
		& \rotatebox{90}{\textcolor{other-ground}{$\blacksquare$} other-gr.\classfreq{otherground}} 
		& \rotatebox{90}{\textcolor{building}{$\blacksquare$} building\classfreq{building}} 
		& \rotatebox{90}{\textcolor{car}{$\blacksquare$} car\classfreq{car}} 
		& \rotatebox{90}{\textcolor{truck}{$\blacksquare$} truck\classfreq{truck}} 
		& \rotatebox{90}{\textcolor{bicycle}{$\blacksquare$} bicycle\classfreq{bicycle}} 
		& \rotatebox{90}{\textcolor{motorcycle}{$\blacksquare$} motorcycle\classfreq{motorcycle}} 
		& \rotatebox{90}{\textcolor{other-vehicle}{$\blacksquare$} other-veh.\classfreq{othervehicle}} 
		& \rotatebox{90}{\textcolor{vegetation}{$\blacksquare$} vegetation\classfreq{vegetation}} 
		& \rotatebox{90}{\textcolor{trunk}{$\blacksquare$} trunk\classfreq{trunk}} 
		& \rotatebox{90}{\textcolor{terrain}{$\blacksquare$} terrain\classfreq{terrain}} 
		& \rotatebox{90}{\textcolor{person}{$\blacksquare$} person\classfreq{person}} 
		& \rotatebox{90}{\textcolor{bicyclist}{$\blacksquare$} bicyclist\classfreq{bicyclist}} 
		& \rotatebox{90}{\textcolor{motorcyclist}{$\blacksquare$} motorcyclist\classfreq{motorcyclist}} 
		& \rotatebox{90}{\textcolor{fence}{$\blacksquare$} fence\classfreq{fence}} 
		& \rotatebox{90}{\textcolor{pole}{$\blacksquare$} pole\classfreq{pole}} 
		& \rotatebox{90}{\textcolor{traffic-sign}{$\blacksquare$} tr. sign\classfreq{trafficsign}} 
		& \rotatebox{90}{mIoU}  \\
		\midrule

		S3CNet~\cite{Yan2020SparseSS} & G & 42.0 & 22.5 & 17.0 & 7.9 & \textbf{50.2} & 31.2 & 6.7 & \textbf{41.5} & \textbf{45.0} & \textbf{16.1} & 39.5 & \textbf{34.0} & 21.2 & \textbf{45.9} & \textbf{35.8} & \textbf{16.0} & \textbf{31.3} & \textbf{31.0} & \textbf{24.3} & \textbf{29.5} \\
		JS3CNet~\cite{Yan2020SparseSS} & G & 64.7 & 39.9 & 34.9 & \textbf{14.1} & 39.4 & 33.3 & \textbf{7.2} & 14.4 & 8.8 & 12.7 & \textbf{43.1} & 19.6 & \textbf{40.5} & 8.0 & 5.1 & 0.4 & 30.4 & 18.9 & 15.9 & 23.8 \\
		Local-DIFs~\cite{Rist2020SemanticSC} & G & \textbf{67.9} & \textbf{42.9} & \textbf{40.1} & 11.4 & 40.4 & \textbf{34.8} & 4.4 & 3.6 & 2.4 & 4.8 & 42.2 & 26.5 & 39.1 & 2.5 & 1.1 & 0 & 29.0 & 21.3 & 17.5 & 22.7 \\
		TS3D+DNet+SATNet~\cite{Behley2019SemanticKITTIAD} & G & 62.2 & 31.6 & 23.3 & 6.5 & 34.1 & 30.7 & 4.9 & 0 & 0 & 0.1 & 40.1 & 21.9 & 33.1 & 0 & 0 & 0 & 24.1 & 16.9 & 6.9 & 17.7 \\
		LMSCNet-SS\cite{Roldao2020LMSCNetLM} & G & 64.8 & 34.7 & 29.0 & 4.6 & 38.1 & 30.9 & 1.5 & 0 & 0 & 0.8 & 41.3 & 19.9 & 32.1 & 0 & 0 & 0 & 21.3 & 15.0 & 0.8 & 17.6 \\
		
		\bottomrule
	\end{tabular}\\
	
	\caption{\textbf{Detailed SSC class performance on SemanticKITTI \cite{Behley2019SemanticKITTIAD} dataset.} Best 5 methods are presented and ordered in decreasing mIoU performance from top to bottom.} 
	\label{table:semantic_kitti_semantics}
\end{table*}

\begin{table}	
	\scriptsize
	\setlength{\tabcolsep}{0.0045\linewidth}
	\newcommand{\classfreq}[1]{{~\tiny(\nyuvtwofreq{#1}\%)}}  
	
	\def\mystrut{\rule{0pt}{1.5\normalbaselineskip}}
	\centering
	\rowcolors[]{2}{black!3}{white}
	
	\begin{tabular}{l| c | c c c c c c c c c c c|c}
		
		\toprule

		Method 
		& \rotatebox{90}{Input}
		& \rotatebox{90}{\textcolor{ceiling}{$\blacksquare$} ceil. \classfreq{ceiling}} 
		& \rotatebox{90}{\textcolor{floor}{$\blacksquare$} floor \classfreq{floor}}
		& \rotatebox{90}{\textcolor{wall}{$\blacksquare$} wall \classfreq{wall}} 
		& \rotatebox{90}{\textcolor{window}{$\blacksquare$} win. \classfreq{window}} 
		& \rotatebox{90}{\textcolor{chair}{$\blacksquare$} chair \classfreq{chair}} 
		& \rotatebox{90}{\textcolor{bed}{$\blacksquare$} bed \classfreq{bed}} 
		& \rotatebox{90}{\textcolor{sofa}{$\blacksquare$} sofa \classfreq{sofa}} 
		& \rotatebox{90}{\textcolor{table}{$\blacksquare$} table \classfreq{table}} 
		& \rotatebox{90}{\textcolor{tvs}{$\blacksquare$} tvs \classfreq{tvs}} 
		& \rotatebox{90}{\textcolor{furniture}{$\blacksquare$} furn. \classfreq{furniture}} 
		& \rotatebox{90}{\textcolor{objects}{$\blacksquare$} objs. \classfreq{objects}} 
		& \rotatebox{90}{mIoU}  \\
		\midrule
		
		SISNet~\cite{Cai2021SemanticSC} & G+T & \textbf{54.7} & 93.8 & \textbf{53.2} & \textbf{41.9} & \textbf{43.6} & \textbf{66.2} & \textbf{61.4} & \textbf{38.1} & 29.8 & \textbf{53.9} & \textbf{40.3} & \textbf{52.4} \\
		CCPNet$^{\dag}$$^\wedge$~\cite{Zhang2019CascadedCP} & G & 25.5 & \textbf{98.5} & 38.8 & 27.1 & 27.3 & 64.8 & 58.4 & 21.5 & \textbf{30.1} & 38.4 & 23.8 & 41.3 \\
		3DSketch~\cite{Chen20203DSS} & G+T & 43.1 & 93.6 & 40.5 & 24.3 & 30.0 & 57.1 & 49.3 & 29.2 & 14.3 & 42.5 & 28.6 & 41.1 \\
		ForkNet$^{\dag}$$^\wedge$~\cite{Wang2019ForkNetMV} & G & 36.2 & 93.8 & 29.2 & 18.9 & 17.7 & 61.6 & 52.9 & 23.3 & 19.5 & 45.4 & 20.0 & 37.1 \\ 
		IPF-SPCNet~\cite{Zhong2020SemanticPC} & G+T & 32.7 & 66.0 & 41.2 & 17.2 & 34.7 & 55.3 & 47.0 & 21.7 & 12.5 & 38.4 & 19.2 & 35.1 \\
		
		\bottomrule
	\end{tabular}\\
	{\scriptsize $^\wedge$ Results reported on resolution different to ($60\times36\times60$).  CCPNet:~($240\times144\times240$). ForkNet:~($80\times48\times80$).}\\
	{\scriptsize $^{\dag}$ Pretraining on SUNCG.}
	\caption{\textbf{Detailed SSC class performance on NYUv2~\cite{Silberman2012IndoorSA} dataset.} Best 5 methods from Tabs.~\ref{tab:results_datasets_perf} and \ref{tab:results_datasets_perf_pretrainSUNCG} are presented and ordered in decreasing mIoU performance from top to bottom.} 
	\label{table:semantic_NYUv2}
\end{table}

\begin{table}	
	\scriptsize
	\setlength{\tabcolsep}{0.0045\linewidth}
	\newcommand{\classfreq}[1]{{~\tiny(\suncgfreq{#1}\%)}}  
	
	\def\mystrut{\rule{0pt}{1.5\normalbaselineskip}}
	\centering
	\rowcolors[]{2}{black!3}{white}
	
	\begin{tabular}{l| c | c c c c c c c c c c c|c}
		
		\toprule

		Method 
		& \rotatebox{90}{Input}
		& \rotatebox{90}{\textcolor{ceiling}{$\blacksquare$} ceil. \classfreq{ceiling}} 
		& \rotatebox{90}{\textcolor{floor}{$\blacksquare$} floor \classfreq{floor}}
		& \rotatebox{90}{\textcolor{wall}{$\blacksquare$} wall \classfreq{wall}} 
		& \rotatebox{90}{\textcolor{window}{$\blacksquare$} win. \classfreq{window}} 
		& \rotatebox{90}{\textcolor{chair}{$\blacksquare$} chair \classfreq{chair}} 
		& \rotatebox{90}{\textcolor{bed}{$\blacksquare$} bed \classfreq{bed}} 
		& \rotatebox{90}{\textcolor{sofa}{$\blacksquare$} sofa \classfreq{sofa}} 
		& \rotatebox{90}{\textcolor{table}{$\blacksquare$} table \classfreq{table}} 
		& \rotatebox{90}{\textcolor{tvs}{$\blacksquare$} tvs \classfreq{tvs}} 
		& \rotatebox{90}{\textcolor{furniture}{$\blacksquare$} furn. \classfreq{furniture}} 
		& \rotatebox{90}{\textcolor{objects}{$\blacksquare$} objs. \classfreq{objects}} 
		& \rotatebox{90}{mIoU}  \\
		\midrule
		
		3DSketch$^{\ddagger}$~\cite{Chen20203DSS} & G* & 97.8 & 91.9 & \textbf{84.1} & \textbf{72.6} & 60.8 & 86.8 & 81.7 & \textbf{68.7} & 52.6 & 75.7 & \textbf{68.2} & \textbf{76.5} \\
		Wang et al.~\cite{Wang2020DeepOC} & G & 98.2 & 92.8 & 76.3 & 61.9 & \textbf{62.4} & \textbf{87.5} & 80.5 & 66.3 & \textbf{55.2} & 74.6 & 67.8 & 74.8 \\	
		CCPNet$^\wedge$~\cite{Zhang2019CascadedCP} & G & \textbf{99.2} & 89.3 & 76.2 & 63.3 & 58.2 & 86.1 & \textbf{82.6} & 65.6 & 53.2 & \textbf{76.8} & 65.2 & 74.2 \\
		ESSCNet~\cite{Zhang2018EfficientSS} & G & 96.6 & 83.7 & 74.9 & 59.0 & 55.1 & 83.3 & 78.0 & 61.5 & 47.4 & 73.5 & 62.9 & 70.5 \\
		EdgeNet~\cite{Dourado2019EdgeNetSS} & G* & 97.2 & \textbf{94.4} & 78.4 & 56.1 & 50.4 & 80.5 & 73.8 & 54.5 & 49.8 & 69.5 & 59.2 & 69.5 \\	
		
		\bottomrule
	\end{tabular}\\
	{\scriptsize $^\wedge$ Results reported on resolution different to ($60\times36\times60$).  CCPNet:~($240\times144\times240$).}\\
	{\scriptsize * Texture input not used due to absence in SUNCG.}\\
	{\scriptsize $^{\ddagger}$ Results provided by authors.}
	\caption{\textbf{Detailed SSC class performance on SUNCG~\cite{Song2017SemanticSC} dataset.} Best 5 methods are presented and ordered in decreasing mIoU performance from top to bottom.} 
	\label{table:semantic_SUNCG}
\end{table}

\begin{table}
	\scriptsize
	\def\mystrut{\rule{0pt}{1.5\normalbaselineskip}}
	\centering
	\rowcolors[]{8}{black!3}{white}
	\setlength{\tabcolsep}{0.005\linewidth}
	\newcommand{\metvariant}[1]{{~~~~#1}}
	
	\def\colortops#1#2#3#4#5#6{
		\def\temp{#1}\ifx\temp\empty
		-
		\else
		\ifdim#1pt=#2pt\cellcolor{red!50}\textbf{#1}\else
		\ifdim#1pt=#3pt\cellcolor{red!40}#1\else
		\ifdim#1pt=#4pt\cellcolor{red!30}#1\else
		\ifdim#1pt=#5pt\cellcolor{red!20}#1\else
		\ifdim#1pt=#6pt\cellcolor{red!10}#1\else
		#1
		\fi\fi\fi\fi\fi
		\fi
	}
	
	\begin{tabular}{rl | c | ccc | c| ccc | c }
		\multicolumn{3}{c}{} & \multicolumn{8}{c}{\scriptsize \textbf{Indoor}} \\
		\cmidrule(lr){4-11}
		\multicolumn{3}{c}{} & \multicolumn{4}{c|}{\scriptsize {Real-world}} & \multicolumn{4}{c}{\scriptsize {Synthetic}} \\
		\multicolumn{3}{c}{} & \multicolumn{4}{c|}{NYUv2~\cite{Silberman2012IndoorSA}} & \multicolumn{4}{c}{NYUCAD~\cite{Firman2016StructuredPO}}  \\
		\multicolumn{3}{c}{} & \multicolumn{4}{c|}{\scriptsize $60\times36\times60$} & \multicolumn{4}{c}{\scriptsize $60\times36\times60$}  \\
		\midrule
		& Method & Input & Prec. & Recall & IoU & mIoU & Prec. & Recall & IoU & mIoU  \\
		\midrule

		2017 & SSCNet~\cite{Song2017SemanticSC} & G & 59.3 & 92.9 & 56.6 & 30.5 & 75.0 & \textbf{96.0} & 73.0 & - \\			
		
		2018 & VVNet~\cite{Guo2018ViewVolumeNF}  & G & 69.8 & 83.1 & 61.1 & 32.9 & 86.4 & 92.0 & 80.3 & - \\ 
		
		& VD-CRF~\cite{zhang2018semantic} & G & - & - & 60.0 & 31.8 & - & - & 78.4 & 43.0 \\
		
		& SATNet~\cite{Liu2018SeeAT} & G+T & 67.3 & 85.8 & 60.6 & 34.4 & - & - & - & - \\ 
		
		2019 & EdgeNet~\cite{Dourado2019EdgeNetSS} & G+T & \textbf{79.1} & 66.6 & 56.7 & 33.7 & - & - & - & - \\
		
		& ForkNet~\cite{Wang2019ForkNetMV} & G & - & - & 63.4$^\wedge$ & 37.1$^\wedge$ & - & - & - & - \\ 
		
		& CCPNet~\cite{Zhang2019CascadedCP} & G & 78.8$^\wedge$ & \textbf{94.3}$^\wedge$ & \textbf{67.1}$^\wedge$ & \textbf{41.3}$^\wedge$ & \textbf{93.4}$^\wedge$ & 91.2$^\wedge$ & \textbf{85.1}$^\wedge$ & \textbf{55.0}$^\wedge$ \\

		\bottomrule
	\end{tabular}\\
	{Input: Geometry (depth, range, points, etc.), Texture (RGB).}\\
	{\scriptsize $^{\wedge}$ Results reported at a different resolution. CCPNet: ($240\times144\times240$). \\ ForkNet: ($80\times48\times80$).}
	\caption{\textbf{SSC performance on indoor datasets with SUNCG pre-training.} Synthetic data pre-training results on slight performance gains for all methods.}
	\label{tab:results_datasets_perf_pretrainSUNCG}
\end{table}
\paragraph{Architecture and design choices.}
One may notice the good performance of hybrid networks~\cite{Chen20203DSS,Li2020DepthBS,Zhong2020SemanticPC,Cheng2020S3CNet,Cai2021SemanticSC} (Fig.~\ref{fig:arch_hybrid}), which we believe results from richer input signal due to the fusion of multiple modalities.
We also argue that multiple neighboring definitions (2D and 3D) provide beneficial complementary signals. 
For instance, S3CNet combines 2D BEV and 3D f-TSDF for late fusion through post-processing refinement, achieving the best semantic completion performance on SemanticKITTI~\cite{Behley2019SemanticKITTIAD} by a considerable margin (+5.7\% mIoU). Qualitative results of the approach are shown in Fig.~\ref{fig:qual_out_c},~\ref{fig:qual_out_d}. Similarly, JS3CNet~\cite{Yan2020SparseSS} ranks second in the same dataset (23.8\% mIoU and 56.6\% IoU) with point-wise semantic labeling through SparseConvNet architecture~\cite{Graham20183DSS} and dense semantic completion using a point-voxel interaction module, enabling to better infer small vehicles as shown in circled areas of Fig.~\ref{fig:qual_out_e},~\ref{fig:qual_out_f}. 
Analogously, PALNet~\cite{Li2020DepthBS} middle fuses depth image and f-TSDF features, achieving good performance on NYUv2 (34.1\% mIoU and 61.3\% mIoU) and NYUCAD (46.6\% mIoU and 80.8\% mIoU) datasets, such performance can also be attributed to its position-aware loss, to be discussed next.

Contextual awareness (Sec.~\ref{sec:mscale_aggregation}) seems also to play an important role for the task. This is noticeable with CCPNet~\cite{Zhang2019CascadedCP} encouraging results given the use of a single geometric input (see Fig.~\ref{fig:qual_in_d},~\ref{fig:qual_in_e}). 
Note however that in addition to its lightweight design, the output of CCPNet is higher in resolution ($240\times144\times240$) which was proved to boost performance \cite{Zhang2019CascadedCP}. 
The performance of SISNet \cite{Cai2021SemanticSC} is also remarkable thanks to instance-wise completion at high resolution and iterative SSC.

On position awareness (Sec.~\ref{sec:position_awareness}) it seems to boost intra-class consistency together with inter-class distinction. 
For example 3DSKetch~\cite{Chen20203DSS} and PALNet~\cite{Li2020DepthBS}, both use position awareness and achieve high performances in indoor scenes with 3DSketch ranking second in mIoU among non-pre-trained methods and third overall (Tabs.~\ref{tab:results_datasets_perf} and \ref{tab:results_datasets_perf_pretrainSUNCG}),
visible in Figs.~\ref{fig:qual_in_a}, \ref{fig:qual_in_b}, \ref{fig:qual_in_c}. Similarly, S3CNet dominates performance in SemanticKITTI as already mentioned, which performance is noticeable in Figs. \ref{fig:qual_out_c}, \ref{fig:qual_out_d}.

An interesting observation is the high density of the completion even regarding the ground truth, visible in Figs.~\ref{fig:qual_out_b}, \ref{fig:qual_out_g}, \ref{fig:qual_out_h}. This relationship is studied in~\cite{Dai2019SGNNSG}, where sparsity is exploited by removing input data to impulse unknown space completion.

\paragraph{Synthetic data pre-training.} Pre-training on large SUNCG is a common workaround to improve performance on the smaller NYUv2 and NYUCAD datasets. Comparison between Tabs. \ref{tab:results_datasets_perf} and \ref{tab:results_datasets_perf_pretrainSUNCG} show that the technique always brings performance gains although the gap becomes less important for most recent methods (i.e. +5.8\% mIoU for SSCNet vs. +2.8\% mIoU for CCPNet). Since SUNCG is no longer legally available, this practice is less common in recent works.

\begin{table}
	\centering
	
	\newcommand{\metvariant}[1]{{~~~~#1}}
	\subfloat[{\scriptsize$60\times36\times60$} prediction (indoor)]{
		\scriptsize
		\def\mystrut{\rule{0pt}{1.5\normalbaselineskip}}
		\centering
		\rowcolors[]{2}{black!3}{white}
		\setlength{\tabcolsep}{0.01\linewidth}
		\begin{tabular}{rlcc} 
			\toprule
			& Method & Params (M) &  FLOPs (G)\\
			\midrule
			2017 & SSCNet~\cite{Song2017SemanticSC}\textsuperscript{a} &  0.93 & 163.8\\
			& VVNet~\cite{Guo2018ViewVolumeNF}\textsuperscript{a} & 0.69 & 119.2\\
			& ESSCNet~\cite{Zhang2018EfficientSS}\textsuperscript{a}  & 0.16 & 22\\
			& SATNet~\cite{Liu2018SeeAT}\textsuperscript{a} & 1.2 & 187.5\\
			2018 & DDRNet~\cite{Li2019RGBDBD} & 0.20 & 27.2\\
			
			& CCPNet~\cite{Zhang2019CascadedCP} & 0.08 & 6.5\\
			2020 & GRFNet~\cite{Liu20203DGR} & 0.82 & 713\\
			& PALNet~\cite{Li2020DepthBS} & 0.22 & 78.8\\
			& AIC-Net~\cite{Li2020AnisotropicCN} & 0.72 & 96.77\\
			& Chen et al.~\cite{Chen2020RealTimeSS} & \textbf{0.07} & \textbf{1.6}\\
			\bottomrule
			\multicolumn{4}{c}{\scriptsize \textsuperscript{a} Reported in~\cite{Zhang2019CascadedCP}.}
		\end{tabular}
		\label{tab:net_stats_indoor}
	}\\

	\subfloat[{\scriptsize$256\times32\times256$} prediction (outdoor)]{
		\scriptsize
		\def\mystrut{\rule{0pt}{1.5\normalbaselineskip}}
		\centering
		\rowcolors[]{2}{black!3}{white}
		\setlength{\tabcolsep}{0.01\linewidth}
		\begin{tabular}{rlcc} 
			\toprule
			& Method & Params (M) &  FLOPs (G)\\
			\midrule
			2017 & SSCNet~\cite{Song2017SemanticSC}\textsuperscript{b}         & 0.93 & 82.5\\  
			& \metvariant{SSCNet-full~\cite{Song2017SemanticSC}}\textsuperscript{b}      & 1.09 & 769.6\\
			2019 & TS3D~\cite{Garbade2019TwoS3}\textsuperscript{b}           & 43.77 & 2016.7\\
			& \metvariant{TS3D+DNet~\cite{Behley2019SemanticKITTIAD}}\textsuperscript{b}     & 51.31 & 847.1\\
			& \metvariant{TS3D+DNet+SATNet~\cite{Behley2019SemanticKITTIAD}}\textsuperscript{b}   & 50.57 & 905.2\\
			2020 & LMSCNet~\cite{Roldao2020LMSCNetLM} & \textbf{0.35} & \textbf{72.6} \\
			& JS3C-Net~\cite{Yan2020SparseSS} & 3.1 & - \\
			& Local-DIFs~\cite{Rist2020SemanticSC} & 9.9 & -\\
			\bottomrule
			\multicolumn{4}{c}{\scriptsize \textsuperscript{b} Reported in~\cite{Roldao2020LMSCNetLM}.}
		\end{tabular}
		\label{tab:net_stats_outdoor}
	}
	\caption{\textbf{Network statistics.} Number of parameters and FLOPs are reported per method, grouped by resolution output: {\scriptsize$60\times36\times60$} for typical indoor datasets~\cite{Silberman2012IndoorSA,Firman2016StructuredPO,Song2017SemanticSC} or {\scriptsize$256\times32\times256$} for outdoor dataset~\cite{Behley2019SemanticKITTIAD}.}
	\label{tab:net_stats}
\end{table}

\subsubsection{Network efficiency}\label{sec:eval_neteff}
In Tab.~\ref{tab:net_stats}, network parameters and floating-point operations (FLOPs) are listed -- where possible -- with separation of indoor and outdoor networks because they have different output resolutions.
Notice the extreme variations between networks, which scale from 1:144 in number of parameters and 1:1260 in FLOPs.
Chen \textit{et al.}~\cite{Chen2020RealTimeSS} and LMSCNet~\cite{Roldao2020LMSCNetLM} are by far the lightest networks with the fewest parameters and lower FLOPs, in indoor and outdoor settings respectively. They also account for the lower number of operations, which can -- though not necessarily~\cite{ma2018shufflenet} -- contribute to faster inference times.
Furthermore, the use of sparse convolutions~\cite{Graham20183DSS} is commonly applied as a strategy to reduce memory overhead in~\cite{Zhang2018EfficientSS,Dai2019SGNNSG, Yan2020SparseSS, Cheng2020S3CNet}.

\begin{figure*}
	\centering
	\scriptsize
	\setlength{\tabcolsep}{0.001\linewidth}
	\renewcommand{\arraystretch}{0.8}
	
	\subfloat{
		\begin{tabular}{ccccccc}	
			\multicolumn{7}{c}{NYUCAD \cite{Firman2016StructuredPO}} \\
			\midrule\\	
			& RGB & Depth & Visible Surface & Ground Truth & SSCNet~\cite{Song2017SemanticSC} & 3DSKetch~\cite{Chen20203DSS}  \\
			(a)~~ & 
			\includegraphics[width=0.22\columnwidth]{Figures/scene1_RGB.png} & 
			\includegraphics[width=0.22\columnwidth]{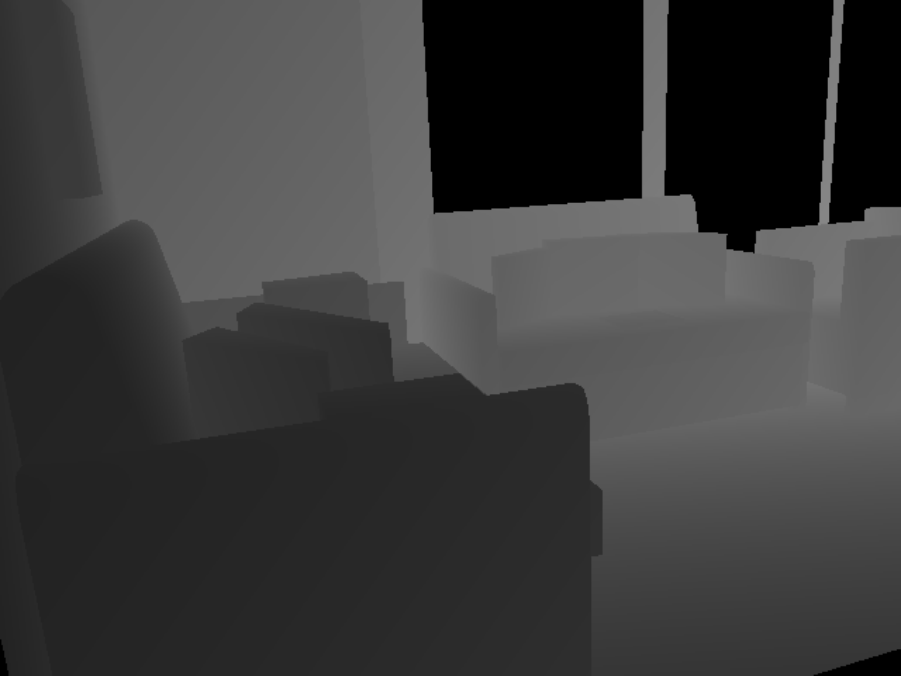} & 
			\includegraphics[width=0.22\columnwidth]{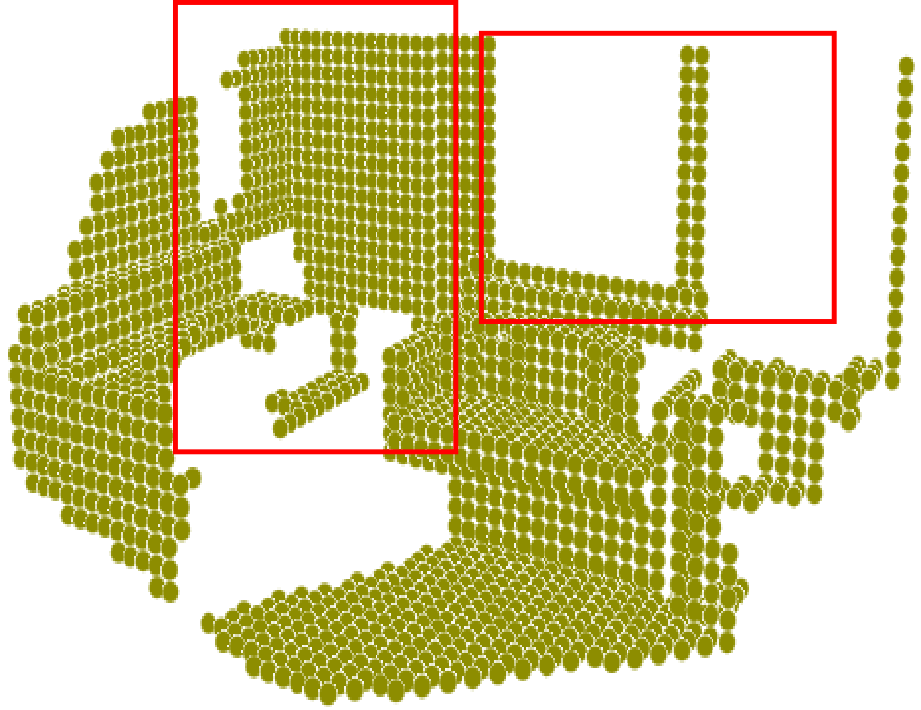} &
			\includegraphics[width=0.22\columnwidth]{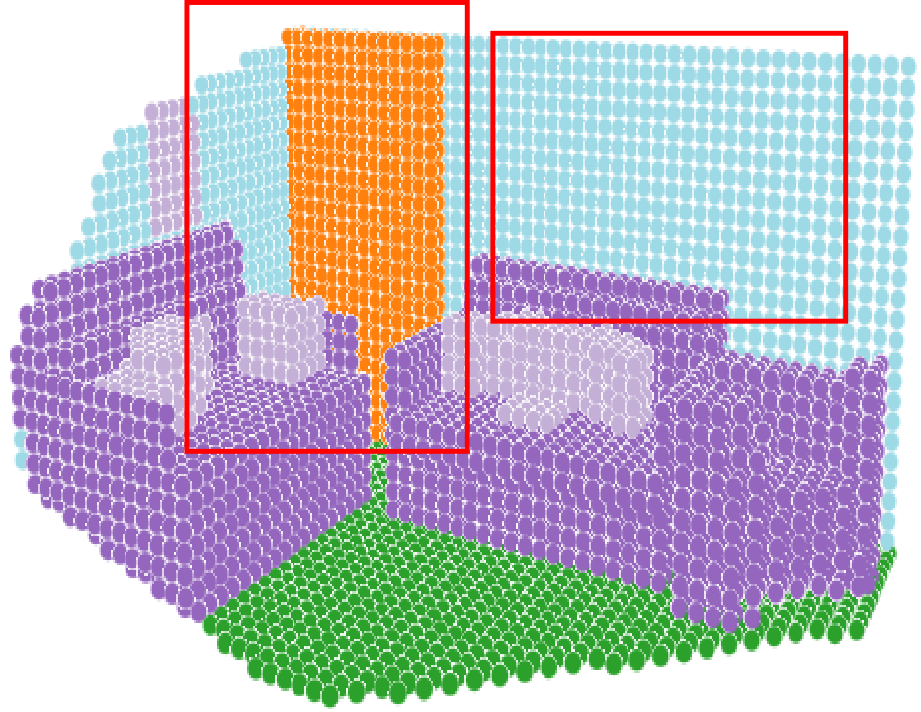} &
			\includegraphics[width=0.22\columnwidth]{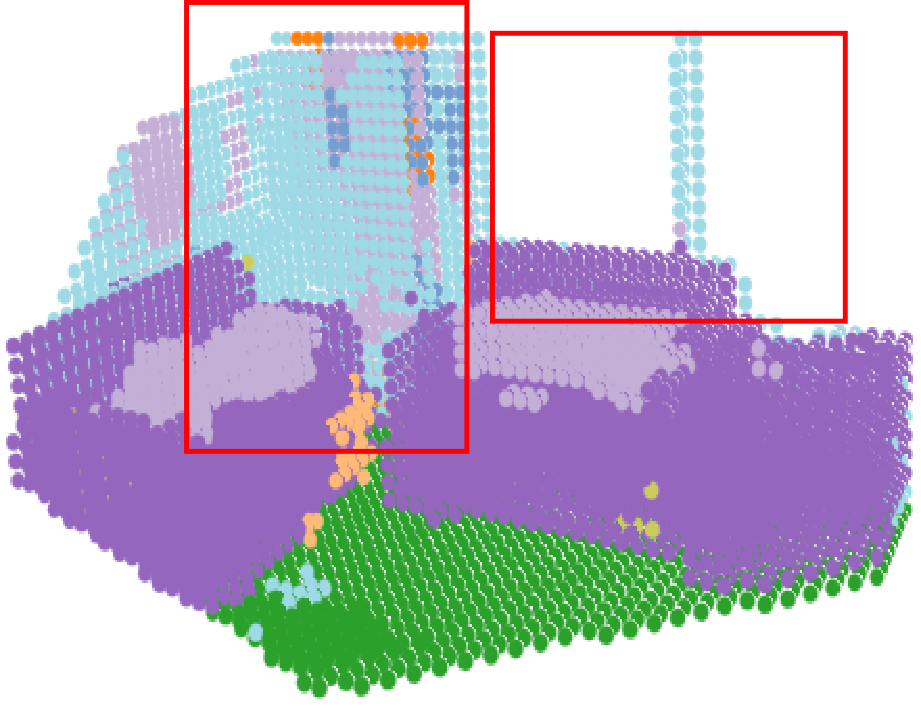} &
			\includegraphics[width=0.22\columnwidth]{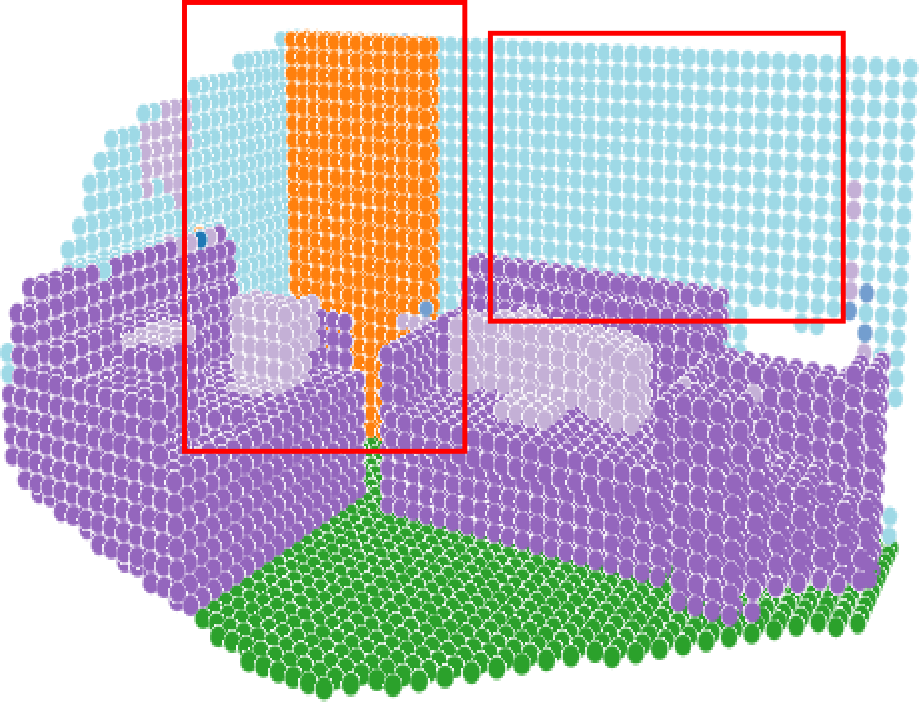} \\
			\label{fig:qual_in_a}
	\end{tabular}}
	
	\subfloat{
		\begin{tabular}{cccccccccc}		
			& RGB & Depth & Visible Surface & Ground Truth & DDRNet~\cite{Li2019RGBDBD} & PALNet~\cite{Li2020DepthBS} & 3DSKetch~\cite{Chen20203DSS} & AIC-Net~\cite{Li2020AnisotropicCN} & SISNet~\cite{Cai2021SemanticSC} \\
			(b)~~ & 
			\includegraphics[width=0.22\columnwidth]{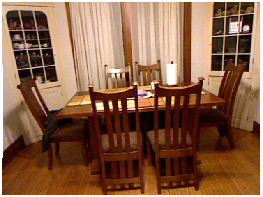} & 
			\includegraphics[width=0.22\columnwidth]{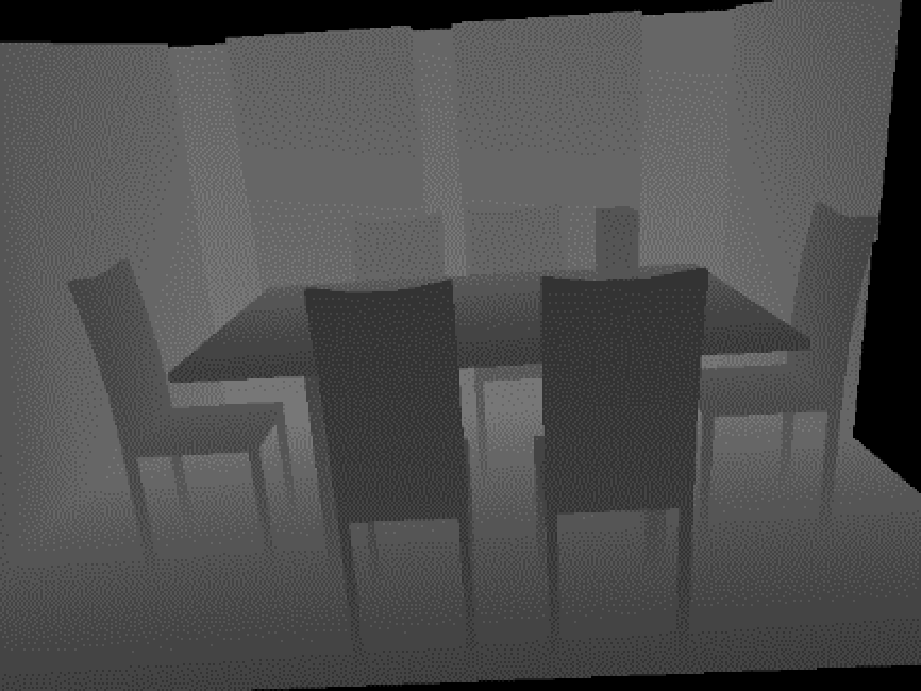} & 
			\includegraphics[width=0.22\columnwidth]{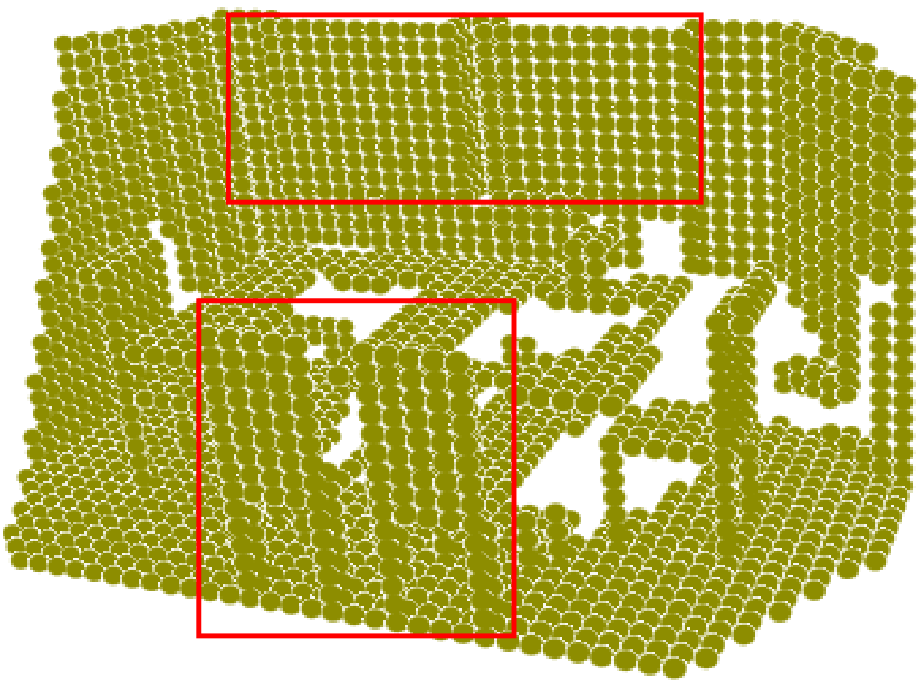} &
			\includegraphics[width=0.22\columnwidth]{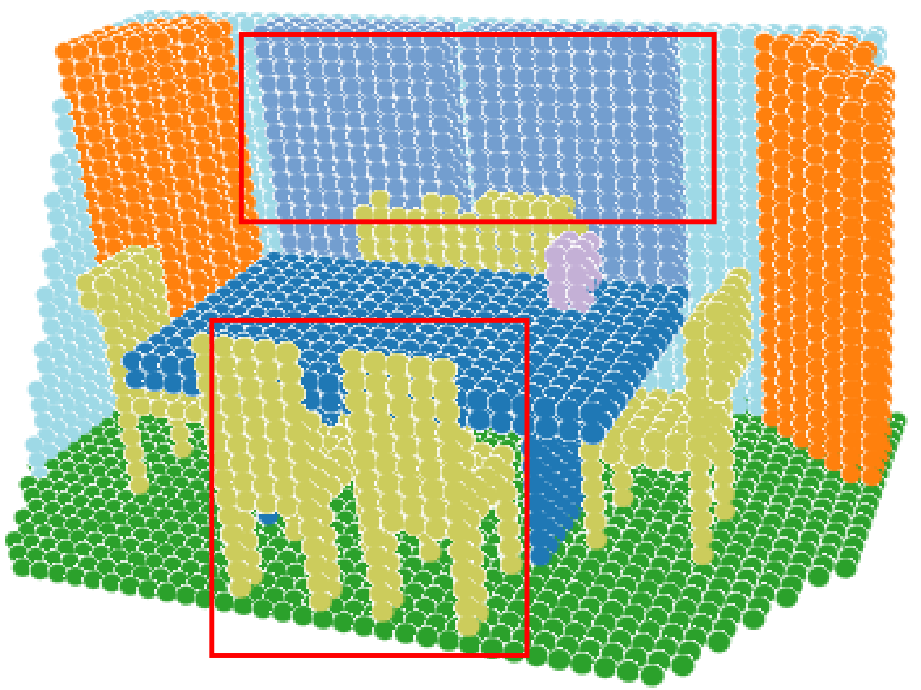} &
			\includegraphics[width=0.22\columnwidth]{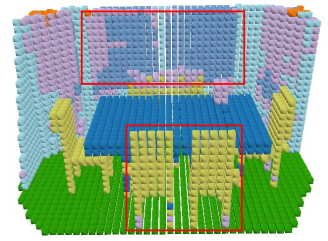} & 
			\includegraphics[width=0.22\columnwidth]{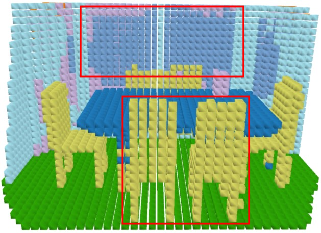} & 
			\includegraphics[width=0.22\columnwidth]{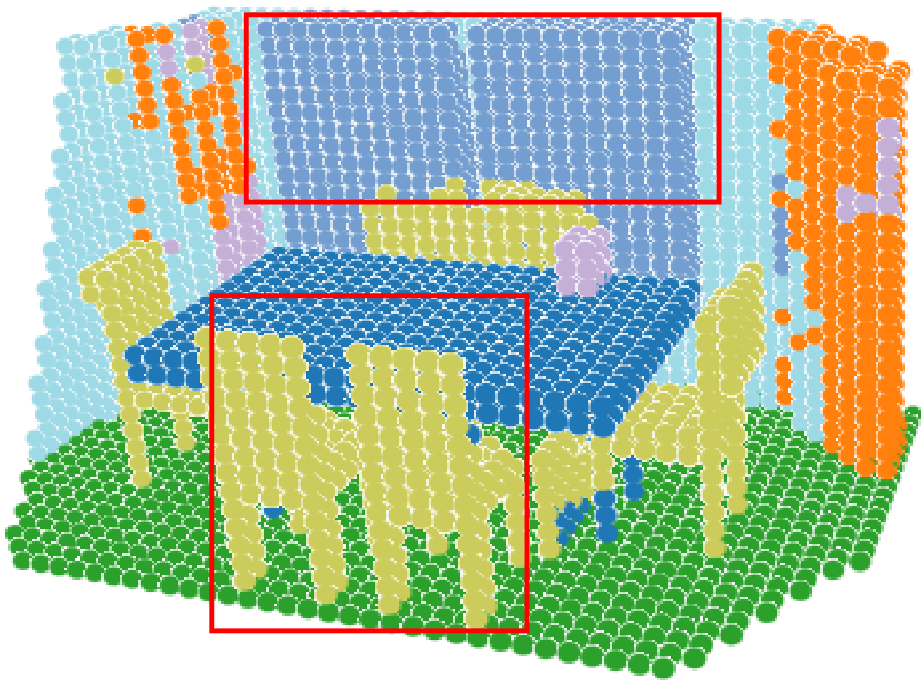} &
			\includegraphics[width=0.22\columnwidth]{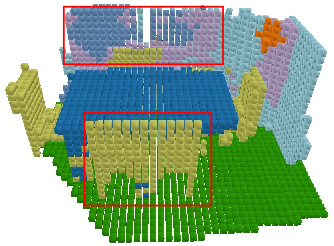} & 
			\includegraphics[width=0.22\columnwidth]{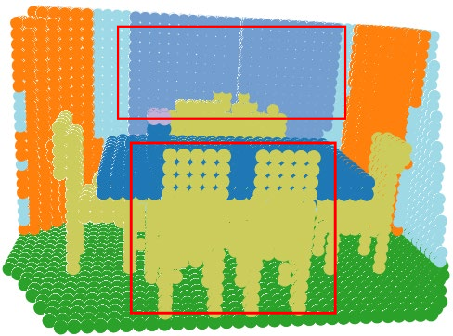}\\
			\label{fig:qual_in_b}
	\end{tabular}}

	\subfloat{
		\begin{tabular}{ccccccccc}		
			& RGB & Depth & Visible Surface & Ground Truth & SSCNet~\cite{Song2017SemanticSC} & DDRNet
			~\cite{Li2019RGBDBD} & PALNet~\cite{Li2020DepthBS} & 3DSKetch~\cite{Chen20203DSS}  \\
			(c)~~ & 
			\includegraphics[width=0.22\columnwidth]{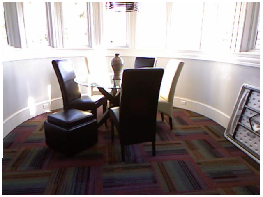} & 
			\includegraphics[width=0.22\columnwidth]{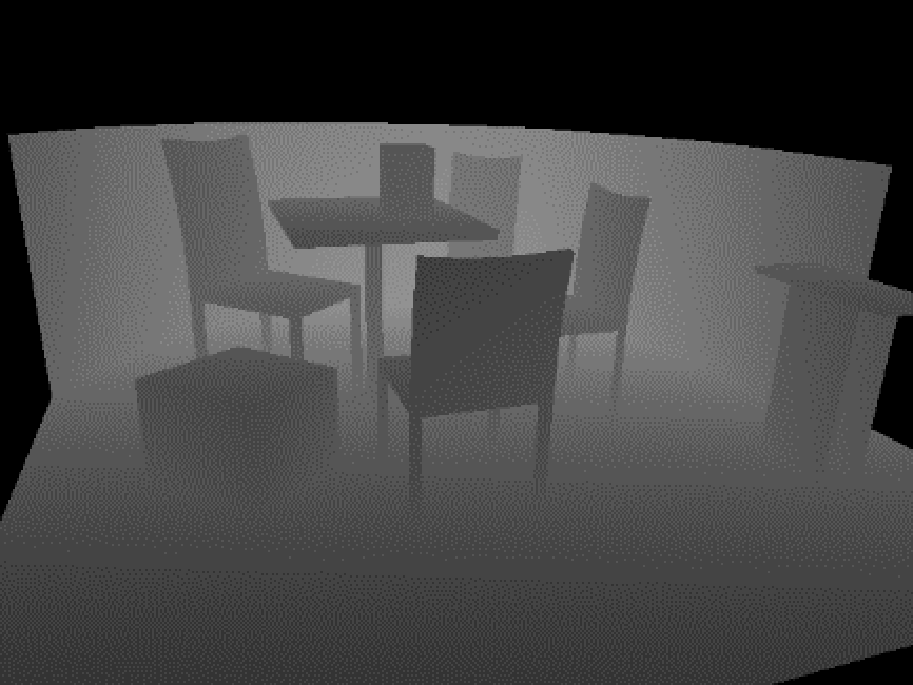} & 
			\includegraphics[width=0.22\columnwidth]{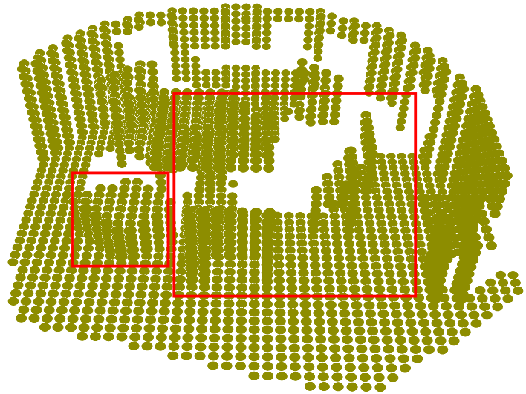} &
			\includegraphics[width=0.22\columnwidth]{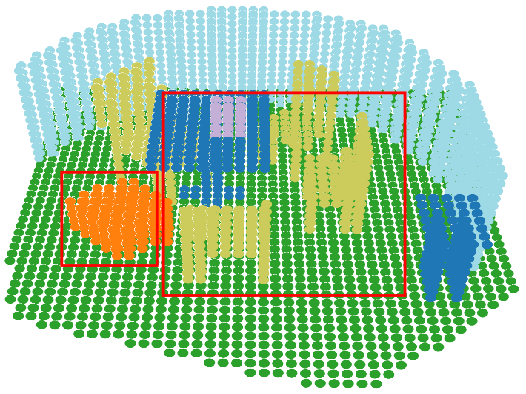} &
			\includegraphics[width=0.22\columnwidth]{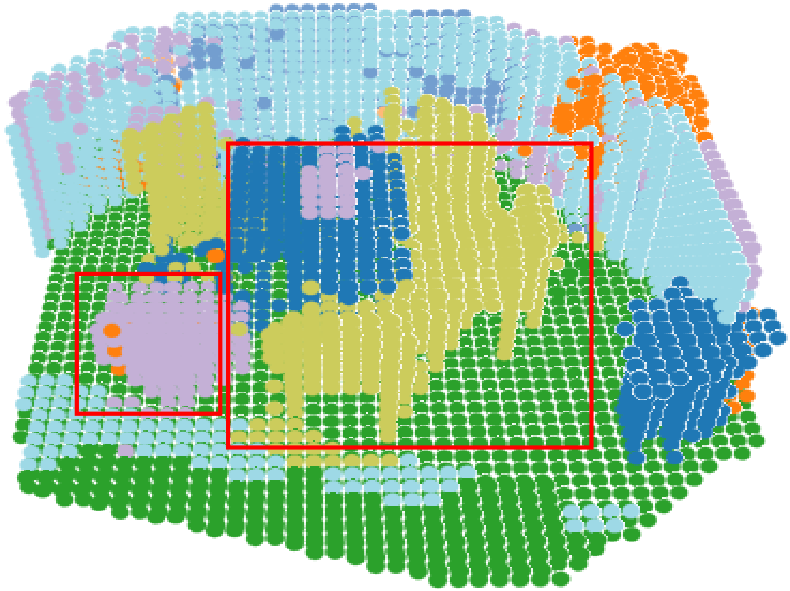} &
			\includegraphics[width=0.22\columnwidth]{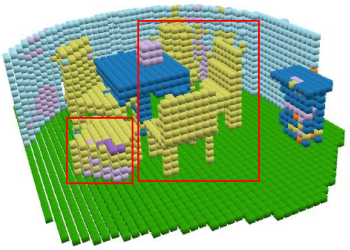} & 
			\includegraphics[width=0.22\columnwidth]{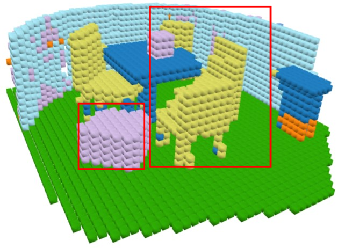} & 
			\includegraphics[width=0.22\columnwidth]{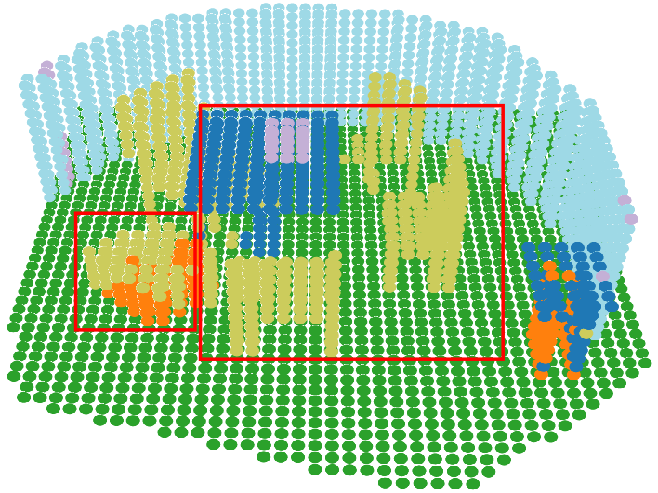} \\
			\label{fig:qual_in_c}
	\end{tabular}}

	\subfloat{
		\begin{tabular}{cccccccc}	
			\multicolumn{8}{c}{SUNCG \cite{Song2017SemanticSC}} \\
			\midrule\\			
			& Depth & Visible Surface & Ground Truth & SSCNet~\cite{Song2017SemanticSC} & VVNet~\cite{Guo2018ViewVolumeNF} & ESSCNet~\cite{Zhang2018EfficientSS} & CCPNet~\cite{Zhang2019CascadedCP} \\
			(d)~~ & 
			\includegraphics[width=0.21\columnwidth]{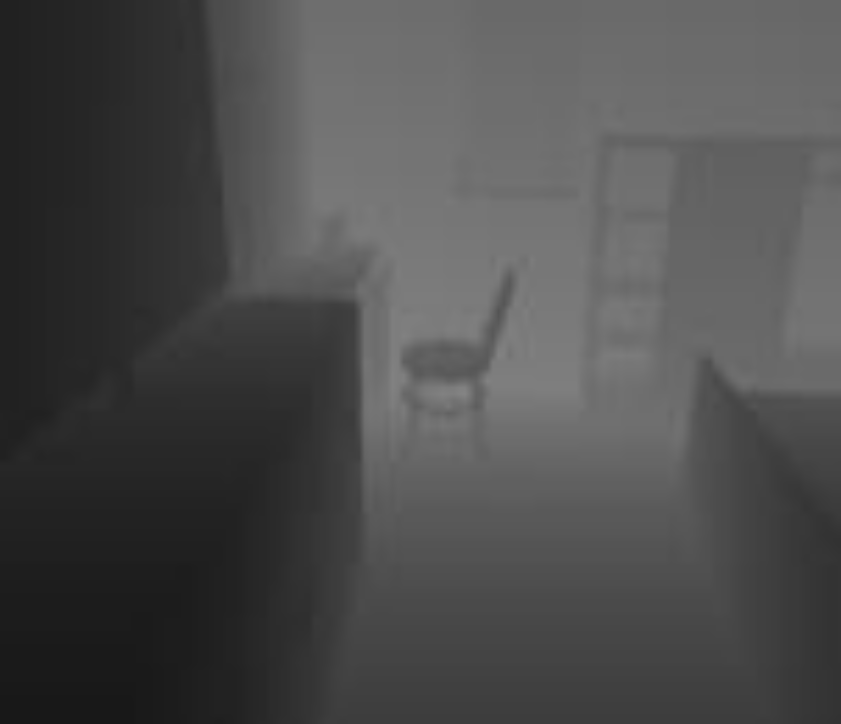} & 
			\includegraphics[width=0.21\columnwidth]{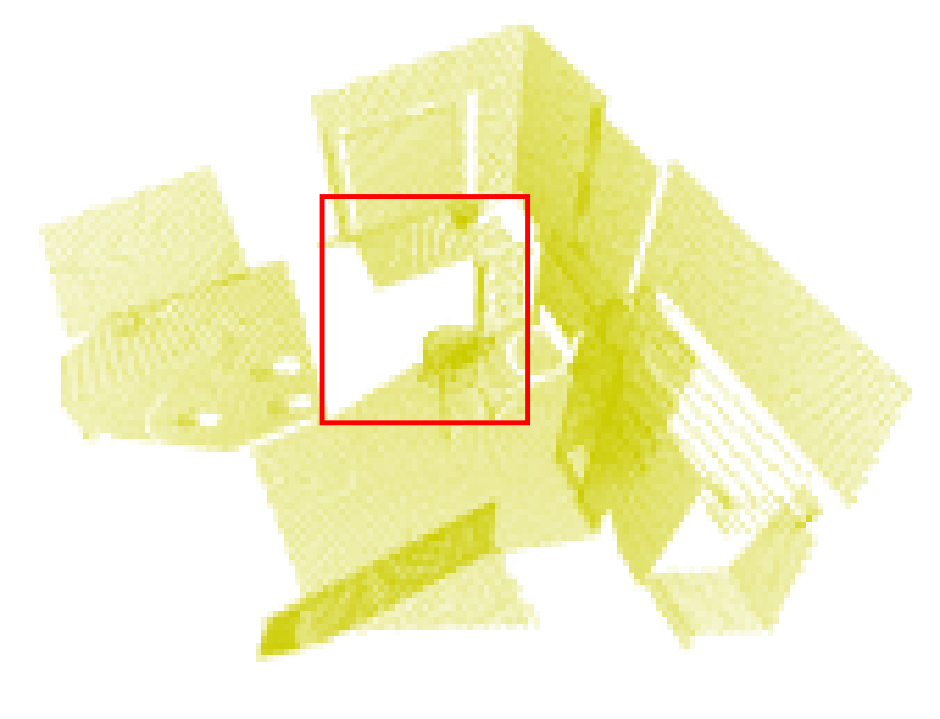} &
			\includegraphics[width=0.21\columnwidth]{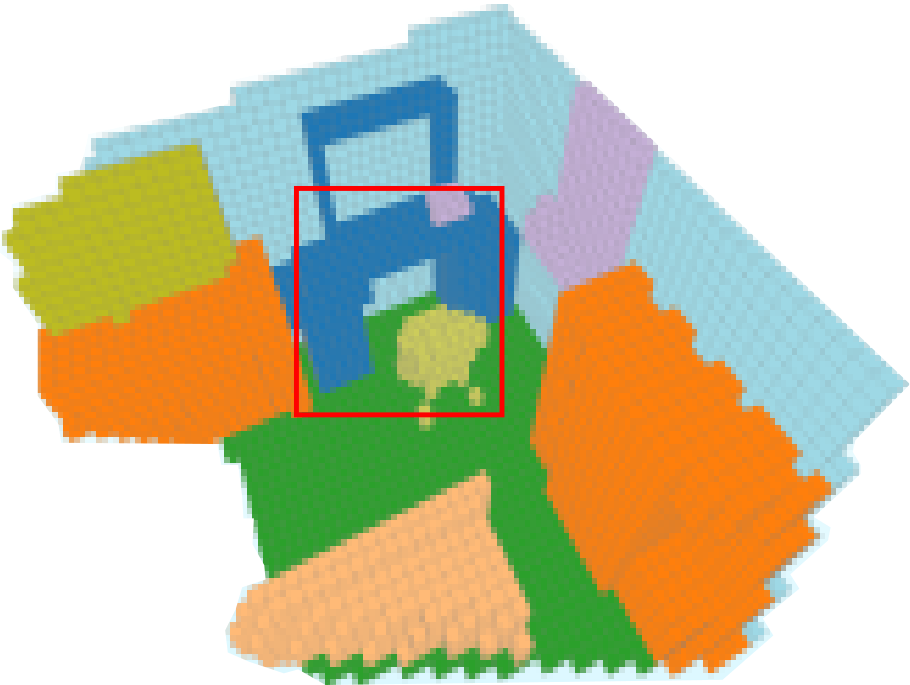} &
			\includegraphics[width=0.21\columnwidth]{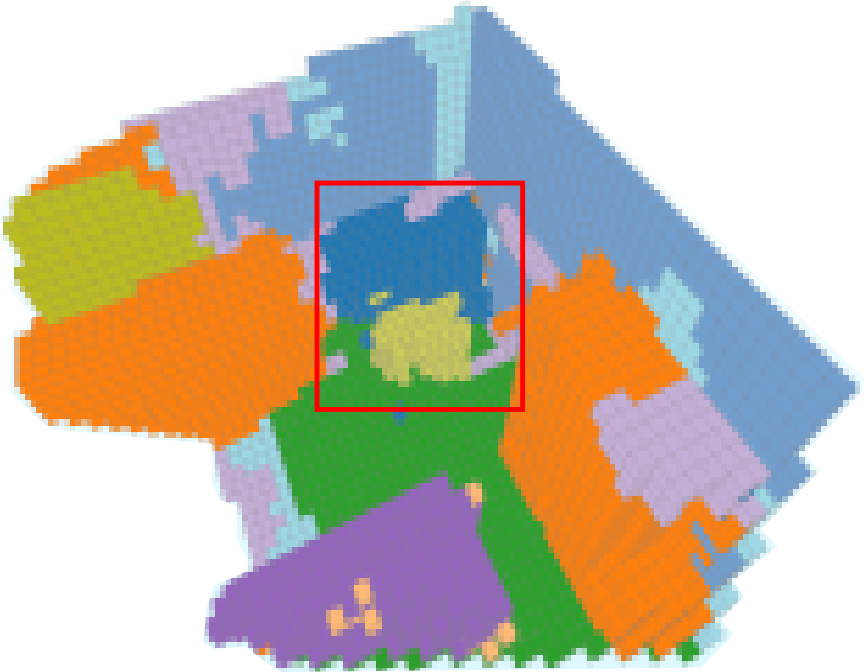} &
			\includegraphics[width=0.21\columnwidth]{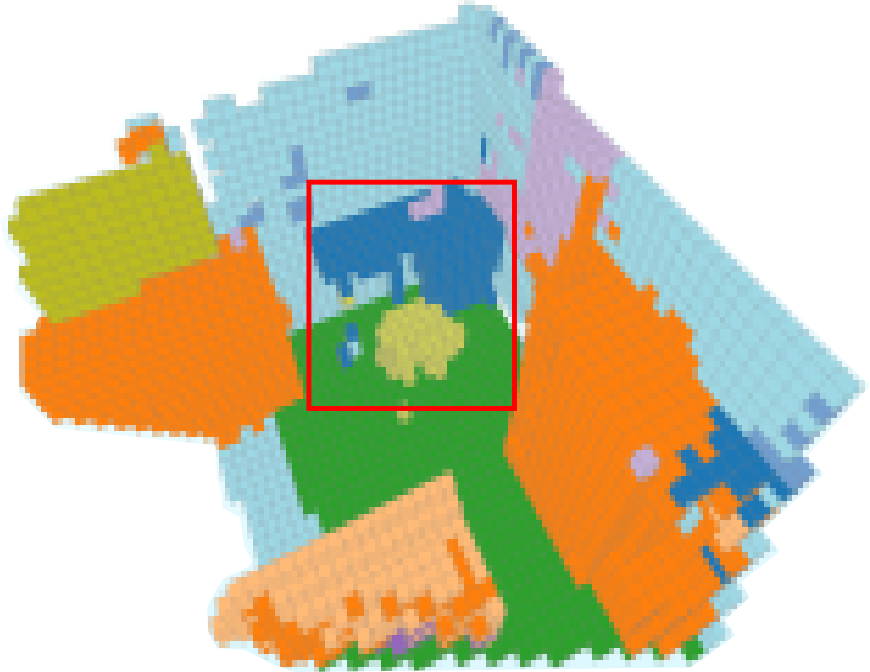} & 
			\includegraphics[width=0.21\columnwidth]{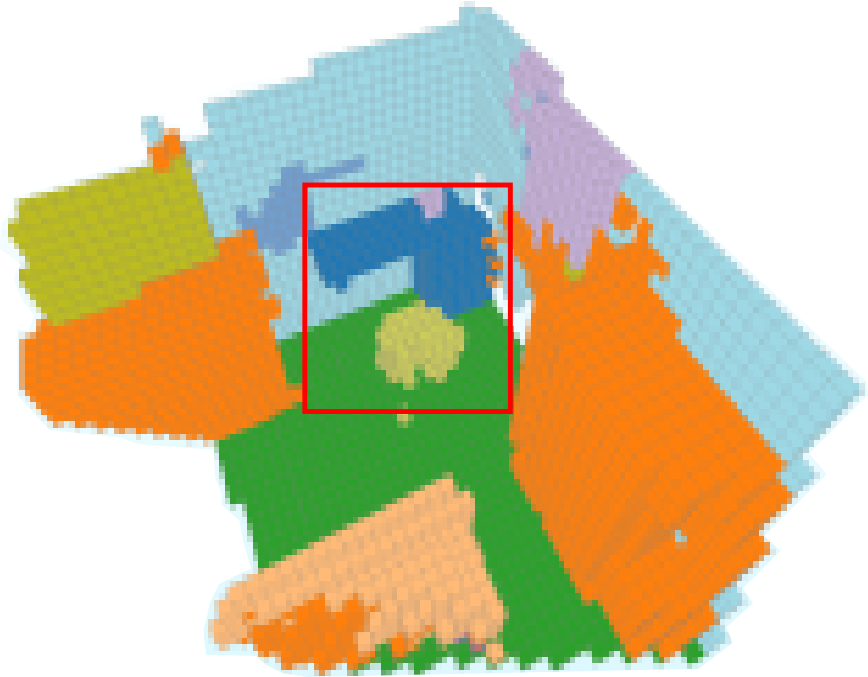} & 
			\includegraphics[width=0.21\columnwidth]{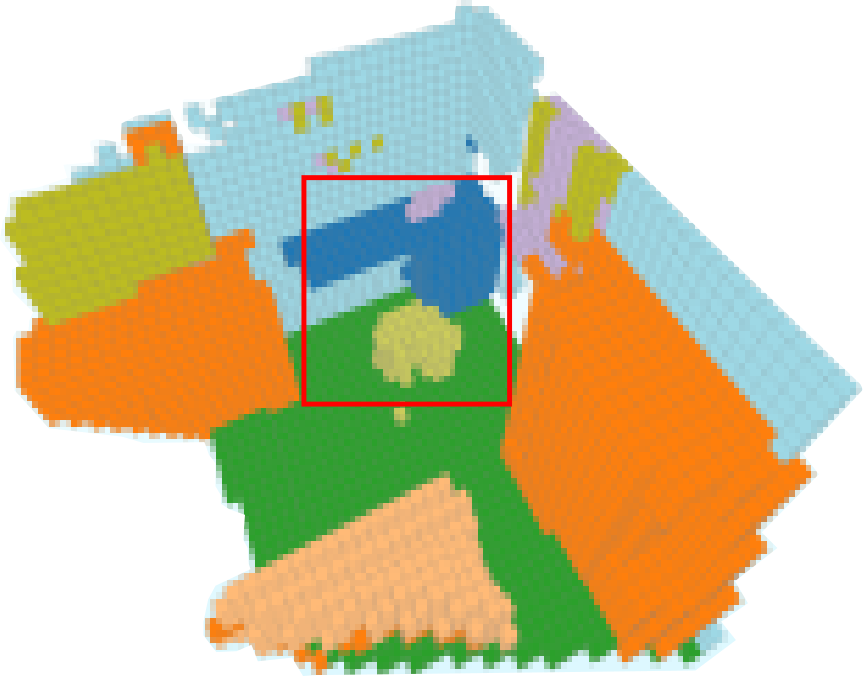} \\
			\label{fig:qual_in_d}
	\end{tabular}}

	\subfloat{
		\begin{tabular}{cccccccc}		
			& Depth & Visible Surface & Ground Truth & SSCNet~\cite{Song2017SemanticSC} & VVNet~\cite{Guo2018ViewVolumeNF} & ESSCNet~\cite{Zhang2018EfficientSS} & CCPNet~\cite{Zhang2019CascadedCP} \\
			(e)~~ & 
			\includegraphics[width=0.21\columnwidth]{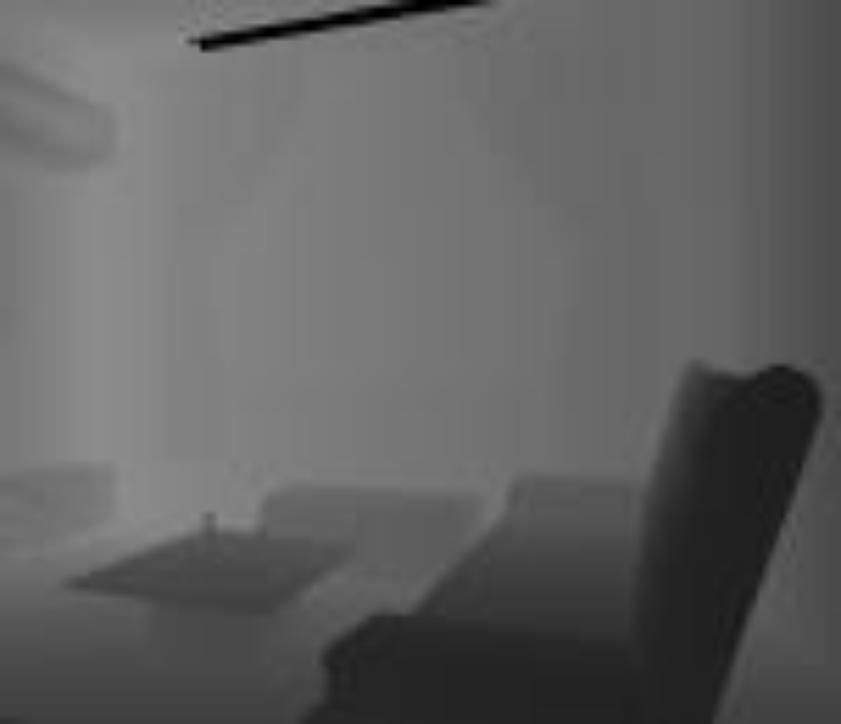} & 
			\includegraphics[width=0.21\columnwidth]{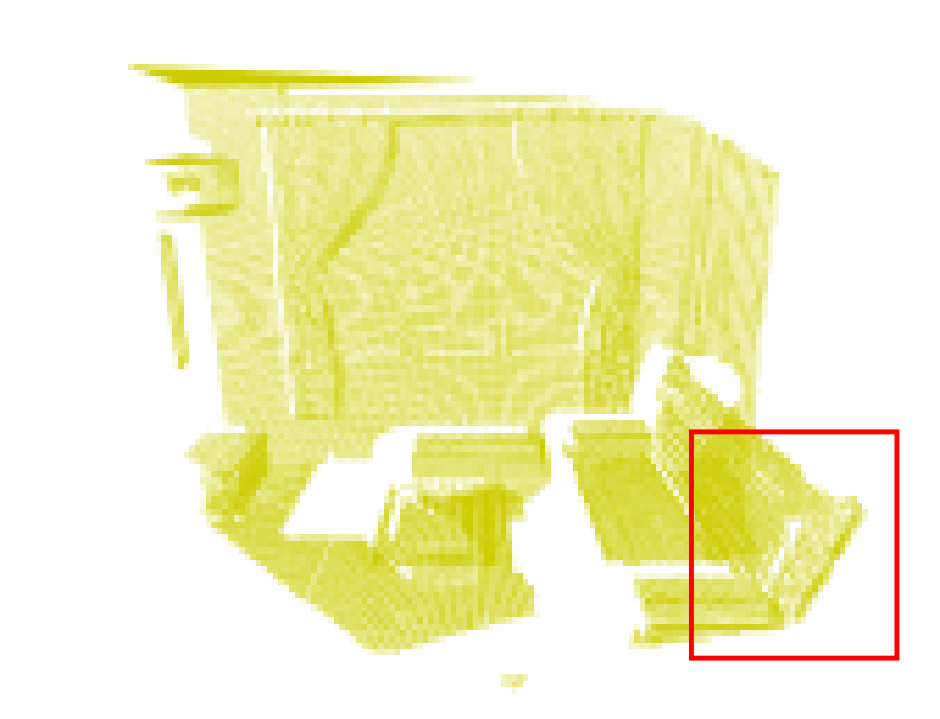} &
			\includegraphics[width=0.21\columnwidth]{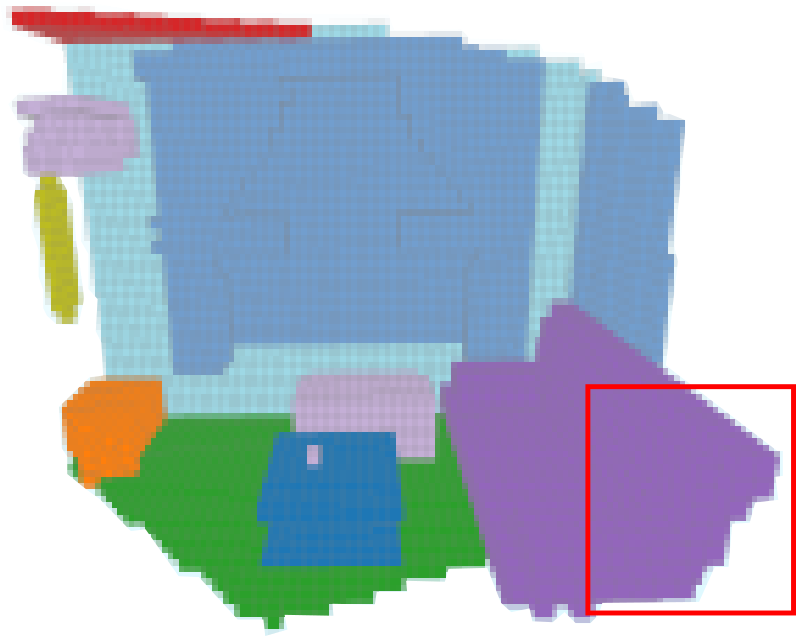} &
			\includegraphics[width=0.21\columnwidth]{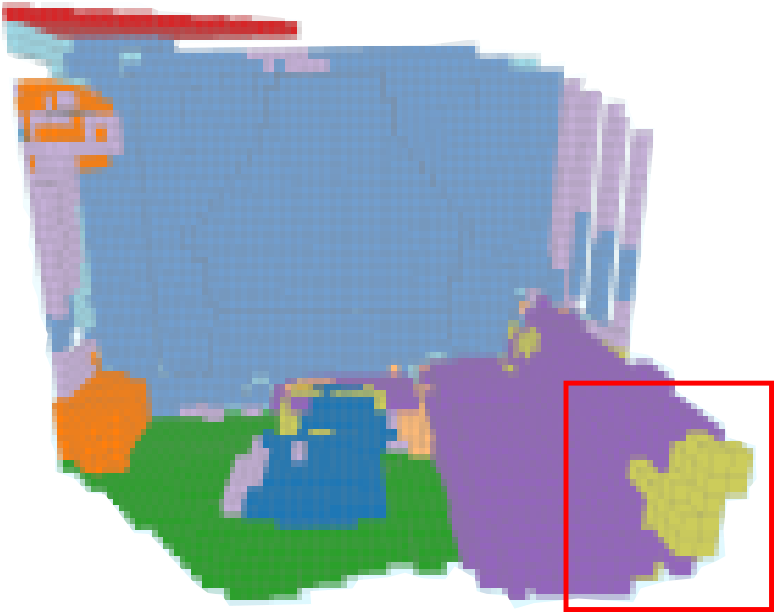} &
			\includegraphics[width=0.21\columnwidth]{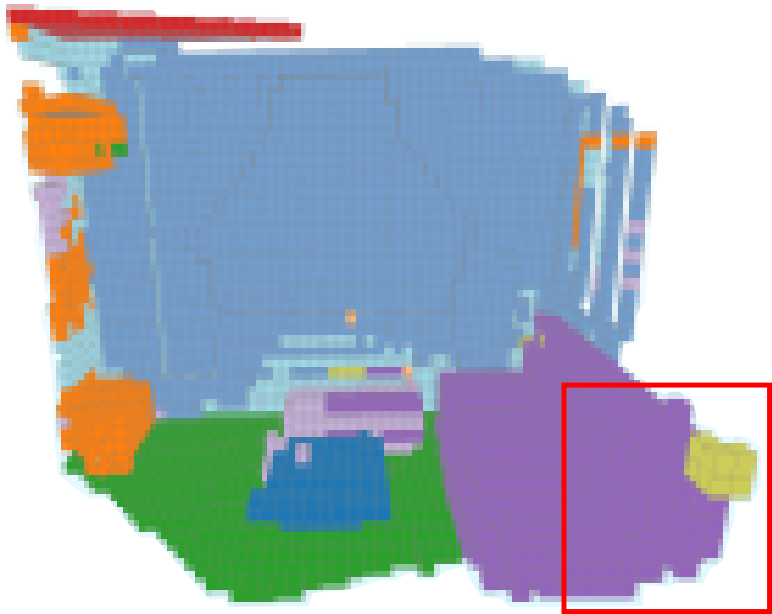} & 
			\includegraphics[width=0.21\columnwidth]{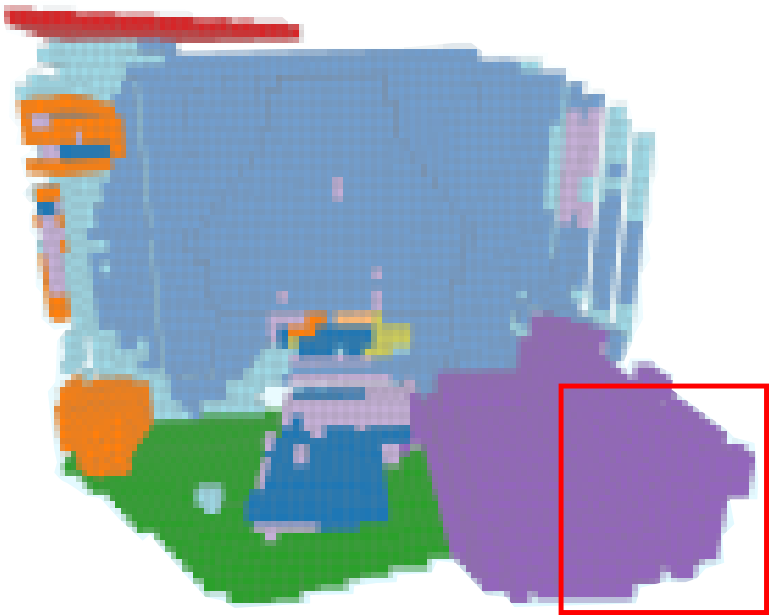} & 
			\includegraphics[width=0.21\columnwidth]{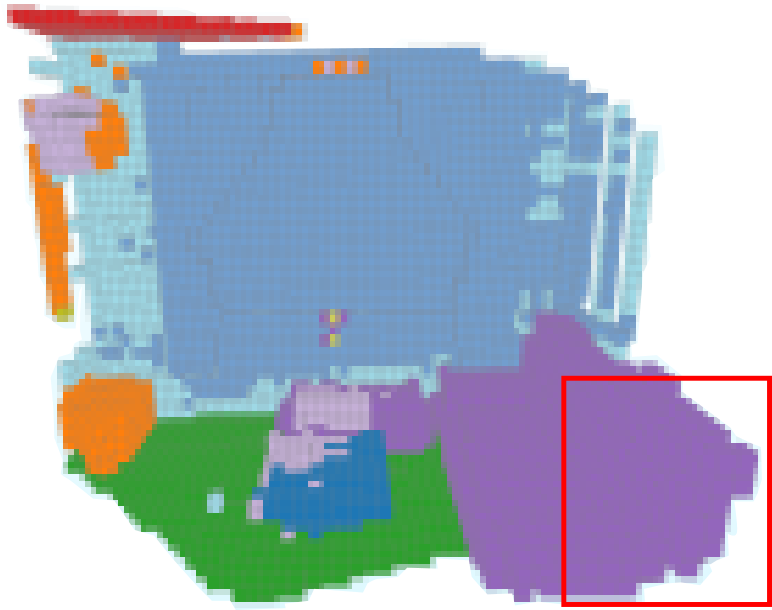} \\
			\\
			\multicolumn{8}{c}{\includegraphics[width=1.4\columnwidth]{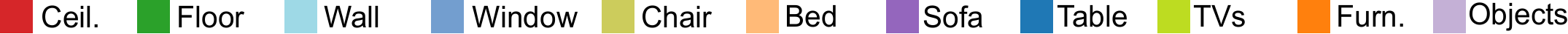}} \\
			\label{fig:qual_in_e}
	\end{tabular}}
	
	\caption{\textbf{Performance of indoor Semantic Scene Completion} on NYUCAD~\cite{Firman2016StructuredPO} and SUNCG~\cite{Song2017SemanticSC}. Methods with RGB  modalities (i.e. 3DSketch) enable detection of color salient objects as the highlighted door in row~\protect\subref{fig:qual_in_a}. Position awareness also contributes to better reconstruction consistency and inter-class distinction as seen in rows~\protect\subref{fig:qual_in_b},~\protect\subref{fig:qual_in_c} by PALNet~\cite{Li2020DepthBS} and 3DSketch~\cite{Chen20203DSS}. Furthermore, SISNet~\cite{Cai2021SemanticSC} overcomes all other methods through their scene-instance-scene loop architecture, seen in row~\protect\subref{fig:qual_in_b}. Multi-scale aggregation also improves reconstruction performance as seen on rows~\protect\subref{fig:qual_in_d},~\protect\subref{fig:qual_in_e}, where CCPNet~\cite{Zhang2019CascadedCP} achieves the best performance on SUNCG~\cite{Song2017SemanticSC}.}
	\label{fig:qualitative_indoor_1}
\end{figure*} 

\begin{figure*}
	\centering
	
	%	\resizebox{\textwidth*2}{!}{
	\scriptsize
	\setlength{\tabcolsep}{0.01\linewidth}
	\renewcommand{\arraystretch}{0.8}
	
	\subfloat{
		\begin{tabular}{ccccc}	
			\multicolumn{5}{c}{SemanticKITTI \cite{Behley2019SemanticKITTIAD}} \\
			\midrule\\		
			& Input & Ground Truth & SSCNet-full~\cite{Song2017SemanticSC, Roldao2020LMSCNetLM} & LMSCNet~\cite{Roldao2020LMSCNetLM}\\
			(a) & 
			\includegraphics[width=0.35\columnwidth]{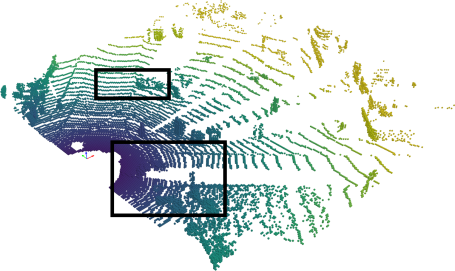} & 
			\includegraphics[width=0.35\columnwidth]{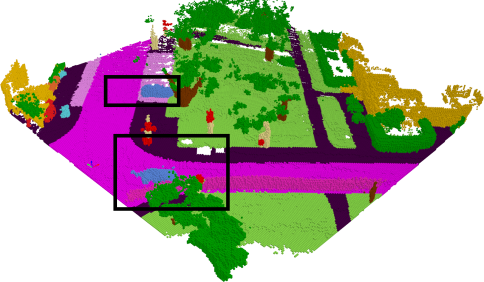} & 
			\includegraphics[width=0.35\columnwidth]{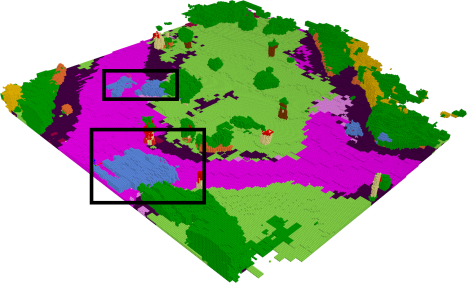} & 
			\includegraphics[width=0.35\columnwidth]{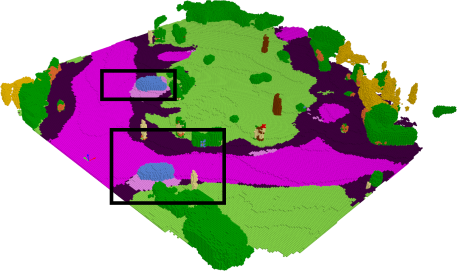}
			\label{fig:qual_out_a}
	\end{tabular}}
	\\
	\subfloat{
		\begin{tabular}{ccccc}		
			(b) & 
			\includegraphics[width=0.35\columnwidth]{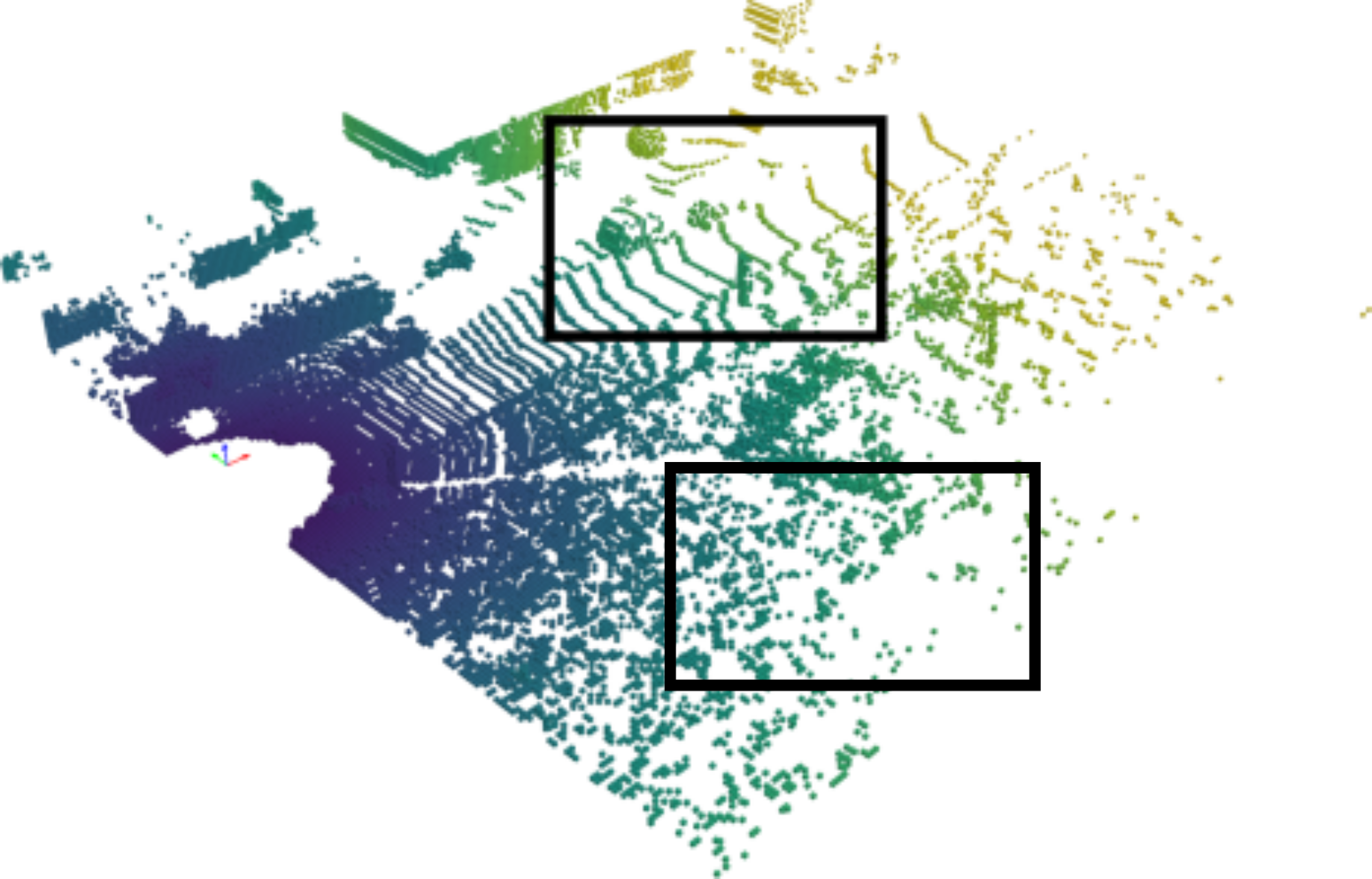} & 
			\includegraphics[width=0.35\columnwidth]{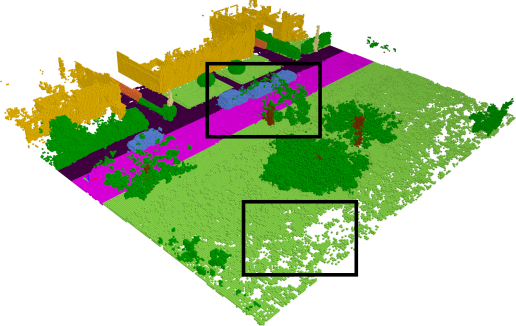} & 
			\includegraphics[width=0.35\columnwidth]{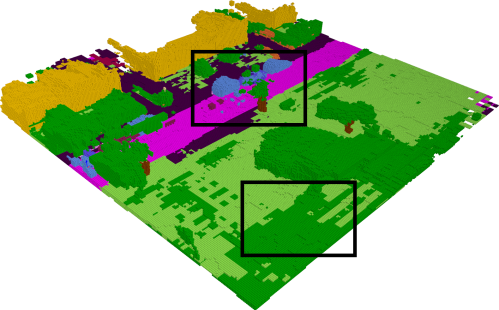} & 
			\includegraphics[width=0.35\columnwidth]{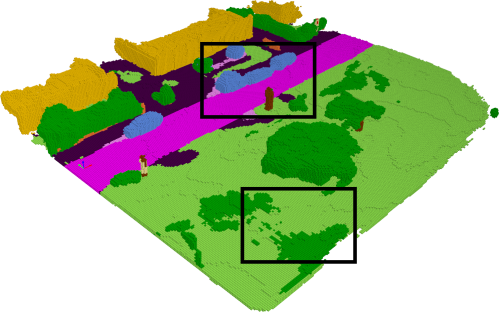}
			\label{fig:qual_out_b}
	\end{tabular}}

	\subfloat{
		\renewcommand{\arraystretch}{0.8}
		\begin{tabular}{cccc}		
			& RGB & Ground Truth & S3CNet~\cite{Cheng2020S3CNet} \\
			(c) & 
			\includegraphics[width=0.35\columnwidth]{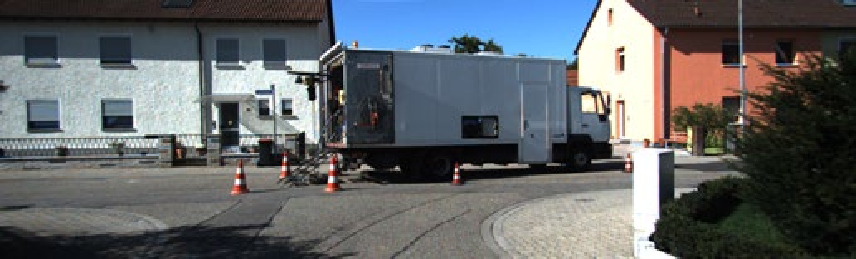} & 
			\includegraphics[width=0.35\columnwidth]{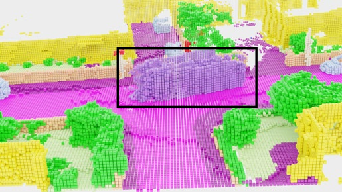} & 
			\includegraphics[width=0.35\columnwidth]{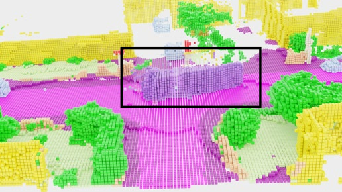}
			\label{fig:qual_out_c}
	\end{tabular}}
	\\
	\subfloat{
		\renewcommand{\arraystretch}{0.8}
		\begin{tabular}{cccc}		
			(d) & 
			\includegraphics[width=0.35\columnwidth]{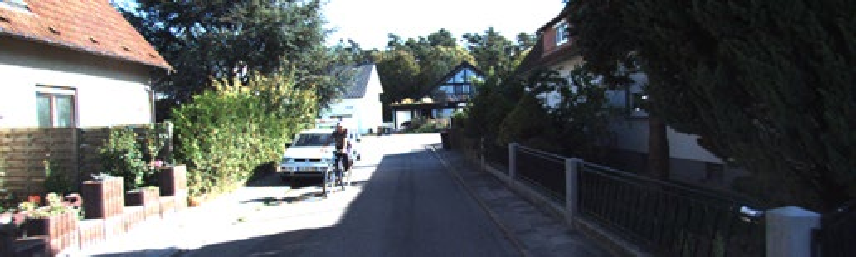} & 
			\includegraphics[width=0.35\columnwidth]{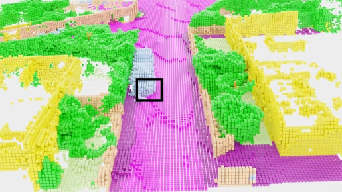} & 
			\includegraphics[width=0.35\columnwidth]{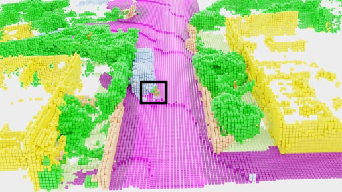}
			\label{fig:qual_out_d}
	\end{tabular}}

	\subfloat{
		\renewcommand{\arraystretch}{0.8}
		\begin{tabular}{ccccc}		
			& Input & Ground Truth & ESSCNet~\cite{Zhang2018EfficientSS} & JS3C-Net~\cite{Yan2020SparseSS}\\
			(e) & 
			\includegraphics[width=0.33\columnwidth]{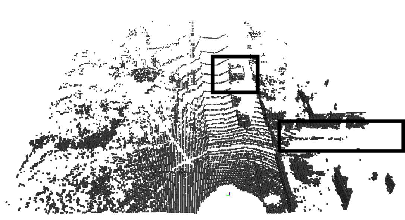} & 
			\includegraphics[width=0.33\columnwidth]{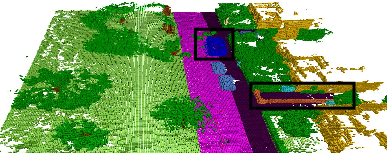} & 
			\includegraphics[width=0.33\columnwidth]{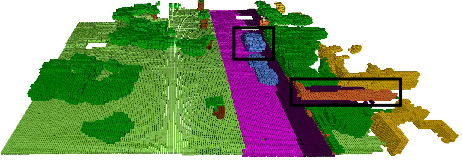} & 
			\includegraphics[width=0.33\columnwidth]{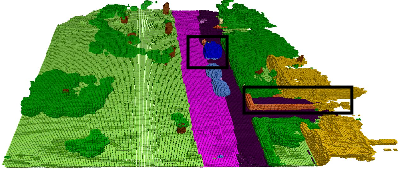}
			\label{fig:qual_out_e}
	\end{tabular}}
	\\
	\subfloat{
		\renewcommand{\arraystretch}{0.8}
		\begin{tabular}{ccccc}		
			(f) & 
			\includegraphics[width=0.32\columnwidth]{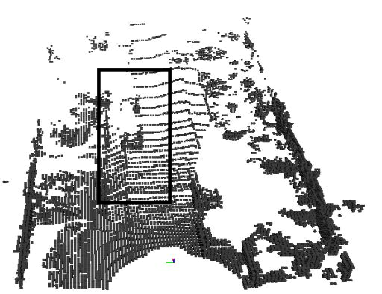} & 
			\includegraphics[width=0.32\columnwidth]{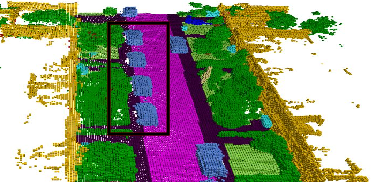} & 
			\includegraphics[width=0.32\columnwidth]{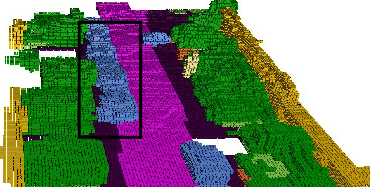} & 
			\includegraphics[width=0.32\columnwidth]{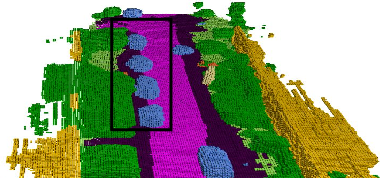}
			\label{fig:qual_out_f}
	\end{tabular}}

	\subfloat{
		\renewcommand{\arraystretch}{0.8}
		\begin{tabular}{ccccc}		
			& Ground Truth & Local-DIFs~\cite{Rist2020SemanticSC} & Input & Local-DIFs~\cite{Rist2020SemanticSC} (Larger extent)\\
			(g) & 
			\includegraphics[width=0.31\columnwidth]{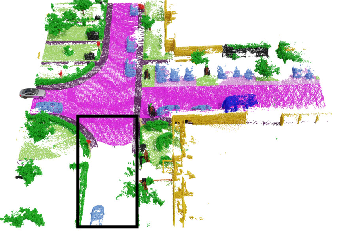} & 
			\includegraphics[width=0.31\columnwidth]{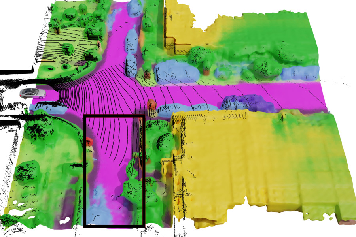} & 
			\includegraphics[width=0.31\columnwidth]{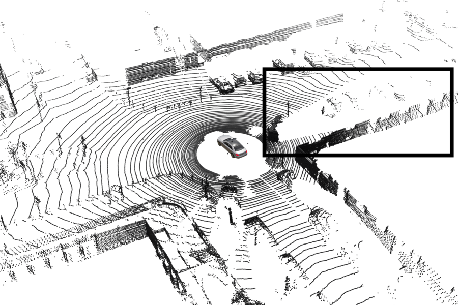} & 
			\includegraphics[width=0.31\columnwidth]{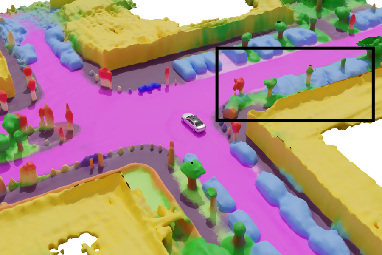}
			\label{fig:qual_out_g}
	\end{tabular}}
	\\
	\subfloat{
		\renewcommand{\arraystretch}{0.8}
		\begin{tabular}{ccccc}		
			(h) & 
			\includegraphics[width=0.32\columnwidth]{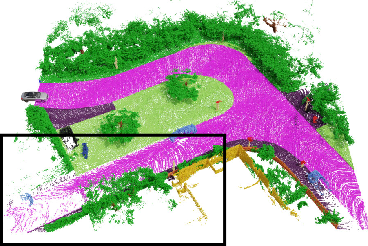} & 
			\includegraphics[width=0.32\columnwidth]{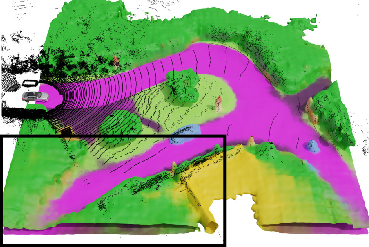} & 
			\includegraphics[width=0.32\columnwidth]{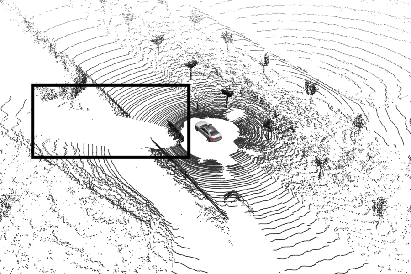} & 
			\includegraphics[width=0.32\columnwidth]{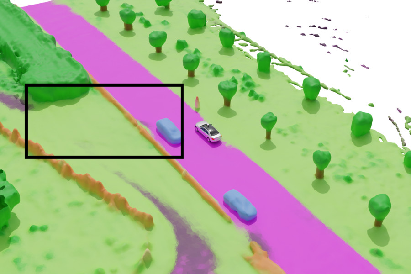} \\
			\multicolumn{5}{c}{\includegraphics[width=1.7\columnwidth]{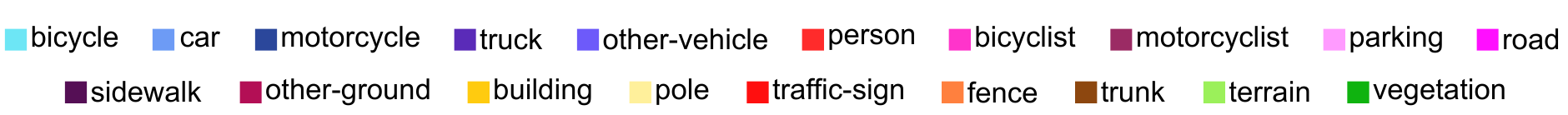}} \\	 
			\label{fig:qual_out_h}
	\end{tabular}}
	%	}
	
	\caption{\textbf{Performance of outdoor Semantic Scene Completion} on SemanticKITTI~\cite{Behley2019SemanticKITTIAD}. LMSCNet~\cite{Roldao2020LMSCNetLM} proposes a lightweight architecture with small performance decrease, rows~\protect\subref{fig:qual_out_a},~\protect\subref{fig:qual_out_b}. S3CNet~\cite{Cheng2020S3CNet} achieves SoA performance by their sparse bird's eye view and 3D f-TSDF feature encoders, rows~\protect\subref{fig:qual_out_c},~\protect\subref{fig:qual_out_d}. Two-stage JS3CNet~\cite{Yan2020SparseSS} performs point-wise semantic segmentation and semantic scene completion sequentially, enabling better completion as seen in rows~\protect\subref{fig:qual_out_e},~\protect\subref{fig:qual_out_f}. Finally, Local-DIFs~\cite{Rist2020SemanticSC} enables continuous surface prediction, thanks to deep implicit functions, which enable predictions of considerably larger spatial extent, rows~\protect\subref{fig:qual_out_g},~\protect\subref{fig:qual_out_h}.}
	\label{fig:qualitative_outdoor_1}
\end{figure*}

	\section{Discussion}\label{sec:discussion}

Despite growing interest, there are still major challenges to solve SSC as the best methods still perform poorly on real datasets (see Tabs. \ref{table:semantic_NYUv2}, \ref{table:semantic_kitti_semantics}). In this section, we wish to highlight important remaining issues and provide future research directions.

\paragraph{Best practices for SSC.}

Among the various viable choices for SSC, some were proven highly beneficial. For instance,
contextual aggregation (Sec.~\ref{sec:mscale_aggregation}) improves the information flow. In that sense, fusion of low and high level features with different receptive fields provides a contextual signal for the network, benefiting both the identification of fine-grained details and the overall scene understanding.
Like for semantic segmentation, features aggregation significantly boosts performance, typically with UNet~\cite{Roldao2020LMSCNetLM,Zhang2018EfficientSS,Wang2017OCNNOC,Cheng2020S3CNet,Yan2020SparseSS,Dourado2019EdgeNetSS}, Cascade Pyramid~\cite{Zhang2019CascadedCP} or Feature Aggregation Modules~\cite{Li2020AttentionbasedMF,Li2020DepthBS,Liu20203DGR,Li2020AnisotropicCN}.
Further geometrical cues often boost SSC, whether if it is multiple geometrical representations (e.g. depth + voxel, Sec.~\ref{sec:input_encoding}) or boundaries (e.g. edges, Sec.~\ref{sec:position_awareness}).\\
Multi-task and auxiliary tasks can also bring a performance boost. This is commonly done in the form of separated semantic segmentation~\cite{Yan2020SparseSS} or instance completion~\cite{Cai2021SemanticSC} to improve input to SSC.
Sketch supervision \cite{Chen20203DSS} -- which comes at virtually no cost -- was shown particularly helpful to boost indoor performance but might be less beneficial for outdoor scenarios given larger semantic geometrical variance (i.e. vegetation).
On networks, sparse convolutions \cite{Yan2020SparseSS,Cheng2020S3CNet,Dai2019SGNNSG} can reduce memory needs and therefore enable higher resolution, although they must be combined with dense convolutions. Similar for dilated convolutions that have been used in the wide majority of SSC works \cite{Song2017SemanticSC,Guo2018ViewVolumeNF,Garbade2019TwoS3,Liu2018SeeAT,Wang2019ForkNetMV,Dourado2019EdgeNetSS,Dourado2020SemanticSC,Zhang2019CascadedCP,Li2020AttentionbasedMF,Chen20203DSS,Li2020DepthBS,Roldao2020LMSCNetLM,Wu2020SCFusionRI}.\\
On training, \cite{Rist2020SemanticSC} shows that free space supervision close to the geometry can provide sharper inference, and we believe adversarial training (Sec.~\ref{sec:training_strategies}) is key to cope with the ground truth ambiguities. Another way to have an additional signal is the use of position aware losses \cite{Li2020DepthBS} which provides additional spatial supervision and was shown to bring performance improvements in both indoor \cite{Li2020DepthBS} and outdoor scenarios \cite{Cheng2020S3CNet}.
On evaluation, we encourage authors to evaluate on both indoor \textit{and} outdoor datasets which exhibit different challenges.
Finally, for real-time applications, more works like~ \cite{Roldao2020LMSCNetLM,Chen2020RealTimeSS} should account for lightweight and fast inference architectures (Sec.~\ref{sec:eval_neteff}).

\paragraph{Supervision bias.} An important challenge for completion results from the big imbalance ratio between free and occupied space (9:1 in both NYUv2~\cite{Song2017SemanticSC, Silberman2012IndoorSA} and SemanticKITTI~\cite{Behley2019SemanticKITTIAD}) which biases the networks towards free space predictions. To deal with this problem, random undersampling of the major free class is often applied~\cite{Song2017SemanticSC}
to reach an acceptable 2:1 ratio. The strategy reportedly improves completion performance (i.e. $+4\%$ IoU~\cite{Song2017SemanticSC}) and is widely employed~\cite{Zhang2018EfficientSS, Garbade2019TwoS3, Liu2018SeeAT, Dourado2019EdgeNetSS, Zhang2019CascadedCP}. Similarly, the loss can be balanced to favor occupy predictions~\cite{Li2019RGBDBD, Liu20203DGR}. Again, few works like~\cite{Rist2020SemanticSC} efficiently benefit from free space information.

\paragraph{Semantic class balancing.}
Imbalance is also present in the semantic labels, especially in outdoor datasets, where there is a prevalence of road or vegetation (see Fig.~\ref{fig:teaser} and~\ref{fig:dset_frequenciess}). Class-balancing can be applied to mitigate imbalanced distribution, usually weighting each class according to the inverse of its frequency~\cite{Roldao2020LMSCNetLM}, though prediction of under-represented classes still suffers (e.g. pedestrian or motorcycle in~\cite{Behley2019SemanticKITTIAD}). This may have a catastrophic impact on robotics applications.
An approach worth mentioning is S3CNet~\cite{Cheng2020S3CNet}, where combined weighted cross entropy and position aware loss (cf.~Sec.~\ref{sec:losses}) achieve impressive improvements in under-represented classes of SemanticKITTI. 
We believe SSC could benefit from smarter balancing strategies.

\begin{table*}
	\scriptsize
	\setlength{\tabcolsep}{0.0035\linewidth}
	\newcommand{\nclass}[1]{\scriptsize{#1}}
	
	\newcommand{\ext}[1]{\tiny{#1}}
	
	\def\mystrut{\rule{0pt}{1.5\normalbaselineskip}}
	\centering
	\rowcolors[]{2}{black!3}{white}
	
	\begin{tabular}{l | c c c c c c c c}
		\toprule
		Dataset & Year & Type & Nature & Data & 3D Sensor & \# Classes & Extension & \#Sequences \\ 
		\midrule		
		IQMulus~\cite{Vallet2015TerraMobilitaiQmulusUP} & 2015 & Real-wolrd & Outdoor & $\rightarrow$ Points & Lidar & \nclass{-} & \ext{Sparse input scene subsampling} & -  \\
		Semantic3D~\cite{Hackel2017Semantic3DnetAN} & 2017 & Real-world & Outdoor &  $\rightarrow$ Points & 3D Scanner & 8 & \ext{Sparse input scene subsampling} & 30  \\
		Paris-Lille-3D~\cite{Roynard2018ParisLille3DAL} & 2018 & Real-world & Outdoor & $\rightarrow$ Points & Lidar-32 & 50 & \ext{Sparse input scene subsampling} & 4 \\
		Replica~\cite{Straub2019TheRD} & 2019 & Real-world\tiny{$^{\dag}$} & Indoor &  $\rightarrow$ 3D Mesh & RGB-D & \nclass{88} & \ext{Sparse input from virtual RGB-D} & 35  \\
		nuScenes~\cite{Caesar2019nuScenesAM} & 2020 & Real-world & Outdoor & Points/RGB $\rightarrow$ & Lidar-32 & \nclass{32} & \ext{Dense scenes registration} & 1000  \\
		Toronto-3D~\cite{Tan2020Toronto3DAL}  & 2020 & Real-world & Outdosor & $\rightarrow$ Points & Lidar-32 & 8 & \ext{Sparse input scene subsampling} & 4   \\
		3D-FRONT~\cite{Fu20203DFRONT3F} & 2020 & Synthetic & Indoor & $\rightarrow$ Mesh & - & \nclass{-} & \ext{Sparse input from virtual RGB-D} & -  \\	
		\toprule

	\end{tabular}\\
	{\scriptsize $^{\dag}$ Synthetically augmented.}

	\caption{\textbf{SSC-extendable datasets}. To promote research on SSC we highlight that existing 3D semantic datasets could be extended for SSC, at the cost of processing work (cf. col. `Extension'). While some extensions could be obtained with little processing (e.g. Replica~\cite{Straub2019TheRD}, 3D-Front~\cite{Fu20203DFRONT3F}), others are significantly more complex (e.g. nuScenes~\cite{Caesar2019nuScenesAM}).}
	\label{table:SSC_extendable_datasets}
\end{table*}

\paragraph{Object motion.} 
As mentioned in Sec.~\ref{sec:datasets}, real-world ground truth is obtained by the \textit{rigid} registration of contiguous frames. While this corrects for ego-motion, it doesn't account for scene motion and moving objects produce temporal tubes in the ground truth, as visible in SemanticKITTI~\cite{Behley2019SemanticKITTIAD} (Fig.~\ref{fig:dset_accumulation}).
As such, to maximize performance, the SSC network must additionally predict the motion of any moving objects.\\
To evaluate the influence of such imperfections for SSC, some works reconstruct target scenes by 
accounting only for a few future scans~\cite{Yan2020SparseSS, Rist2020SemanticSC}.
Results show marginal completion improvement from the application of such a strategy.
An alternative proposal~\cite{KimK20a}, is to remove dynamic objects from the detection of spatial singularities after frames registration.
On the challenging SemanticKITTI~\cite{Behley2019SemanticKITTIAD}, because there are few insights to classify dynamic objects, all methods tend to predict vehicles as stationary (cf. Fig.~\ref{fig:qualitative_outdoor_1}) -- producing appealing results but being punished by dataset metrics. This obviously results from the dataset bias, given the abundance of parked vehicles.

The introduction of larger synthetic datasets~\cite{Dosovitskiy2017CARLAAO, Ros2016TheSD} could be an interesting solution to fight ground truth inaccuracies.

\paragraph{Datasets extendable for SSC.}

Because semantic labeling is the most complex and costly, we denote that a large amount of existing 3D semantics datasets \cite{Vallet2015TerraMobilitaiQmulusUP,Hackel2017Semantic3DnetAN,Roynard2018ParisLille3DAL,Straub2019TheRD,Caesar2019nuScenesAM,Tan2020Toronto3DAL,Fu20203DFRONT3F} could also be extended to SSC at the cost of some processing effort. A selective list of these \textit{SSC-extendable} datasets is in Tab. \ref{table:SSC_extendable_datasets} and we believe that their use should be encouraged to serve the interest of research on SSC.
Interestingly, most need little processing for SSC (e.g. sparse input generation from 3D meshes or point clouds, virtual sensor configurations)~\cite{Vallet2015TerraMobilitaiQmulusUP,Hackel2017Semantic3DnetAN,Roynard2018ParisLille3DAL,Straub2019TheRD,Tan2020Toronto3DAL}, though some require more complex processing (e.g. aggregation of sparse inputs~\cite{Caesar2019nuScenesAM}). 
We also encourage the use of autonomous driving simulators such as CARLA~\cite{Dosovitskiy2017CARLAAO}, SYNTHIA~\cite{Ros2016TheSD}
for \textit{synthetic} dataset generation, devoid of dynamic objects and subsequent registration problems. More extensive surveys on RGB-D and Lidar datasets are provided in~\cite{Gao2020AreWH, Firman2016RGBDDP}.

	\section{Conclusion}

This paper provided a comprehensive survey on contemporary state-of-the-art methods for 3D Semantic Scene Completion. We reviewed, and critically analyzed major aspects of proposed approaches, including important design choices to be considered, and compared their performance in popular SSC datasets. We believe that this survey will support further development in the field, aiming to provide new insights and help inexpert readers to navigate the field.

	\bibliographystyle{spmpsci}
	\bibliography{biblio}

\end{document}